Review Article

**Title:** A review of technical factors to consider when designing neural networks for semantic segmentation of Earth Observation imagery

**Authors:**

Sam Khallaghi, J. Ronald Eastman, Lyndon D. Estes

**Affiliation:**

The authors are affiliated with Clark University's Graduate School of Geography [Worcester, MA]. Ron Eastman is with Clark Labs. Sam Khallaghi is the corresponding author.

**Abstract:**

Semantic segmentation (classification) of Earth Observation imagery is a crucial task in remote sensing. This paper presents a comprehensive review of technical factors to consider when designing neural networks for this purpose. The review focuses on Convolutional Neural Networks (CNNs), Recurrent Neural Networks (RNNs), Generative Adversarial Networks (GANs), and transformer models, discussing prominent design patterns for these ANN families and their implications for semantic segmentation. Common pre-processing techniques for ensuring optimal data preparation are also covered. These include methods for image normalization and chipping, as well as strategies for addressing data imbalance in training samples, and techniques for overcoming limited data, including augmentation techniques, transfer learning, and domain adaptation. By encompassing both the technical aspects of neural network design and the data-related considerations, this review provides researchers and practitioners with a comprehensive and up-to-date understanding of the factors involved in designing effective neural networks for semantic segmentation of Earth Observation imagery.

**Keywords:**

# 1. Introduction

In the past 50 years, remote sensing has provided researchers with unprecedented means to study the earth, particularly in monitoring natural and human-generated land cover types. Such efforts usually produce outputs such as LULC thematic maps, in which each image pixel is assigned a nominal label through a classification scheme chosen to meet the objectives of the research project (*1*). Satellite-based LULC products are mostly used as a surrogate to describe landscape structure, and play an important role in downstream applications to understand land transformation, ecosystem dynamics, terrestrial biodiversity, and human-environment interactions at different scales (*2*). Such thematic products also include more specific applications, such as crop field mapping (*3*), forest inventory (*4*), tree species mapping (*5*), delineations of building footprints (*6*) and roads (*7*), or burned-area detection (*8*).

Pixel-wise classification resulting in dense prediction has a long history as a research topic in remote sensing, and has covered a wide variety of classifiers, ranging from traditional statistical approaches (*9*, *10*, *11*, *12*) to more elaborate learning-based algorithms, such as kernel-based methods (e.g. support vector machine (SVM) and its variants) (*13*, *14*, *15*, *16, 17*), ensemble models like Random forest (RF) (*18*, *19*, *20*, *21*), genetic and swarm models (*22*, *23*), and Artificial Neural Networks. Alongside these algorithms, there has been significant effort devoted to techniques for handling mixed pixels in coarser spatial resolution images, such as spectral mixture analysis (*24*, *25*) and fuzzy-based classifiers (*26*, *27*, *28*). Given the moderate to low spatial resolution of most freely available Earth Observation (EO) satellite imagery, coupled with their typically high, scientifically calibrated spectral resolution (e.g., multi- to hyper-spectral bands), the RS community has traditionally relied on the spectral response of objects as the primary source of discriminative information. For this reason, many existing classifiers aim to predict each pixel's class separately, based on its spectral profile or with added engineered features (e.g., spectral indices; SI). This approach is mainly due to the high spatial extent of each pixel, which weakens the autocorrelation between pixels and negatively impacts the segmentation performance. However, the increasing availability of Very High Resolution (VHR) images has highlighted the importance of spatial contextual information related to object shape and textural patterns. Consequently, this has led to a shift in methodologies since the early 2000s, and this trend continues to gain momentum with the increasing accessibility of imagery from newer sensing platforms such as high-resolution small satellites (*29*) and uncrewed aerial systems (*30*). To this end, much research and effort is spent on the development of effective feature-selection and engineering methods based on geometric and morphological features (*31*, *13*), wavelets, dimensionality-reduction, and textural and contextual information (*32*, *33*, *34*, *35*). One of the most prominent examples of such efforts is the object-based image analysis (OBIA) procedure (*36*, *37*, *38*). In OBIA, a VHR image is first segmented using different techniques into pixel groups called image objects. At the next stage, spectral and contextual features that the human analyst finds useful will be extracted from these image objects and fed to the classifier of choice. The skill of the prediction in these models is highly dependent on the quality of the segmentation procedure. Furthermore, since the properties of many natural phenomena manifest at multiple spatial scales, it is usually necessary to segment imagery at multiple scales to build a hierarchy of objects as inputs to a classifier, in order to improve the accuracy of the prediction (*39*). Hossain and Chen (*40*), Wang et al. (*41*) and Fotso et al. (*42*) provide comprehensive overviews of segmentation and feature extraction procedures developed for RS imagery and Belgiu and Drăguţ (*18*), Phiri and Morgenroth (*43*),

Talukdar et al. (44) provide an in-depth reviews of the typology and inspiration behind many conventional satellite-based image classification techniques.

Besides the domain expertise required to manually engineer discriminative features, the major shortcoming of these approaches is that they separate the processes of feature extraction and classification (45), resulting in the selection of features that are too localized and do not generalize to larger spatial extents (46). Deep learning (DL) models, on the other hand, automatically extract the discriminative deep hidden features across a hierarchy of scales through cascaded layers in feedforward (e.g. convolutional neural networks; CNNs) or recursive neural networks (e.g. RNNs) and perform the classification as a single process, resulting in a model with higher expressive capacity and stronger generalization capability (47). Another major attraction of DL models, especially CNNs, is their ability to use both the spectral and contextual information to classify each image pixel, a process that is referred to as semantic segmentation, a term that originated in the computer vision community that is increasingly used in the rapidly growing segment of the RS community that uses DL models. More specifically, semantic segmentation describes the process of partitioning an image into exhaustive homogeneous regions by grouping pixels belonging to each object and assigning to them a semantic category (48). Semantic segmentation therefore simultaneously performs both classification (the assignment of semantics) and localization (the delineation of object boundaries) (49). It is important to note that semantic segmentation stops short of distinguishing individual objects, or instances of a nominal class, and that there are numerous DL algorithms like mask-RCNN (50, 51) and YOLO (52) and for instance segmentation but they are out of the scope of the current manuscript.

Although DL models reduce the expertise required to engineer useful features, they require a new set of skills related to building or adapting models to make them effective for a particular problem. Unfortunately, this skill-set is not easy to acquire, for several reasons. First, the field of DL is relatively new but is growing fast and already has a huge family of models (e.g. deep belief networks, graph networks, CNNs, RNNs, GANs), with an astonishing number of subgroups and variations regarding both the architectural design (e.g. component modifications, hybrid models, addition of specific-purpose-build modules) and the type of tasks (e.g. object detection, semantic segmentation, instance segmentation, image captioning). Many of these models are experimental, and require further modifications to work when applied to real-world datasets. Second, the bulk of the published papers on DL come from other disciplines that focus on classifying conventional photos (e.g. distinguishing cats from dogs) or medical images. Models such as Unet were originally designed for semantic segmentation of medical imagery, but have proven useful for satellite imagery, as the concept of localization and image dependencies are important for both kinds of imagery, and both face similar issues of increasing intra-class and decreasing inter-class variance as image spatial resolution gets finer. However, despite such similarities that ease the application of DL to remote sensing imagery, satellite imagery has a number of complexities that require special adaptations to DL models. These include the larger variety in imaging sensor characteristics (i.e. multi-modal data), the distinctive spectrum characteristic of the target classes, and the quality and size of the training dataset (53, 54). Another factor is heterogeneity within the image arising from the large swaths covered by overhead sensors, which causes similar objects to appear different because of variation in viewing angle (*55*). Furthermore, classes of interest are often arranged as dense clusters of small objects, with low contrast and ambiguous boundaries (*56*, *57*).

There are already several existing articles that review the applications of DL models in remote sensing, which provide a broad overview of important model families with specific examples of common applications for each model family (*58*, *59*, *60*, *61*, *49*). There are also a number of more narrow reviews that focus on specific semantic segmentation use cases, such as agriculture (*62*, *63*), vegetation (*64*), roads (*65*) and urban features (*66*), or specific image types like HIS (*67*, *68*, *69*), or data collected from Uncrewed Aerial Systems (*70*). Yet there are still other reviews fully dedicated to semantic segmentation like Vali et al. (*71*), Borba et al. (*72*), Yuan et al. (*73*), although these may lack details regarding the technical aspects of DL models.

This paper attempts to complement this existing body of literature by providing a comprehensive technical review of the common design patterns of network architectures optimized for semantic segmentation as applied to satellite imagery. To this end, we focus on several prominent families of models, including CNNs and vision transformers, for their superb performance on grid-like input and ability to consider contextual information, as well as RNNs, for their ability to process sequential input, which is becoming increasingly important in modern remote sensing as the temporal interval of satellite missions shrinks and the length of record in image archives increases. In addition to the primary focus on remote sensing, our review also includes promising architectures and procedures from other disciplines, such as medical image processing and computer vision. We also introduce generative models, specifically GANs, and explain their use cases for image synthesis and translation, and how these structures are utilized in advanced training frameworks with semi-supervised learning and domain adaptation. Although this class of models has only recently been applied to remote sensing, they show great potential and are increasingly used for semantic segmentation in other areas. We assume that readers are already familiar with the basic terminology, procedures, and components of neural networks, such as backpropagation, optimization techniques, learning rate policies, and initialization and regularization techniques. For an overview of these aspects, we recommend Li et al. (*74*), Ruder (*75*), Andrychowicz et al. (*76*), Smith (*77*), Wu et al. (*78*).

Our review is structured as follows: Section 2 builds the technical foundation and covers the required terminology, mainly for CNNs, through an in-depth investigation of the basic components of such networks, and the key models for semantic segmentation. Section 3 puts the components together and explores the design pattern of CNNs used for feature extraction. Section 4 covers important design considerations for CNN structures optimized for dense prediction, and explores strategies for boundary refinement and adaptations for handling multi-modal datasets. It also covers attention mechanisms, and investigates the different formulations of these structures and the ways they are integrated into existing CNN networks to boost segmentation performance. Section 5 is devoted to network designs that can handle sequential inputs, including both the CNN and RNN families. Section 6 introduces vision transformer models and explores common architectures used for semantic segmentation. As no design consideration can be separated from the input data and its characteristics, especially in data hungry neural networks, Section 7 focuses on data considerations, and is divided into 4 parts. Part 1 covers common pre-processing steps, such as chipping large images and normalization techniques. Parts 2-4 cover the characteristics of labeled datasets that have important implications for model design and the learning approach, including the impact of limited labels (part 2), class imbalance (part 3), and label quality (part 4). The manuscript is concluded in section 8.

## 2. Basic components of Convolutional Neural Networks

Convolutional neural networks (CNNs) are specialized, feed-forward neural networks inspired by the mammalian brain's visual cortex (*79*), which excel at processing grid-like structures like images or videos. A CNN employs a hierarchical structure of convolution and downsampling layers, incrementally processing visual patterns to extract semantic-rich features for recognition tasks (*80*). Primitive features like gradients, edges, or blobs are detected in early layers, while complex, semantic-rich features are identified in later layers through combinations of lower-level features (*81*). The resulting feature maps are processed by fully connected layers and a classifier like SoftMax[1], providing a probability distribution over class labels (*6*). CNNs' use of sparse connections with high parameter-sharing reduces trainable parameters without losing many generalities, leading to an easier optimization problem compared to simplifying optimization compared to fully connected layers in Multi-layer Perceptrons (MLPs). Modern CNNs comprise specific components, each with a purpose, and understanding these offers insights into the workings of CNNs and directions for active research.

### 2.1. Convolution layer

A convolution (conv) operation multiplies the values of a pixel (e.g. central or skeletal pixel) and its neighbors by their corresponding values in a weight matrix using a dot product operation. In this way, each pixel is transformed into a weighted linear combination of the values within a spatial neighborhood determined by the kernel size (*82*). A kernel with the same size and weights is applied through a sliding window strategy to each pixel and its neighborhood, effectively dividing the image into small blocks and looking for the same pattern in each block. This approach guarantees translation equivariance in the network (*83, 84*), which means that a slight shift in the input (e.g. movement of the object of interest in the image) doesn't change the model's prediction (*85*). Each filter in a convolution layer acts as a feature detector that can be trained to detect a particular type of spatial-spectral pattern from the same transformation of the input space. The output of a conv filter is a 2D representation of the local patterns known as a feature map. Usually each conv layer needs multiple trainable filters to detect different useful features required to recognize complex patterns. The number of feature maps produced by a conv layer is known as the layer width, which is strongly related to the expressive power of the model (*86*). As a sliding window operation, convolutions are commonly categorized based on the number of directions the operation can be applied, which is determined by the dimensionality of the input tensor and shape of the sliding kernel. 1 to 3D convolutions are common and supported by all DL packages. When applying a 2D standard convolution on a 3D input, such as a multi-spectral satellite image, the filter depth must equal the number of input channels. A 2D convolution applied in this way learns from the joint spatial-spectral distribution of the input imagery, by leveraging the spatial correlation among neighboring pixels in all the channels to detect contextual dependencies (87). In other words, detected dependencies are local in the spatial but global in the spectral domain as illustrated in Figure 1.

---

[1]Softmax is a mathematical function that turns each logit, which is the output produced by the final layer of a network, into a probability by taking the exponent of the logit and normalizing it by the sum of the logits so that the entire vector sums to 1.

Convolution operations have several important free parameters. Kernel size defines the spatial extent of a region searched for local patterns also known as receptive field (RF). The choice of kernel size can impact the model's capacity and generalization ability. Networks developed with small kernels are typically better at preserving small objects, but can divide larger objects into separate pieces, while larger kernels better capture larger objects, but might group together small adjacent objects (88). If a kernel of size 1 is used, the conv operation is analogous to a single hidden layer MLP that acts on each pixel over the input features and produces an output with the same spatial size in which each pixel (e.g. spatial position) is a linear combination of the values in the input channels (89). Such 1×1 convolution are typically used to compress or expand the feature space. Another important free parameter is stride, which controls how the kernel is moved over the input image. Stride values can be 1 or larger, with larger values increasing the receptive field while downsampling the output at a factor equal to the chosen stride. Strides smaller than the kernel size cause the kernels to overlap. Luo et al. (90) showed that pixels in the center of the receptive field has higher impact on the response of an output unit with quick decay in the importance as moving away from the RF center showing a Gaussian distribution and call it effective receptive field which is a fraction of the theoretical RF. According to Sharir and Shashua (91) conv layers with overlapping receptive fields exponentially increase the expressive capacity of the network compared to CNN designs based on non-overlapping conv layers.

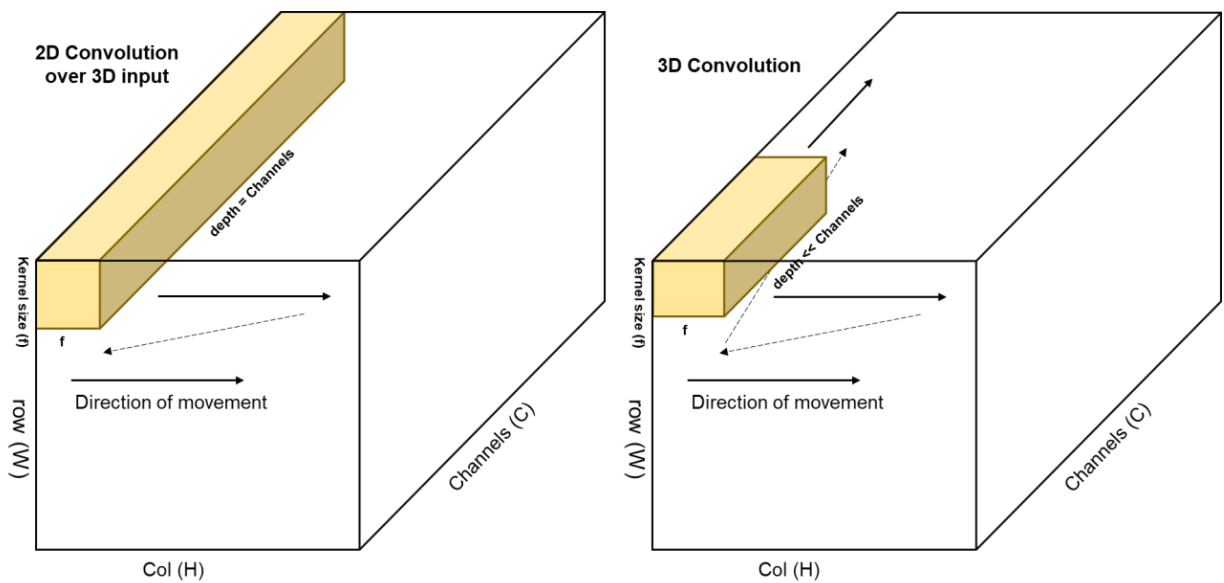

Figure 1. Comparison of 2D and 3D convolution. In a 2D convolution over a 3D image cube, the filter always extends to the full depth of its input and outputs a 2D feature map. The operation is performed by multiplying each kernel of the filter with the corresponding image band elementwise and adding the values. The output is a vector with the length of C. The sum of this vector generates a single position in the conv output. Summing all the C values for each pair of band and kernel as the kernel slides will build the final 2D output. In 3D convolution the filter depth is usually much smaller than the number of input channels which results in a 3D output.

Even with a stride of 1, the size of the convolution output is ($kernel\ size - 1$) pixels smaller in each of the spatial dimensions than the input image, which is referred to as a valid convolution. To include pixels near image borders in the convolution calculations while preventing shrinkage of the extent of the output, the input image is padded with extra rows and columns of pixels, which are either filled with zero or filled with duplicates of the edge values (reflection) or values from the opposite edge of the image (mirroring) (92). To make the conv output the same size as the input, the padding parameter should be set to $\frac{f-1}{2}$ for each of the spatial dimensions, where f represents the kernel size. For an input of dimension $R^{H \times W \times C}$ where $H = W = n$, the output of the convolution operation with kernel $k \in R^{f \times f \times C}$ and stride of $s$ and padding of $p$ is of the size $\lfloor \frac{n+2p-f}{s} + 1 \rfloor$. The number of trainable parameters for a kernel will be $[(f \times f \times C) + 1] \times M$ where M represents the number of different filters applied on the input and 1 accounts for the bias term associated with the kernel. The total number of operations associated with the conv layer is the multiple of the number of kernel parameters by the spatial dimensions of the convolution output. The number of trainable parameters in a 2D convolutional layer increases linearly with the kernel size and the number of input and output channels. Based on findings of Islam et al. (93), CNN models that use convolution layers with zero-padding significantly outperforms the model with no padding as zero-padding can implicitly encode absolute position information in feature maps throughout the network. Murase et al. (94) argue that positional encoding can improve CNNs' robustness towards image degradation, but its usefulness depends on the task and dataset. For instance, the position and orientation of objects in satellite images are less predictable than in typical photographs found in databases such as ImageNet and PASCAL VOC due to the varying angles and heights from which satellites capture images. This means that the biases found in standard image databases, such as central object placement, might not be as prevalent. The nature of satellite images might require CNNs to rely more on other information, such as relative spatial relationships, texture, and color, rather than absolute position. It is also worth considering that position information might still be relevant in some contexts in satellite imagery, such as consistent landscape features, roads, or buildings, but it would need to be evaluated on a case-by-case basis. Different variations of the convolution operation have been introduced, with each bringing a new set of functionalities, as well as extra hyperparameters that need to be selected. Some of these variations include dilated convolution (95), fractionally strided transposed convolution (96), deformable convolution (97), and depthwise separable convolution (98), which will be discussed in subsequent sections.

## 2.2 Activation Function layer

Each convolutional layer is accompanied by an activation layer, which applies a non-linear, differentiable function to the convolution output of the current layer (e.g. pre-activation values after the affine transformation of the input) element-wise to produce activation maps of the same dimensions. The activation (or transfer) function plays a crucial role in deep networks by preventing the network from degenerating into a single-layer linear network (99, 100). There are a number of activation functions available, the choice of which can significantly impact model performance, as some functions better approximate the distribution of certain inputs than others (*101, 102, 103*). Further details on available activation functions are provided by Nwankpa et al. (*104*) and Zheng et al. (*105*). Here we review several of the more widely used variants.

The sigmoid (Logistic) function squashes the entire real line into the interval [0,1], centering at point (0, 0.5) that can approximate a step function to any arbitrary level of accuracy and was initially the most widely used non-linear function, but has been largely replaced because of problems with saturation near small negative or large positive values (e.g. horizontal asymptotes of the sigmoid), which causes the gradients to tend towards zero (*106*). This prevents the propagation of gradients from higher layers in the network to the earlier layers, which is commonly known as the vanishing gradient problem that stalls the learning process (*102*). However, the probability-like output of the sigmoid function makes it a popular choice for gating mechanisms in Recurrent Neural Network (RNN) variants or attention modules, and it is also used as a simple classifier at the top of the CNNs to convert prediction scores (e.g. logits) to proper probability for binary classifications. Hyperbolic tangent (tanh) is a scaled version of sigmoid, which is centered around the origin and monotonically maps its output to the range [-1,1] which brings it the advantage of mapping a negative input to a negative range. Tanh is a common choice in RNNs, but, like the sigmoid function, still suffers from saturation around its bounds, which is common to all s-shaped activation functions. A deeper explanation for why gradients vanish or explode arise from changes in the scale of the gradient magnitude as it propagates through layers in the forward or backward paths. Besides the choice of the activation function, the architectural design is an important factor in controlling the flow of gradients and stabilizing the scale change, as we see later on in the design of ResNet in CNN models and gating cell-structures among RNN models.

Sigmoid functions are now largely replaced by the Rectified Linear Unit (ReLU) family of activation functions (*106*). ReLU (*107*) is a piecewise linear function that thresholds its input based on the sign where the positive values are unchanged and negative ones become zero. The power of ReLU is in its ability to treat parts of its domain differently, with the function imposing more non-linearity as the model gets deeper. ReLU does not suffer from the saturation issue for positive values and is much faster compared to S-shaped functions, but it cannot naturally handle zero values, as it is not differentiable at this point on its domain, and it omits all the negative inputs (*103*). There are variants of ReLU that are treating negative values differently to address the "dying ReLU" problem when the gradient of the ReLU becomes zero, resulting in a dead neuron that cannot be activated during training, and thus its corresponding weights are not updated, leading to a suboptimal solution (*108, 109*). Leaky ReLU (*101*) uses a small constant like 0.01 instead of zero as the slope to deal with negative values in its domain. Parametric ReLU (PReLU) (*110*) uses a linear function with a learnable slope parameter to vary the activation function on different channels for non-positive inputs. The exponential linear unit (ELU) (*111*) uses the identity function for positive input and an exponential function in the form of $\alpha(e^x - 1)$ otherwise. ELU is differentiable at zero and, by reducing the bias shift in feature activations by setting the mean activation of each channel to zero, produces a normalization effect, generally leading to a faster and easier learning process. Scaled Exponential Linear Unit (SELU) (*112*) is an extension of ELU with more self-normalization effect, stabilizing both the mean and the variance of the layer by introducing an extra learnable parameter. Gaussian Error Linear Unit (GELU) (*113*) is a non-monotonic smooth function that incorporates elements of non-linearity across both positive and negative domains. This infers that the derivative of the function is non-zero across all input values, thereby allowing for continuous gradients and more nuanced learning in neural networks. GELU integrates the activation function with dropout, and zone-out regularizers. The latter two are techniques employed to improve the model's generalization capabilities and prevent overfitting, a situation where the model becomes excessively specialized to the training data. Dropout, specifically, operates by randomly nullifying a proportion of neurons in a layer during training. This is akin to temporarily "silencing" them, thus preventing certain paths in the network from being overly dominant (114). The effect of dropout is a simpler network with reduced correlation between connected units (e.g. less co-adaptation), which mitigates overfitting effects when the model is large or training dataset is limited. Zone-out, on the other hand, works by randomly retaining the previous values of some neurons during training, instead of updating them. This induces a form of noise or randomness that ensures the network doesn't overly rely on any specific features from the input data. GELU can be conceptualized as a form of stochastic regularizer, which means it introduces a random element (e.g. noise), to the inputs of the neural network, with the goal of enhancing its ability to generalize from the training data to unseen data. The noise is derived from a zero-one mask, which is essentially a randomly generated filter following a standard Gaussian Cumulative Distribution Function (CDF). This mask is applied to the inputs of the GELU function to determine which inputs are retained and which are dropped out (i.e. set to zero). The effect of the zero-one mask is contingent on the value of the input. Lower valued inputs have a higher likelihood of being dropped out, whereas higher valued inputs are more likely to be retained. This dynamic application of the mask thus adds a further layer of randomness and prevents the model from becoming overly reliant on specific inputs or features. Hard-Swish is also a non-monotonic activation function that can smoothly interpolate the non-linearity between ReLU and a linear activation (115). Softplus is a smooth version of ReLU formulated as $ln(1 + e^x)$, which is also differentiable at zero and produces non-zero gradients over the negative part of its domain. Softplus is effective when paired with dropout regularization, in which the activations from some units (e.g. image positions) are temporarily removed from the forward or backward calculations based on a probability threshold (116). Smooth activations like GELU and Softplus are the most common non-linearity used in vision transformers (117).

## 2.3. The Batch Normalization (BN) layer and its alternatives

Batch Normalization (BN) is a process in which the outputs from a prior network layer are standardized, re-scaled, and shifted in order to keep the form of layer distributions fixed, which results in faster training and improved performance of the network. BN layer is commonly added after each trainable (e.g. weight) layer and before the activation layer, converting the distribution in each channel of the mini-batch into a standard distribution with mean of zero and unit variance (118). Batch normalization is followed by a channel-wise affine transformation with learnable parameters, which adjusts how closely the channel distribution approximates the Standard Gaussian distribution. To enable different batch sizes to be used during training and inference, a pre-computed exponentially weighted moving average (EWMA, also known as the running average algorithm) is used as a rough estimate of the training mini-batch mean and variance during inference. The EWMA algorithm uses a hyperparameter, momentum, to control the contribution of earlier mini-batches in the running statistics calculation (119).

BN is tremendously effective in improving training efficiency, although the reasons why are still debated. One explanation is that BN reduces the Internal Covariate Shift (ICS), which is the change in the statistical distribution of the inputs to a layer as the previous layers get updated during each training epoch. This can make it difficult for the neural network to learn because the optimal parameters for one layer may not be optimal for the next layer. BN can help to reduce ICS by normalizing the inputs to each layer to have zero mean and unit variance. Another possibility is that BN simplifies the optimization problem by smoothing the loss function, reducing the number of sharp local minima, better gradient flow, less sensitivity to the initial state setting of the network, and provides the opportunity to use a larger learning rate (*120, 121, 122*).

In addition to placing the BN layer between the conv and activation layer, it can also be placed after the activation layer or before the conv layer. An empirical study by Hasani and Khotanlou (*123*) explored the effects of BN position on the required number of iterations for the model to achieve a desired accuracy. The study found that the effect of BN position varied between models and datasets, but ResNet architectures achieved the highest performance with the BN layer placed before the conv layer. This is consistent with the proposed structure of residual block v2 (*124*). The study also suggested that in this particular configuration, BN acts as a gradient noise suppressor similar to the Normalized Least Mean Square (NLM) function used in conventional Adaptive Filter Theory (*125*).

The sensitivity of BN to different batch sizes and independence of samples within a batch is a known issue. As several studies (*126–128*) indicate, BN performs poorly on small batch sizes and/or when samples in mini-batches exhibit strong autocorrelation. Statistics computed from small batches during the training stage, particularly those lacking independence, can be unrepresentative of the broader population, resulting in a distributional shift between the training and inference processes. This distributional shift can negatively impact the model's ability to generalize, which is a critical factor for deep learning models. As a result, a number of other normalization techniques have been developed (see Figure 2 for a visual representation of these techniques compared to BN). One potential fix offered by Ioffe (*129*) is Batch Renormalization (BR), which uses an extra affine transformation to relate mini-batch moments to running average moments, correcting for the difference between the population expectation and mini-batch statistics. Despite being an improvement over BN for small batch sizes and non-independent samples, BR is still inferior to using BN with larger batch sizes, typically 32 or more, for visual recognition tasks.

Another strategy is to break the dependence of normalization on batch dimension. In Layer Normalization (LN) (*126*), the statistics are calculated over the channel and spatial dimensions of each individual sample in the mini-batch, rather than over the entire batch. This ensures that the normalization factors are not affected by the statistics of other samples in the batch. LN fits a distribution with shared mean and variance over all channels for each sample, thereby assuming equal contribution by all feature maps. LN is a popular choice for sequential input in both recurrent networks and transformers. Instance Normalization (IN) (*127*) is a normalization technique that calculates the mean and variance statistics for each channel of each sample in the mini-batch along the spatial dimensions, thereby enforcing the assumption of spectral independence. The benefit of this assumption is that it allows IN to better preserve the texture and style information of the input image. Since the statistics are calculated for each channel independently, IN can be used to normalize the style information of an input image without affecting the content information which makes IN a common choice in style transfer approaches (see Section 7.2.1). Group Normalization (GN) (*128*) is a trade-off between LN and IN. It aims to overcome the limitations of BN and to provide effective normalization for small batch sizes. GN divides the channels of a feature map into multiple groups, where the statistics for each sample are calculated independently within each group, using the mean and variance of the activations in that group. Reducing the number of groups to 1 turns GN into LN, while prior work has found that any group number between these two extremes has a better normalization performance than LN and IN.

As with BR, these alternative methods of normalization that try to solve the problem using normalization layers independently from the batch dimension, are typically outperformed by BN with large batch sizes. In contrast, Filter Response Normalization (FRN) (*130*) outperforms BN in various batch size regimes and can handle inputs with varying magnitude or dynamic range. The FRN layer has two components. The first is a normalization step that calculates the second moment for each channel of each sample. This independent normalization of each sample in the mini-batch removes scaling artifacts associated with filter weights and pre-activations. The second component is a Threshold Linear Unit (TLU) activation function, which is ReLU with an extra learned threshold that compensates for the lack of mean-centering in FRN. The threshold is learned during training and is specific to each channel of the input. For best performance the TLU should have a warm-up phase in which the model is first trained for a number of epochs through a policy that increases the learning rate very slowly from an initial value close to zero up to a user-defined peak value. FRN is used with promising results with encoder-decoder structures in the road extraction task (*131*, *132*).

All the normalization layers discussed so far generate global statistics over the spatial dimensions. In contrast, Positional Normalization (PN) (*133*), operates on the spatial dimensions of the input tensor, normalizing each position of each sample independently across all channels. This can be particularly useful for generative models, where the goal is to learn a probability distribution over the input space. PN can be applied in conjunction with other normalization techniques, including BN. In fact, the original paper on PN recommends using it together with BN, as the two types of normalization can complement each other. For example, BN can help to control the scale of the activations, while PN can help to normalize the spatial structure of the input. Local Context Normalization (LCN) (*134*) is a technique that considers the statistics of a local neighborhood in the spatial domain over grouped channels for the normalization. Depending on the selection of parameters, LCN can function as GN, LN or IN. LCN is specifically designed for visual recognition tasks with competitive or better results compared to BN, and is helpful with RS image analysis, where image chipping procedure is common but shifting the input patch by a few pixels can lead to different predictions over the patches. There are also normalization techniques that instead of the feature (activation) maps work on the weights directly (*135*, *136*, *137*, *138*, *139*). Summers and Dinneen (*140*), Wu and Johnson (*119*) further review the caveats of BN and explore some of the proposed solutions and work-arounds.

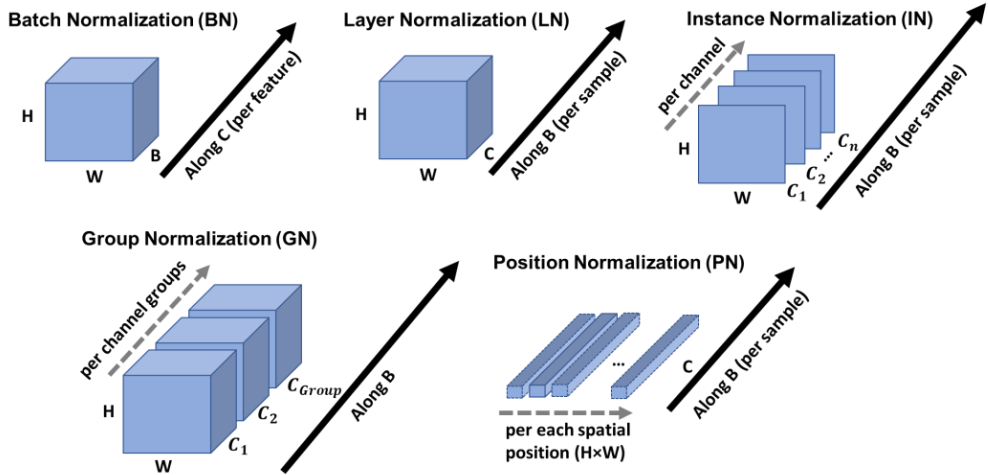

Figure 2. An illustration of the difference between normalization types. The best way to interpret the figure is to imagine that each element along the black line calculates the normalization statistics from a multidimensional array of values represented by the blue cubes. B, C, H and W are respectively the batch dimension, channel dimension and height and width spatial dimensions.

## 2.4 Downsampling layer

Downsampling, the process of reducing the spatial resolution of the input, can be achieved either using a convolution layer with a stride larger than 1 (see 2.1.1), or through a pooling operation. The aim of the downsampling layer is to achieve translation invariance and improve upon the linear increase in the receptive field when placed after the standard conv layers (*141*). Both strided convolution and pooling increases the network's receptive field multiplicatively (*142*). Strided convolution subsets values from fixed positions in the conv window, and thereby disregards the importance of the selected activation values, which can limit translation invariance in the output (*143*). A layer with a pooling operation aggregates values within blocks dividing the spatial dimensions of the inputs for each channel, typically using functions such as mean (*144, 145*) or max (*146, 147*), resulting in the same number of outputs as the input feature maps, but with spatial dimensions reduced to $\lfloor \frac{n-f+s}{s} \rfloor$, where n, f, and s represent the height and/or width of a square input, square kernel size and stride respectively (*148*). Kernel size and stride control the size and arrangement of pooling bins. When the kernel size equals the stride (f=s), the pooling operation will use non-overlapping windows. This is the most common setup, as it ensures a simple and straightforward downsampling of the feature map. On the other hand, if the kernel size is larger than the stride (f>s), the pooling operation will use overlapping windows. While this setup can be more complex to implement, it has been found to sometimes help in reducing overfitting and improving the model's accuracy and expressiveness, according to research by Krizhevsky et al. (*149*) and Gao et al. (*150*).

The deterministic nature of mean or max operations means that a pooling layer using such operations does not require any trainable parameters, which implies its operational cost purely depends on the input size. As a kernel-based operation, pooling can be done at different scales. When the kernel size of the pool is equal to the extent of the input features, the operation is called global pooling, which acts as a global descriptor that can be used in novel ways to boost the performance of semantic segmentation models (see 4.3 for further details). Besides increasing the receptive field non-linearly, pooling prevents the propagation of local patterns into the neighboring receptive fields, which causes the model to be less sensitive to image clutter and deformations (*148*). The level of translation invariance encoded into the model is controlled by the pooling kernel size and the operation used. This invariance is essential in semantic assignment as it focuses on the existence of a pattern rather than its location in the image. Shrinking the spatial extent through the pooling operation also reduces the number of trainable parameters for the consequent layers in the network, which is commonly leveraged to increase the network width.

Max pooling, the most commonly used pooling function for recognition tasks, assumes that objects in an image are locally consistent, and that this consistency can be used to understand an object from its parts. Max-pooling partitions the domain of a function describing the spatial distribution of activation maps, and uses the most representative value in each partition as the function descriptor in that particular subdomain, which simplifies the function and reduces its variance within the output representation, as the less relevant features are discarded. However, the assumption that the strongest activation is also the most discriminative in the sliding window is not always valid, as, for example, it might represent a high value noise (*143*). Mean-pooling assumes the same importance for activations within the pooling kernel, which causes blurry object outlines (*151*). Prior work has shown that for both conventional feature extraction methods and CNNs, the optimal pooling strategy lies somewhere between max and mean pooling, depending on the dataset (*152,* (*141*). One drawback of deterministic pooling strategies is that the kernel size imposes a fixed and non-adaptive importance measure on the local neighborhood, disregarding the individual characteristics of the input image. This imposition can actually increase the translation variance at inference time, and coupled with the chipping and stitching strategy can increase the misclassification rate (*153*).

Research into more efficient pooling strategies has led to the introduction of different functional forms, such as $L_p$ Norm pooling (*154*), in which a normalized measure of the activations that fall within the window are considered as aggregated output. In stochastic pooling (*155*), the aggregated value is sampled from the multinomial distribution fitted to activations inside the pooling kernel. Local Importance-based Pooling (LIP) (*156*) is particularly successful in preserving small objects, and acts as a form of local attention per channel, which learns the discriminative criterion for features in order to produce a softmax-based importance map that quantifies the contribution of activations inside its local neighborhood. Universal Pooling (*157*) is proposed as the generalization of strided conv, max, or mean pooling, and formulated as a local spatial attention module that is applied to each channel separately, determining the pooling mechanism by learning the softmax-normalized weights for each pooling kernel.

Regularized pooling (*158*) was designed to address the drawbacks of max-pooling, which is its overcompensation for image deformations, such as object displacement, that are actually present inside the local patch, which can result in confusion between similar classes, slow down network convergence, and increase the possibility of ending in a bad local minima, especially in early stages of training. As a solution, to fit to the actual existing deformations, the Max-pool deformations in a feature map are first captured for each pooling kernel through a local displacement measure, which can be conceptualized as the direction of an arrow, with its tail on the central pixel and the head on the local maximum. Then these local max-value directions are smoothed using a kernel-based moving average with the most regularizing effect for deformations, where the neighboring kernels shift to random directions. Unlike mean pooling, this strategy allows a non-maximum value to be pooled without directly smoothing over the actual input feature map, thereby helping to preserve the shape of objects. This approach is best suited for patterns with periodic structure. A limitation of regularized pooling is that it needs a large pooling kernel to outperform max-pooling as the overcompensation effect of max-pooling increases by the kernel size. A comprehensive review of pooling strategies is done by Akhtar and Ragavendran (*148*).

**2.5 Fully Connected Layer (FC) and classifier**

A fully connected layer (aka. dense layer) is equivalent to a single layer MLP, where each input unit is connected to every output unit. FC layers are placed at the top of the network. The interactions between pixels in the FC layer encode the global relationships between features in the semantically rich feature space of the last kernel-based layer (e.g. conv or pooling) of the network. The FC layer can also be formulated as a standard convolution with kernel size equal to its input spatial extent. The important difference between these two formulations is that the input to the standard FC layer is flattened into a vector, making it agnostic to input data structure, but the conv equivalent requires a grid-like input. Theoretically a conv layer can be converted to an FC layer by reshaping it to permit matrix multiplication, but such formulation is avoided in practice because it requires large amounts of memory (*159*). CNNs developed for image-level classification usually use 2 or 3 FC layers towards the end of the network. The number of trainable parameters for a FC layer directly depends on the input size and the required number of units for the current hidden state, which controls the function approximation ability of the layer. FC layers are computationally expensive and usually responsible for most of the trainable parameters in a CNN network. This expense motivated the replacement of one FC layer with a global pooling layer in some modern architectures like GoogLeNet and ResNet (*160*). The output of the last FC layer is raw prediction scores in the form of real numbers expressed as logits, which are then passed through a sigmoid to produce class probabilities for a binary classification, or through a softmax in a multi-class scenario. In models designed for semantic segmentation the last layer is a linear convolution layer that produces the prediction outputs as a vector for each pixel that provides the probability of membership for each class in the classification scheme. This type of structure is known as one-hot-encoding, which demonstrates how model output can be realized as a multi-band raster in which the number of bands is equal to the number of modeled classes. The output probability maps can be converted to a single class prediction by assigning to each pixel the class value corresponding to the maximum probability in its vector of predicted class probabilities.

**3. Backbone Networks**

CNN architectures were originally developed for assigning a label (e.g. cat, dog, car) to an entire image. These architectures are essential for extracting the semantic-rich features needed to categorize images, and are integrated into semantic segmentation models as so-called backbone networks or feature encoders. Backbone networks typically have a modular structure, with convolutional layers arranged into blocks consisting of different arrangements (e.g. parallel or serial) of a few convolutions and activations plus normalization layers followed by a downsampling layer. The effectiveness of CNNs relies on several design factors, including the depth of the network (i.e., the number of layers), the width of the network (i.e., the number of feature maps produced in each stage), and the resolution of the input images (i.e., the size of the image patch and kernel type and size). The choice of these design factors can affect the representational power of the features extracted by the network, as well as the model complexity and computational efficiency. For instance, a common design practice is to shrink the spatial size as the network progresses but increase the number of feature maps, which provides a trade-off between network approximation capacity and model complexity and roughly keeps the FLOPs stable among blocks (*86, 161*). Increasing the width and resolution are more costly than increasing the depth dimension, as doubling depth will double the FLOP (floating point operations required to perform a forward pass through the network ) cost, but doubling width or resolution will quadruple the cost (*162*). This section briefly introduces some of the most common backbone architectures and their design choices.

LeNet (*145*) and AlexNet (*149*) are examples of early networks that were successfully applied to the MNIST digit dataset and ImageNet dataset, respectively. VGGNet (*163*) is another old but still popular architecture that uses a simple stack of convolution layers, each with a 3×3 kernel, which serve to gradually increase the receptive field (RF) of the network while holding down the number of model parameters. The idea is that two consecutive convolutions with kernel size of 3 can resemble the RF of a single 5×5 convolution with lower number of parameters and a more non-linear form. Among different variations of VGG, models with 16 (VGG16) and 19 (VGG19) layers have become most popular as backbones for the semantic segmentation task due to their ease of implementation and adaptability for transfer learning.

Inception is another popular family of CNN models developed by Google and used extensively for feature extraction, with four main versions. Inception-v1, also known as GoogLeNet (*164*), uses multi-branching to develop complex, non-sequential networks that can capture invariances at different scales with reasonable computational cost. The structure is designed to learn independent and multi-scale features by organizing several parallel branches of convolutions with different kernel sizes, plus a pooling layer, into blocks called inception modules. Each branch in the block first captures cross-channel correlations using 1×1 conv, and then looks for the spatial correlations in a compressed space through multiple branches, with each examining a different spatial scale by using different kernel sizes. Outputs from these branches are fused through concatenation to generate the block's output. The other three versions of the Inception model mainly vary with respect to the stem subnetwork and how spatially separable convolutions are used to speed up computation by compressing a full rank matrix into two first rank matrices with fewer parameters (*118, 165*). A spatially separable convolution is a technique that factorizes a convolution into two separate operations in order to reduce the number of parameters and the computation cost of a convolutional layer while maintaining its effectiveness. More specifically, a spatially separable convolution decomposes an f×f×c filter into two filters of size f×1×c and 1×f×c where the first filter performs a 1D convolution along the vertical dimension while the second acts on the horizontal dimension (*166*). The output feature map is obtained by convolving the input volume with both the vertical and horizontal filters and summing up the results. Figure 3 illustrates the details of inception-v3. Inception-v4 along with many other modern designs are moving towards simpler networks that chain together a smaller number of specific modules with fixed kernel sizes. This approach reduces the complexity of adapting the network for a new dataset or task. However, the split-transform-merge approach for learning multi-scale features lives and is used in many architecture designs like Inception-ResNet (*167*), MobileNet family (*168, 169*) and ResNeXt (*161*) to name a few.

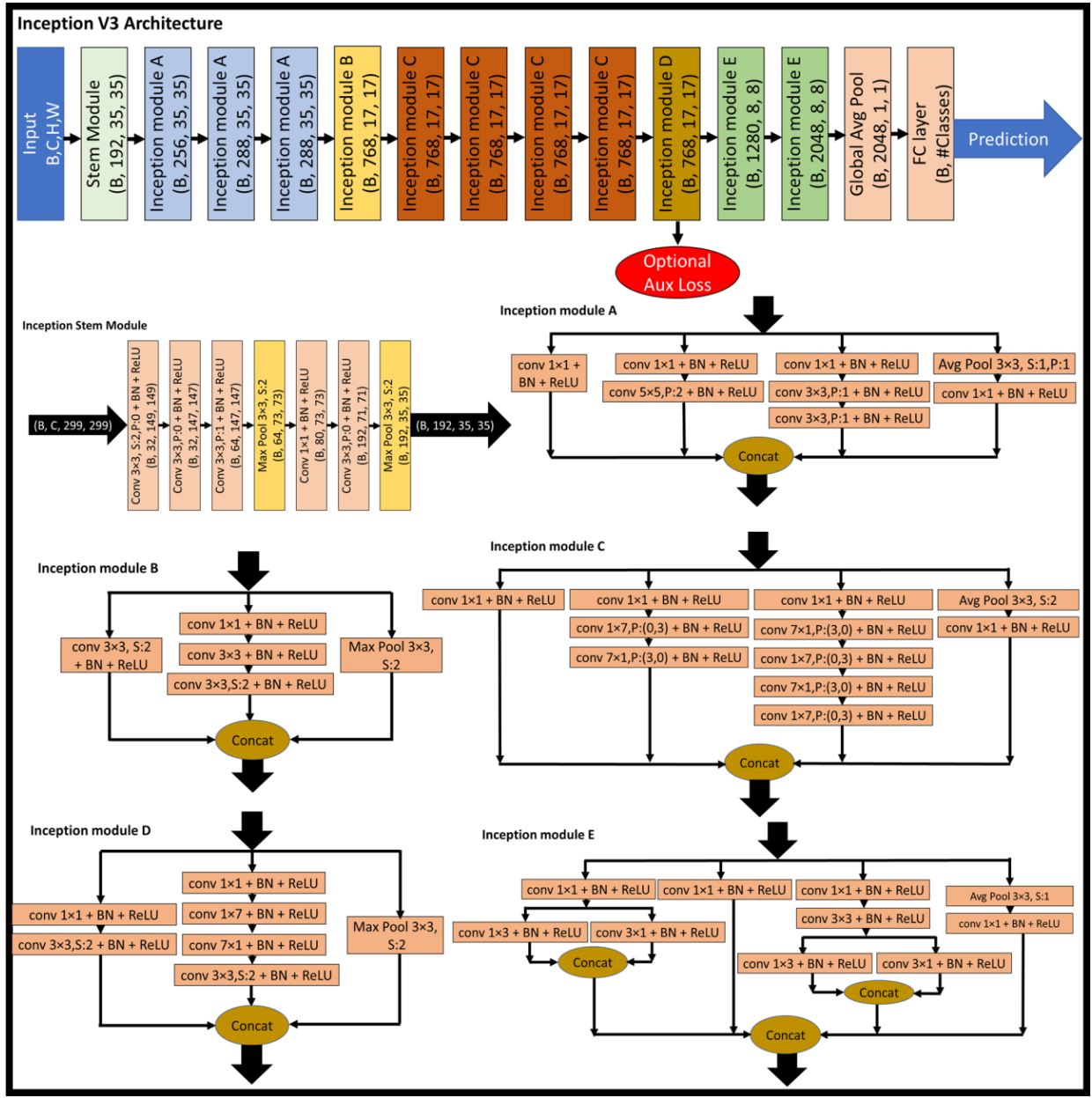

Figure 3. Architecture of Inception V3. This is one of the few CNN designs that use a complex set of modules and customized kernel sizes with tested high performance on different benchmark datasets.

ResNet (*170*) introduced the idea of residual connections, in which the output of every conv block is the summation of two branches: 1) the identity transformation of the input to the block; 2) a series of the convolution, BN, and activation layers (Figure 4), which are responsible for learning the residual between the input and output of the block. The residual connection branch provides the flexibility for the feature maps of the previous layer to bypass some convolutional layers, making the design an effective strategy against vanishing gradients, promoting easier back-propagation and faster convergence, while providing an easy way to increase network depth without hampering performance. Different versions of residual blocks were introduced that alter the ordering of the convolution, batch normalization, and nonlinearity operations within the residual branch (Figure 4) *(124)*.

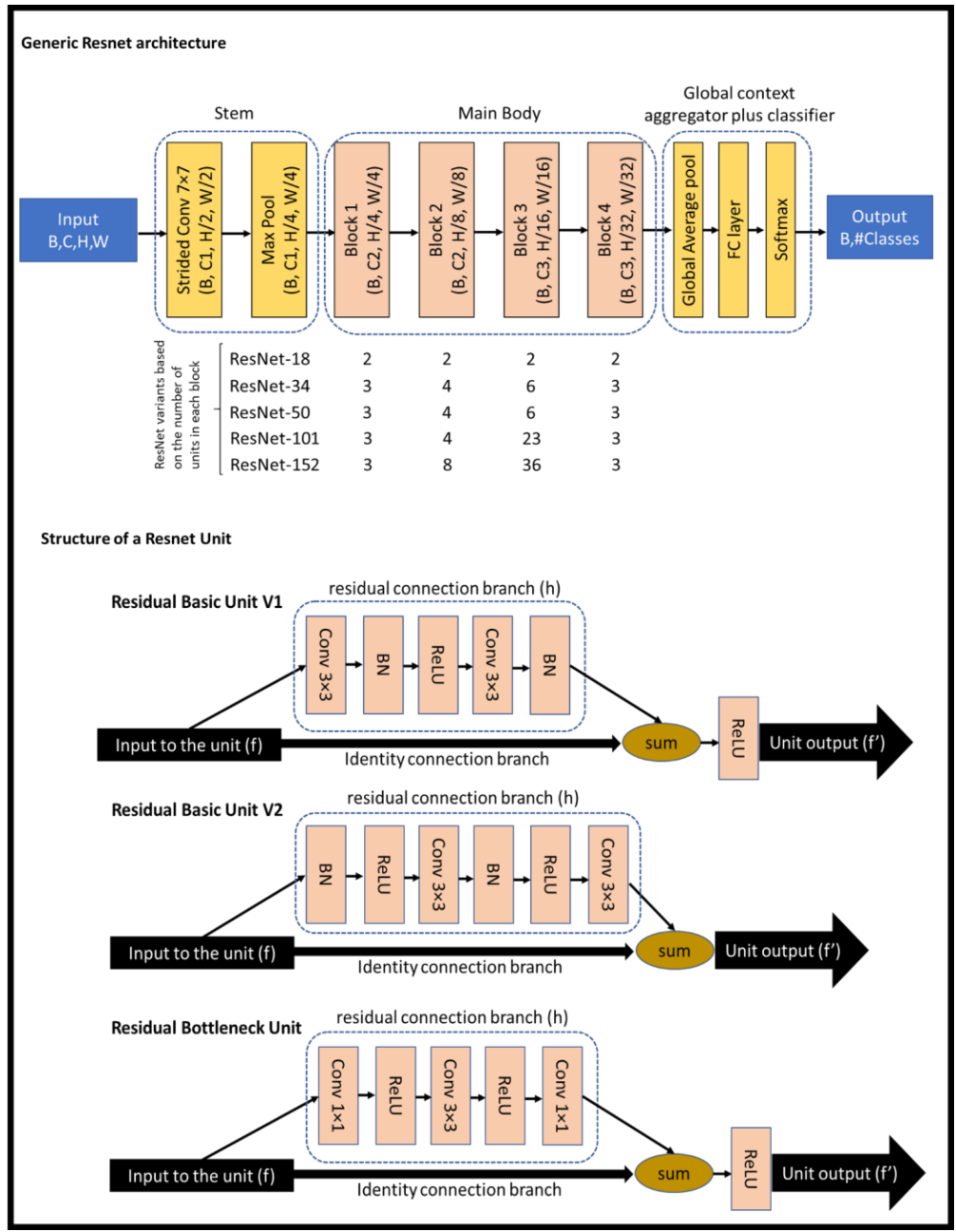

Figure 4. Schematics of generic Resnet networks. The stem subnetwork is used to reduce the scale to ¼. The main body usually consists of 4 blocks. Each block consists of chained primitive units. Notice that the bottleneck unit is a thinner (e.g. compressed embedding) version of the basic unit mostly used in deeper variants of ResNet to minimize the number of learnable parameters. The first 1×1 conv layer is used for feature reduction and the second 1×1 conv layer is used for feature expansion.

There are a number of ResNet variants that alter other architectural dimensions besides depth. For instance, WideResNet *(171)* tries to solve the diminishing feature reuse[2] problem in deep networks by increasing the number of features (e.g. width) of each Resblock, by using the widening factor hyperparameter, controlling the complexity of the model by adding drop-out to the residual connections of each layer in the network, and carefully choosing the depth of the network. They showed that a wide 16-layer deep ResNet can be as accurate as a 1000-layer ResNet model with thin blocks (e.g. using residual bottleneck unit), with a comparable number of parameters but a much faster training speed. ResNeXt *(161)* is a modification of the ResNet architecture that uses the split-transform-merge strategy within the ResNet blocks. It introduces bottleneck units that utilize group convolution, which helps to reduce the number of parameters and training time. The group convolution used in ResNeXt is a more general form of depth-wise separable convolution and divides the input channels into a user-defined number of non-overlapping groups controlled through the cardinality hyperparameter, before applying convolution within each group separately. These branches have the same topology to avoid introducing further hyperparameters, and their output gets merged through summation with a residual strategy, as illustrated in figure 5a. Using group convolution reduces the number of trainable parameters to $\frac{1}{number\ of\ groups}$ of the total parameters in a standard 2D conv layer *(173, 174)*. The authors argue that cardinality is an important design consideration, and demonstrate that increasing cardinality is a more effective strategy for improving classification accuracy than going deeper or wider. The authors of ResNeXt also proposed a new regularization technique called "DropPath" that randomly drops entire residual blocks during training to improve generalization. ResNeSt *(175)* replaces the original ResNet blocks with Split-Attention blocks, which are composed of channel-wise attention mechanisms that apply attention maps on the features groups in conv layers with a multi-branching arrangement and residual connections inside each block (figure 5b). DenseNet *(176)* takes the idea of residual skip connections to the extreme and connects all layers to each other through feature concatenation. This design is based on the rationale that gradient flow is improved because shallow layers are able to directly optimize deeper layers, and that the number of parameters required can be minimized by using smaller numbers of feature maps in each layer by relying on the dense connections from previous layers to provide information. In DenseNet the number of direct connections between layers increases from N to $\frac{N(N+1)}{2}$.

---

[2]The problem is equivalent to vanishing gradients but instead occurs in the forward pass of the deep networks where the propagated effect of convolution with randomly-initialized weights can make it hard for later layers in such a deep network to find meaningful patterns. This problem is also known as feature wash-out. An interesting remedy is introduced in Deep Networks with Stochastic Depth *(172)* which randomly drops entire layers during training to encourage the network to learn more informative features.

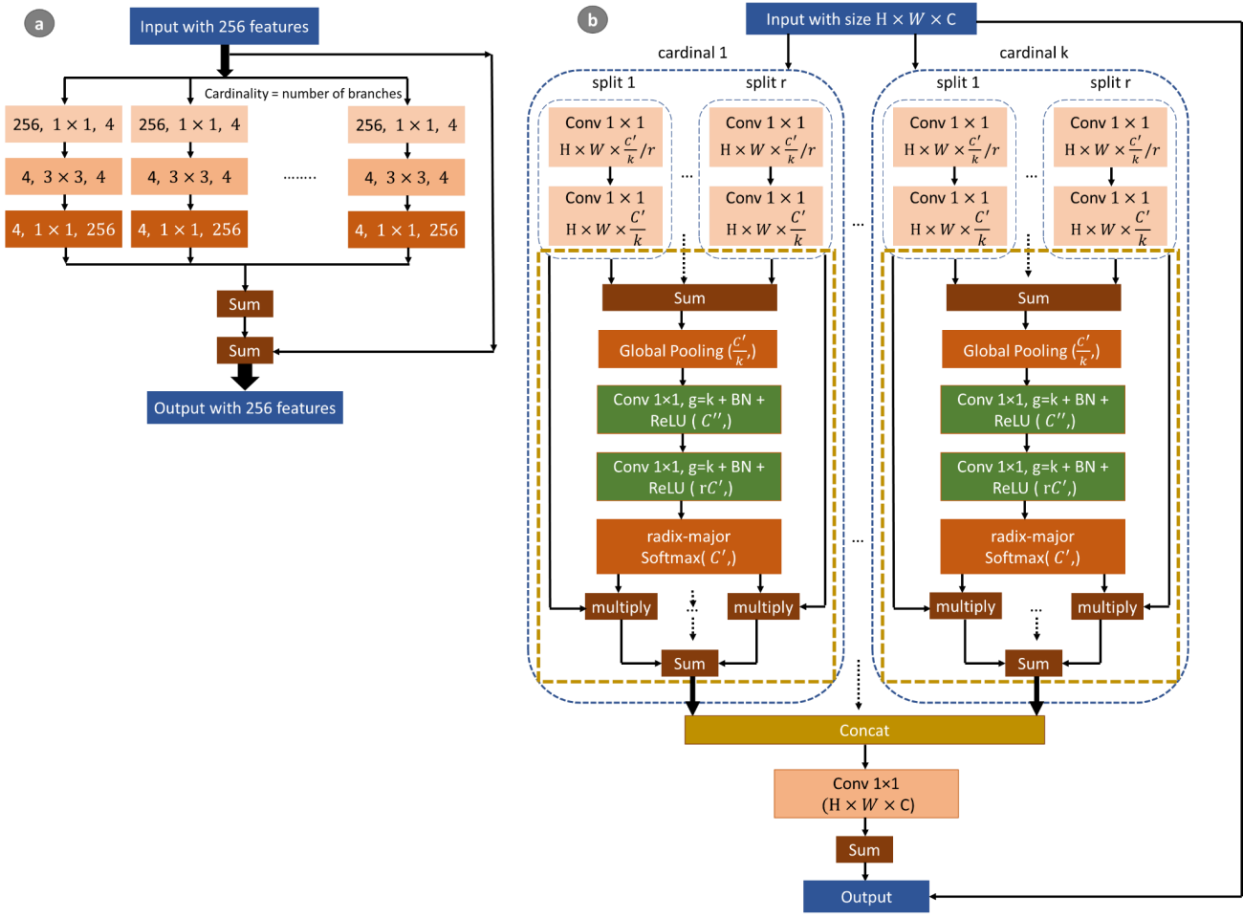

Figure 5. Two variants of the ResNet block. a) structure of a ResNeXt block illustrating the details of aggregated residual transformations with cardinality of 32 and 4 as the width of each group. The comma-separated numbers in each block are the number of input features, filter size, and number of output features respectively. Cardinality branches are in reality implemented using group convolutions. b) Conceptual structure of a ResNeSt block. This Block has an extra hyper parameter called radix that further splits each cardinality into smaller groups and passes them through a split attention before aggregating them to generate the block's output.

Capturing multiscale representations is also being investigated both inside a layer or throughout the whole encoder. Res2Net (*177*) uses a hierarchical structure to combine features with multiple scales in each layer, where each layer is divided into multiple groups, and each group has a separate receptive field size. The number of divisions is controlled by a scale hyper-parameter that decides on the number of splits to apply to the input channel dimension. Like group convolution, the input is split into multiple groups but the key difference with Res2Net is that each group can split multiple times, creating a multi-scale representation of the input feature maps. The "scale" hyper-parameter is controlling the number of splits and determines the level of detail captured in the feature maps. To ensure better fusion of multi-scale features, Res2Net introduces a hierarchical aggregation mechanism between the branches where each split in the same stage is aggregated by the following split in a tree-like structure (Figure 6a). This aggregation helps to preserve the spatial information and improves the representation capability of the network. On the other hand, MultiResUNet (Figure 6b) (*178*) uses a single branch serialized aggregation of multiscale context with a residual strategy to incorporate multiscale features into the encoder.

High-Resolution Net (HRNet) (*179*) is a neural network architecture for image classification and semantic segmentation tasks where high-resolution representation is maintained throughout the network by creating a multi-branch architecture with parallel processing and dense connections. The HRNet architecture consists of four stages, each of which contains a set of residual units. In contrast to the standard ResNet architecture, HRNet creates a new branch at each downsampling stage in addition to the original branch. This creates a multi-branch network with varying lengths, where each branch has a different resolution that remains constant along its length. The outputs of each branch are fused together through a dense connection at each branching stage. This parallel design and context aggregation strategy helps the network to maintain high spatial details with strong semantics (Figure 7).

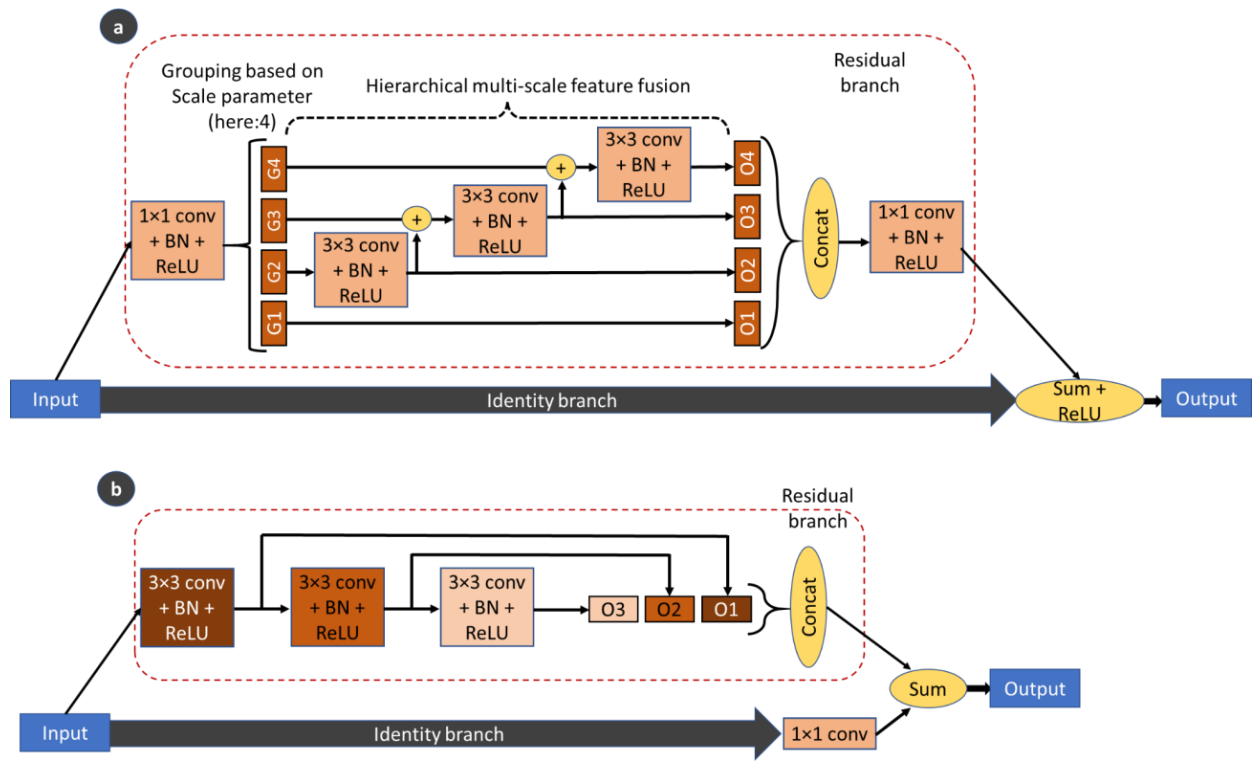

Figure 6. a) Structure of a layer in Res2Net implemented with multi-branching and hierarchical feature fusion. b) structure of a layer in MultiResUNet with simple multiscale aggregation using a single branch.

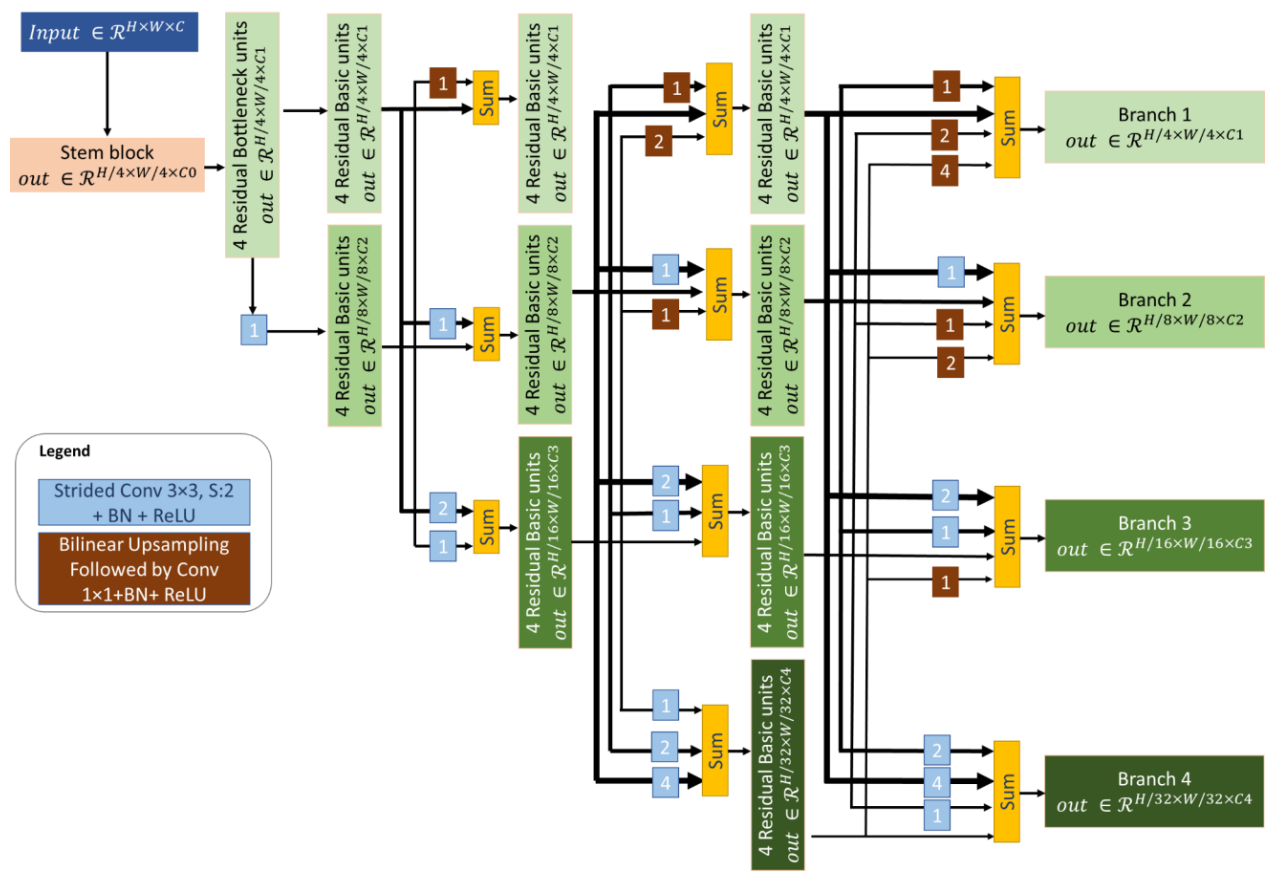

Figure 7. Details structure of HRnet network. Numbers on dense connections represent the number of layers color coded in the legend to downsample or upsample the input by the factor of 2. Different thickness of dense connections is only to ease visualization. At the start of each stage, the input image is first processed by a convolutional layer with a small kernel size to extract low-level features. The outputs of this layer are then fed into all the branches in the stage. Each branch performs a series of residual operations that increase the number of feature maps while maintaining the spatial resolution. The residual blocks in HRNet are similar to those in ResNet, except that they use a more efficient bottleneck structure with 1x1 convolutions to reduce the computational cost. After the residual blocks, the outputs of the branches in each stage are fused together through a dense connection, which sums the feature maps which are then fed into the next stage.

There are two hybrid versions of the Inception and ResNet models that were developed by the Google team, which combines the strengths of Inception and ResNet, using Inception modules for their ability to capture multi-scale features and ResNet-style residual connections for improved gradient flow and training stability (*167*). Another popular hybrid is the Xception network (*98*), which replaces the Inception-v3 convolutions with depth-wise separable convolutions that are further augmented by residual connections, resulting in a full decoupling of the spatial and channel-wise correlations that occur in a standard spatial-spectral convolution. A depth-wise separable convolution is a two-step process. First, a depthwise f×f convolution is applied to each input channel independently to capture spatial correlations, producing the same number of channels as the input. Then, these feature maps are concatenated along the channel dimension and fed through a pointwise convolution (i.e. 1×1 conv) to capture channel-wise correlations and generate the output feature maps. This factorization significantly reduces the number of trainable parameters in the convolution layer to just $[(f^2 \times C) + CM]$ where C and M represent the input and output channels, respectively. This approach is particularly effective when the input has high spatial correlation and few channel-wise dependencies (*180, 181*). The computational efficiency of depth-wise separable convolution makes it popular for building light-weight networks used for real-time semantic segmentation on resource-constrained platforms, such as mobile devices (*182–187*).

## 4. CNNs optimized for Semantic segmentation

To successfully classify remote sensing imagery, which typically entails assigning each pixel a label describing a class of interest, two contradictory challenges have to be overcome: 1) the assignment of proper semantics, which requires the model to be transformation-invariant, and 2) the preservation of spatial details, which requires the model to be transformation-sensitive (*188*). The intrinsic equivariance of the convolution operation and further downsampling through the pooling layer addresses the first challenge, but the output of backbone networks is much smaller in the spatial dimension than the original input, thus the model's output has to be rescaled to the input's dimensions while minimizing localization inaccuracies, particularly near object boundaries (*189*). Many architectural designs specialized for semantic segmentation have been proposed, which can be roughly categorized into three main groups: 1) encoder-decoder models, 2) designs based on dilation convolution and/or pyramid sampling strategies, and 3) those using aggregation modules based on attention mechanisms. In addition, there are also a vast diversity of models that mix aspects of some or all of these three primary design strategies. Many models are also accompanied by a post-processing step, such as Conditional Random Fields (CRF), which helps to improve predictions by using information on pairwise similarities of neighboring pixels in which the algorithms maximizes the similarity of those pairs that are assumed to belong to the same spatial context based on a predefined prior distribution (*190*). Many of these designs might also contain extra modules or training procedures designed for boundary refinement, and may be equipped with domain adaptation techniques to improve the performance and generalization of the model.

### 4.1 FCN

The Fully Convolutional Network (FCN) (*96*) was a pioneering effort to apply CNNs to semantic segmentation, and served as the basis for subsequent encoder-decoder models. Prior to FCN, many segmentation models used patch-wise training strategies, which involved feeding small image patches to a CNN to predict a label for the central pixel of each patch and repeat the process for every image position to generate a dense prediction output (*191*) This approach was computationally expensive and didn't fully capture the interdependencies between image pixels (*192*).

The original FCN uses a VGG16 as the backbone structure but modifies the network such that the fully connected layers are replaced with convolutional layers. This modification has three advantages: 1) it enables the model to preserve the spatial dimension of class probabilities; 2) it dramatically reduces the number of learnable parameters; 3) the model becomes invariant to the spatial dimension of input imagery used during training at inference time. The extracted features produced by the modified VGG16 are $32x$ smaller in spatial dimensions than the original input. To return the encoder output to the input dimensions, FCN introduced three models differentiated by name based on the direct upsampling factor used in the model (32, 16, and 8). In all three FCNs (Figure 8), raw prediction scores are obtained through a linear 1×1 conv layer at the top of the network that maps the learned features to the label space. FCN-32 does a direct upsampling with a factor of 32 from the last layer of the encoder to recover the original input dimensions. However, the last decoding layer by itself is ineffective in reconstructing precise object boundaries, resulting in blocky predictions (*193*). FCN 16 and 8 use trainable upsampling and skip connections to integrate multi-scale features from the hierarchy of encoder layers, in order to achieve sharper predictions. The most precise results are obtained using FCN-8, which uses transposed convolution for incremental upscaling and multiple skip connections that leverage the high spatial precision of earlier layers in the encoder by incorporating them in the decoder path. The skip connections used in FCN-8 dramatically improve boundary refinement compared to the other variants. Transposed convolutions, also known as fractionally strided convolutions or sometimes mistakenly referred to as deconvolutions, are essentially convolutions performed in reverse that map features back to their input spatial dimensions (*194*). To better understand the relationship between convolution and its transposed counterpart and the arithmetic of the operations, refer to the works of Dumoulin and Visin (*195*) and Shi et al. (*196*). To fuse the multi-scale features from different stages of the network, FCN converts scores from intermediate layers of the encoder into class probability maps, then adds them to the equivalent layers from the decoder path.

In a study by, Maggiori et al. (*6*) it was found that Fully Convolutional Networks (FCNs) outperformed conventional Convolutional Neural Networks (CNNs) with patch-wise training strategy for RS imagery due to their robustness to small translations and rotations. FCNs have been successfully employed in various applications involving different image types. For instance, Bittner et al. (*193*) used FCN-8 with normalized DSM data to extract building footprints from urban areas, and post-processed the results using Conditional Random Fields (CRF) to refine boundaries. Moreover, FCNs have been applied to extract road networks from both very high-resolution (VHR) optical imagery (*197*) and Synthetic Aperture Radar (SAR) imagery (*198*), demonstrating promising results. Fu et al. (*199*) modified the FCN architecture by adding skip-connections and atrous convolutions to enhance multi-scale learning and reduce the downsampling factor to 8X in the encoder, and used this enhanced network to map complex urban scenes from GF-2 satellite imagery. Yang et al. (*200*) used FCN-8 and a modified version with an extra skip-connection they called FCN-4 to map building footprints on the continental US. They found that both FCN-8 and FCN-4 tended to under-segment and round the edges of buildings in dense urban areas, and that class probabilities from early encoder layers, such as pool1 and pool2, were noisy due to their limited ability to encode semantics. For multi-scale fusion, FCN-8 only used layers from pool3 onward. Despite the success of FCN, the model still suffers from inconsistencies over large regions and lacks spatial details, resulting in rough edges and missed small objects (*201*).

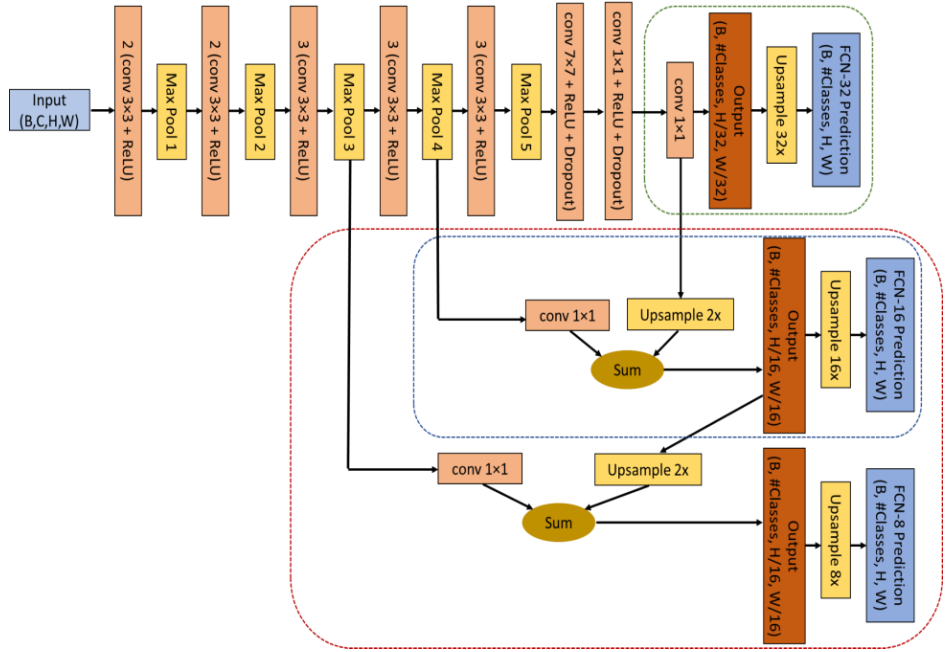

Figure 8. Different versions of FCN, differ in the direct upsampling and how they incorporate multi-scale features in producing prediction output. Decoders for different FCN variants are shown with colored dashed boxes where the green belongs to FCN-32, Blue is representing FCN-16 and FCN-8 is in red.

### 4.2 Encoder-Decoder designs

The encoder path of a convolutional neural network produces feature maps with varying receptive fields, resulting in features with different scales. Early layers in the encoder path provide rich spatial information but are poor in semantic cues, whereas deeper layers provide improved semantic information at the expense of spatial details (*177*). Encoder/decoder networks are designed to exploit this complementary characteristic of the multi-scale information produced in each stage of the encoder path, and use it in the decoder part to refine localization and generate predictions with sharper boundaries, while preserving the semantic consistency for objects of different sizes. These designs are usually implemented through a gradual upsampling procedure with extensive use of lateral skip connections between the encoder and decoder sub-networks to fuse multi-level features, typically using concatenation or summation operations (*202*, *203*).

Unet (*204*) is a commonly used model that utilizes symmetric encoder and decoder sub-networks to progressively refine the prediction. Unet was originally developed for medical imaging using VGG16 as the encoder, but other backbones are also commonly used (*205*). For instance, Poortinga et al. (2021)(*206*) uses a modified Unet where the VGG encoder is replaced with the first 13 blocks of MobileNetV2 to map sugarcane fields in Thailand. Both encoder and decoder sub-networks in Unet have the same number of scale levels (e.g. 5 levels in the original design), each of which are fused by skip connections. Each level in the encoder sub-network consists of a convolution block that doubles the number of feature maps and halves the spatial dimension as the input passes through the encoder path. Each block in the decoder subnetwork upsamples the feature maps by a factor of 2 through bilinear or transposed convolution, but halves the number of feature maps (204). This strategy enables Unet to use a deeper, more complex decoder than FCN, without substantially increasing the number of model parameters. In comparison to FCN, skip connections in the Unet directly use the feature maps from the two paths and use a concatenation operation as the fusion method.

Feature Pyramid Network (FPN) (202) is an encoder-decoder architecture that learns from a single scale input across a hierarchy of scales. Similar to Unet, FPN uses lateral skip connections to fuse multi-scale features. However, FPN augments skip connections with a 1×1 convolution layer and uses summation as the fusion strategy. Each stage of the decoder produces an independent prediction, making it suitable for dense object detection and instance segmentation, especially for small objects (207–209). FPN is also used for semantic segmentation with a simple head that passes the output of each stage of the decoder through two 3×3 convolution layers, upsampling all of them to the proper spatial extent, concatenating the features, then passing them through another 3×3 convolution layer to mitigate artifacts, and maps the features to raw class scores using a 1×1 linear convolution (210, 211). This fusion strategy is also popular for adapting outputs from true multi-branch architectures, such as HRNet, into dense predictions (179).

SegNet (212) is another symmetric encoder-decoder architecture with a similar design to Unet, including the VGG16 backbone, but with a novel upsampling strategy. SegNet (Figure 9) strictly uses max-pooling as the downsampling operator and keeps track of the pooling indices in the sequence of downsampling layers in the encoder path. The equivalent unpool layers in the decoder path use these position indices to first upsample the feature maps to a sparse representation, and then makes them dense through the subsequent convolution block. This method of upsampling is known as "unpooling". Computationally, unpooling is less expensive than transposed convolution and a popular choice for scenarios where preserving spatial information is crucial, such as in medical image segmentation or building multi-branch or ensemble networks (213, 214). DeconvNet (215) is similar to SegNet, but also has an extra 1×1 convolution layer that serves to compress the feature space between the encoder and decoder sub-networks.

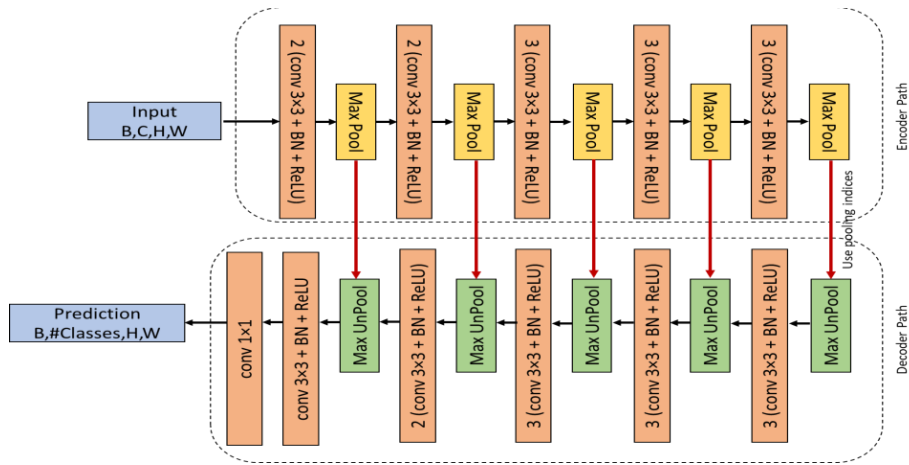

Figure 9. Architecture of SegNet.

Designing an encoder-decoder network involves improving the process of multi-scale feature fusion to overcome the inconsistencies in associating features across different semantic and resolution levels. In practice, this is achieved by introducing semantic information into low-level features and including more spatial information in high-level features. Special modules are designed to facilitate this feature fusion and can be easily integrated into existing encoder-decoder designs. For example, Exfuse (*216*) (Figure 10) introduces three solutions to enrich low-level features with semantic information. The first solution adds more conv layers to the early blocks in the encoder path, but this substantially increases the trainable parameters. The second solution forces the encoder to learn more meaningful semantics by assigning an auxiliary loss to early stages in the encoder and calculating the overall loss as a weighted summation of all loss branches. Using auxiliary loss functions (also known as deep supervision) during the training phase is a common strategy to help the earlier layers in the network have a meaningful gradient during backpropagation (*214*), and is also used for regularization in designs such as Inception (*164*). The third solution is provided by a custom module called a Semantic Embedding Branch (SEB), which guides multi-level feature fusion in a similar manner to Unet but uses element-wise multiplication as the aggregation operation. Exfuse also includes a module called Explicit Channel Resolution Embedding, which is designed to consider more context by dividing the channel dimension into groups and letting the prediction extend to adjacent pixels between the groups. Another special module, Densely Adjacent Prediction, is added to the end of the decoder path, embedding spatial information into high-level layers by using a combination of sub-pixel upsampling, average pooling, and auxiliary loss. The prediction for each output position is then obtained by averaging all these adjacent semantic scores. The Global Convolutional Network (GCN) (*217*), is another encoder-decoder design that uses convolution layers with large kernel sizes to capture more contextual information. Kernels in GCN are factorized in the form of 1×k + k×1 to control the increase in the number of trainable parameters (see Figure 10e). A boundary refinement module, structured as a residual block, is also added to GCN to improve localization accuracy (figure 10d). Scanning convolutional networks (*218*) use the same vertical and horizontal kernels as GCN to make a small-sized network that can be trained quickly

All of the introduced encoder-decoder designs so far follow the standard direction of the decoder, which starts from the coarsest resolution and gradually progresses towards the finest resolution, refining the boundaries of higher-level semantic features. In contrast, Deep Layer Aggregation (*219*) proposes a different approach that aggregates features from the finest to the coarsest resolution which refines the prediction output by iteratively adding more semantics into fine resolution features.

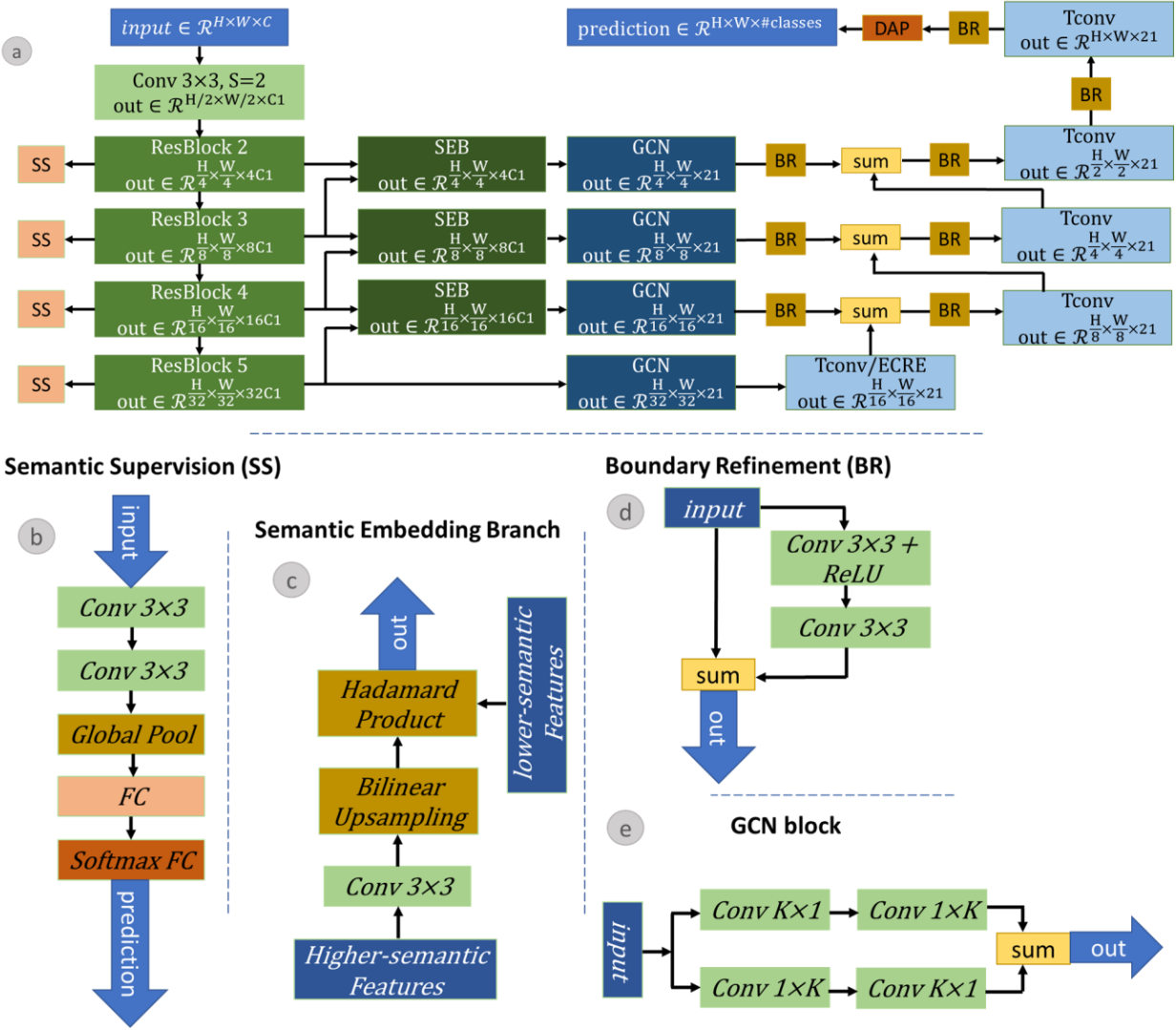

Figure 10. a) structure of Exfuse network. b) Semantic supervision for each stage of the encoder. c) SEB block for multi-scale fusion. d) boundary refinement module as residual correction. e) Global convolution block with large asymmetric kernels. Notice the similarities with figure 15. Tconv is short for transposed convolution.

Exfuse employs a residual unit to refine the boundaries in the fused multi-scale features, which is a common strategy in many encoder-decoder designs. For instance, Chen et al. (220) proposed an error correction module that fuses features from the encoder and decoder by concatenating them and passing them through a residual bottleneck (Figure 4). The residual module learns the semantic differences and produces a more robust feature representation. ScasNet (54) consists of three main parts: 1) an encoder to extract features; 2) a context aggregator module that sequentially aggregates features from the global to the local scale through pairwise aggregation of convolution branches ordered from largest to smallest dilation rate; and 3) a decoder that gradually upsamples the encoder's output to the original image extent using skip connections. Skip connections between encoder and decoder blocks, as well as the connections in the self-cascaded multi-context aggregation module are heavily augmented with inverted residual correction modules. The inverted residual block was introduced in MobileNetV2 (*169*) as a computationally efficient replacement for the standard bottleneck residual unit. The main difference is that, unlike the standard bottleneck unit, the inverted residual branch has more features than the input and output of the unit. In practice, the inverted residual branch is implemented using an expanding 1×1 convolution, followed by a depthwise separable 3×3 convolution, and then a contracting 1×1 convolution. This structure allows for efficient feature extraction and reduces the number of computations needed. RefineNet (*202*) proposes a strategy to efficiently fuse multi-level information in an encoder-decoder network by adding a flexible number of refinement modules in the decoder path. This network uses ResNet as the backbone and the decoding head receives features generated by different levels of the encoder as input. A RefineNet module consists of three processing steps: 1) input from the encoder and decoder paths goes through separate branches of residual convolution, 2) these multi-resolution branches are fused using summation, and 3) the fused features pass through a chained residual pooling block to capture the long-range spatial dependencies between image regions.

There are also designs that try to break free of the restriction of having skip connections limited to the same levels between the encoder and decoder paths. Unet++ (*221*) is an example of such a design that provides an adaptive network depth. The architecture consists of an ensemble of UNets with different depths acting as a unified single network (Figure 11). The encoder is partially shared among the UNets in the ensemble to enable knowledge sharing, but each network has its own decoder. Layers of the same level in both the shared encoder and dedicated decoders are densely connected through skip connections. This dense connectivity provides a pathway for the gradient to flow from the deeper subnetworks in the ensemble to their shallower counterparts, facilitating convergence by gradual fusion of multi-level feature maps between the encoder and decoder layers of the deeper UNets in the ensemble. The authors also proposed optional deep supervision for each network in the ensemble, which enables model pruning by taking an average loss of all ensemble networks or by choosing the results from only one of the networks at inference time. Supervision of intermediate layers, also reported by Mohammadi et al. (*222*) to significantly enhance performance for crop mapping tasks, particularly when models are evaluated across diverse regions. Unet 3+ (*223*) goes a step further and introduces a design with dense skip connections between the stages in the encoder and decoder paths of the original UNet, as well as dense connections between different levels of the decoder, aiming to further improve the process of multi-level feature fusion.

1x1 convolution with softmax activation is commonly used as the classifier on top of the decoder which analyzes each position on a feature map for dense predictions, using a global approach where kernels learned from training samples are fixed and universally applied. Yu et al (*224*) argues that this global approach struggles to account for intra-class distinctions, leading to misclassification of pixels within the same category that exhibit different appearances. The proposed conditional classifier which can be integrated into any FCN-like architecture is a two-part solution designed to overcome this issue: first, the class-feature aggregation module uses weighted averages to identify the distinctions of the same category within one sample, and second, the kernel generation module creates sample-specific kernels based on these distinctions. These kernels are then applied to the input sample to predict the semantic masks more effectively.

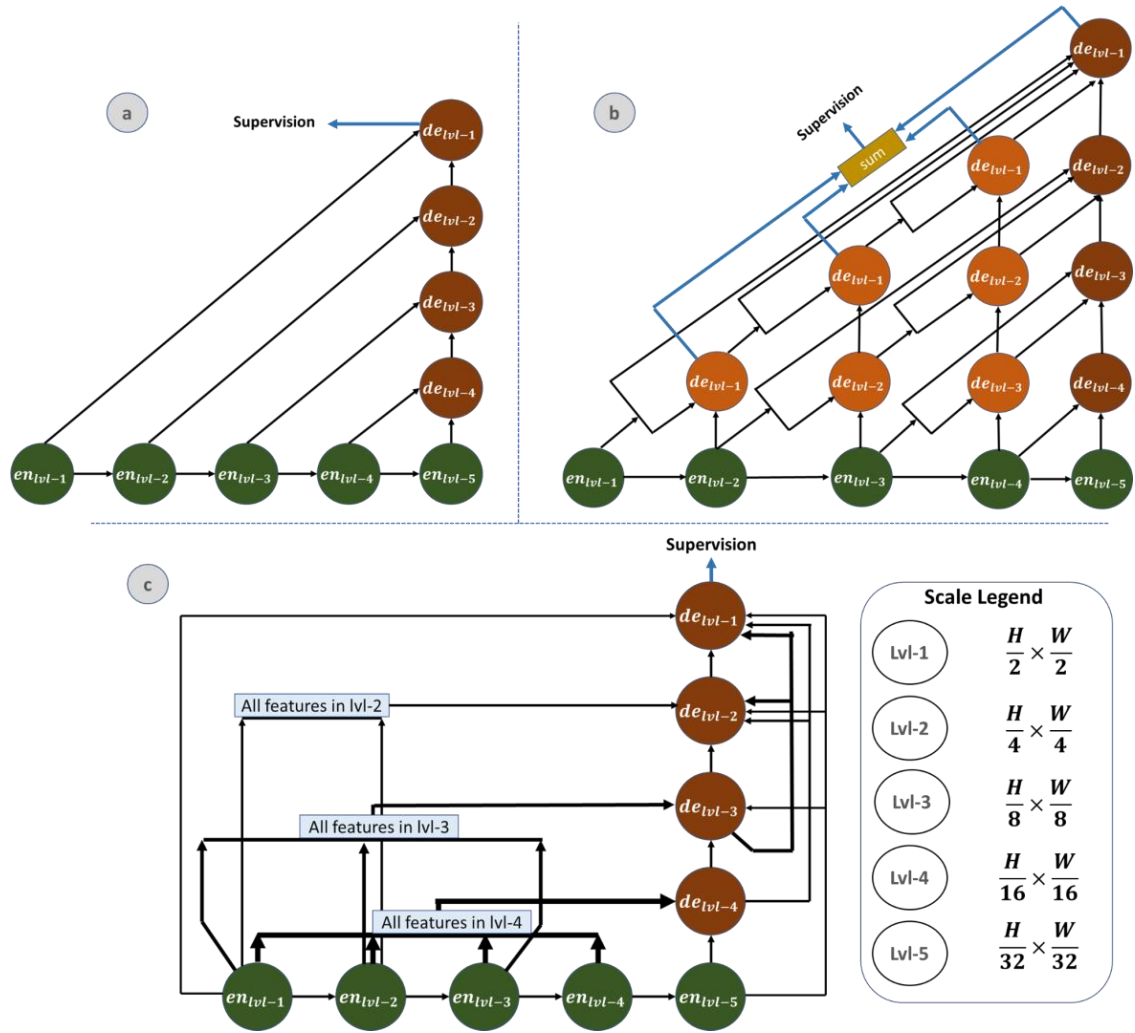

Figure 11. Skip connections in a) Unet b) Unet++ and c) Unet3+. In all the three designs each node represents the feature output of a different scale level where the horizontal nodes (green nodes) belong to the encoder and vertical nodes (shades of brown) belong to the decoder path.

The importance of the encoder versus the decoder in semantic segmentation is a topic of ongoing research. While some studies, like Chen et al. (225), consider the decoding process more challenging and propose a modified Unet model with a light encoder and a complex decoder, others, such as ENet (226), argue that the encoder's role is crucial, and the decoder's job is only to fine-tune the upsampled output. Chen et al. (225) found that their modified Unet performed better for extracting road networks from high-resolution satellite imagery. On the other hand, ENet focuses on optimizing the computational efficiency of the encoder to speed up the inference process while maintaining good accuracy. Ultimately, the relative importance of the encoder and decoder may depend on the specific task and dataset and requires further investigation.

## 4.3 Designs based on pyramid strategies and dilated convolution

As mentioned previously, a CNN requires spatial contextual information to make a prediction for each output pixel. In most applications, a large amount of spatial context must be taken into account in order to successfully label a pixel, thus the total RF of the last layer provides a measure of how much contextual information is used to make the prediction for each pixel in the output. Downsampling through a pooling operation or strided convolutions are common techniques to increase the RF in CNN networks but impose a cost in terms of lost spatial detail. A common shortcoming of FCN and many encoder-decoder designs, is their inability to sufficiently incorporate global contextual information, primarily because spatial support from the conv kernel is fixed and usually small in size, which can lead to inconsistent object segmentation and caused a significant portion of misclassification errors (*47, 227, 228*).

To address this limitation, modules based on pyramid strategies (e.g. multi-scale branches) were developed to encode long range spatial dependencies. One of these modules is the spatial pyramid pooling (SPP), which consists of a set of pooling layers that divide the input image into a fixed number of spatial bins, resulting in a constant output size (see Figure 12). SPP was first introduced in He et al. (*229*), as a means of enabling CNN models with fully connected layers to handle inputs of varying sizes during training, where the SPP module was placed before the first fully connected layer. This approach improves the model's performance by increasing its scale-invariance, reducing its sensitivity to object deformations, and making it less prone to overfitting. Inspired by SPP, Zhao et al. (*230*) proposed PSP-Net, which uses SPP to capture both global and regional relationships for the purpose of improving dense prediction. Pyramid sampling modules like SPP and atrous spatial pyramid pooling (ASPP) are often incorporated at the top of the encoder in semantic segmentation models. Their output can either be directly upsampled to the original input size or further refined by additional decoder layers. These modules have been shown to effectively encode long-range spatial dependencies, and have been applied successfully in semantic segmentation of very high-resolution remote sensing imagery (*231, 232*).

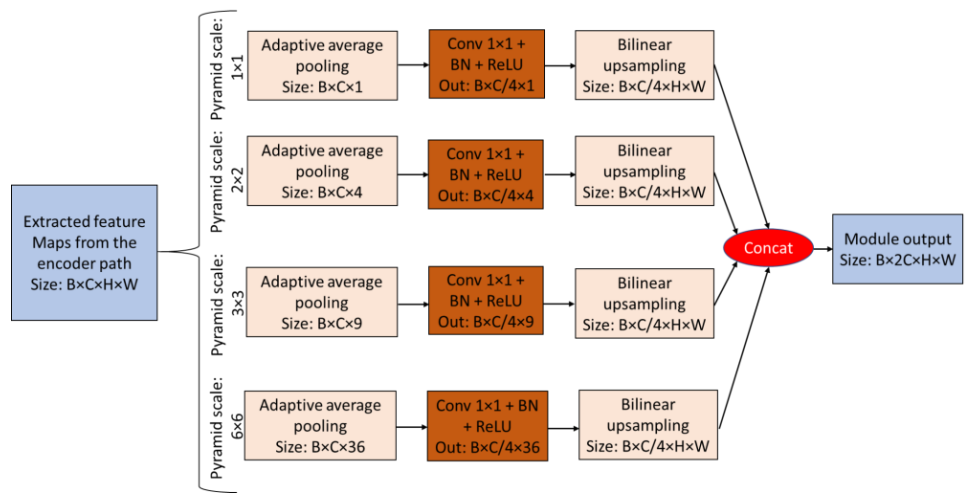

Figure 12. Overview of the pyramid parsing module. Each branch represents different sub-regions, and is followed by upsampling and concatenation layers to form the final feature representation, which carries both local and global context information. source: (Zhao et al., 2017)

Dilated convolution, also known as atrous convolution, is a technique that increases the receptive field of a layer non-linearly without relying on downsampling operations, which can lead to loss of spatial detail (233). This is achieved by introducing a new dilation rate parameter (r), which expands the effective kernel size by inserting (r-1) zeros between each element of a standard f×f kernel. As a result, the receptive field of the layer is enlarged to f + (r -1) (f - 1) × f + (r -1) (f - 1). The default and minimum value of dilation rate is 1. The maximum dilation rate that can be used depends on the spatial dimensions of the input feature map, and for a square input of size m×m, the upper bound is given by $\lfloor\frac{m-f}{2}\rfloor + 1$ (234). Dilated convolution allows the network to incorporate larger spatial context while minimizing the number of extra parameters as only a fraction of pixels are actually sampled from available pixels within the RF, but it can also produce artifacts and checkboarding effects due to the structured sampling of pixels within the enlarged receptive field.

A common design strategy for maintaining high-resolution feature maps during the feature extraction process is the sequential arrangement of dilated convolutional layers in the encoder. If dilated convolutions are arranged serially and the dilation rate is fixed among consecutive layers, the resolution will be preserved, thus preserving local details important for semantic segmentation, and the receptive field will increase linearly (Figure 13b). However, a series of dilated convolutions with increasing dilation rate will maintain the resolution while exponentially increasing the receptive field (Figure 13c) (90, 95). The DeepLab family of networks (235–237) also use dilated convolutions in the encoder to reduce the downsampling factor of the original input, in order to make the decoder's task easier. To speed up the computational time and avoid the memory footprint associated with using dilated conv layers in deep encoders like ResNet-101, FastFCN (238) uses an alternative approach called *joint pyramid upsampling (JPU)*. JPU uses outputs from the last three blocks of a standard five-level downsampling encoder (Figure 13a), and approximates the output feature maps from an encoder with dilation layers (Figure 13b) that maintains the feature resolution at three levels of downsampling. Joint upsampling (Figure 13d) uses high-resolution features to embed detailed spatial structure into low-resolution target features. Onim et al. (239) used FastFCN for land cover mapping of Gaofen-2 Image Dataset (GID-2) and reported class-specific accuracy metrics of above 90%.

Dilated convolution layers are also widely used to capture and fuse multi-scale features when arranged in parallel. The atrous spatial pyramid pooling (ASPP) introduced by DeepLab is an example of this, which aggregates contextual information to produce smooth predictions. The prediction is then upsampled directly to the input extent using simple bilinear interpolation and refined through post-processing. In DeepLabV1 and V2, conditional random fields (CRF) are used for this refinement, but the step is dropped from later versions such as DeepLabV3 (236) and V3+ (240). The latest ASPP module introduced in DeepLab V3 and V3+ consists of multiple branches. These include a branch with a 1×1 conv layer with unit dilation rate, three 3×3 depthwise separable convolution branches with dilation rates >1, where the chosen rate depends on the downscaling factor of the encoder, and a branch with global average pooling to capture image-level clues. Experimental results have shown that global pooling is a better descriptor than using convolutions with very high dilation rates, which can cause edge effects or reduce to a 1×1 conv when the dilated kernel is close to the spatial extent of the input feature map (236). DeepLab V3+ adopts an encoder-decoder design by fusing the ASPP output and an early encoder layer using simple skip connections (*240*). It has been observed that the performance of both SPP and ASPP modules is affected by the size of the mini-batch used in training, with bigger batch sizes resulting in better performance (*241*).

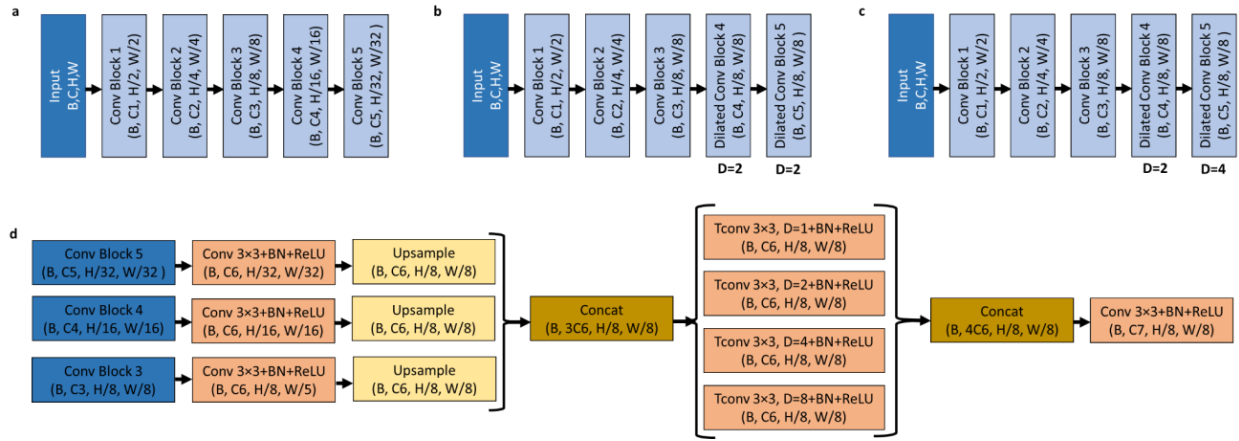

Figure 13. Different arrangement of dilated conv layers. a) A normal encoder consisting of convolution layers with the dilation of 1 and with 5 blocks and 5 levels of downsampling. b) Replaces the conv layers in the last two conv blocks with dilated ones to maintain the resolution at 8× reduction of the input extent (block 3 of network a) but increases the RF roughly similar to conv block 4 in network a. c) Uses dilation blocks of increasing rate in the encoder, which maintains the resolution at 8x but increases the RF of the output comparable to the RF of block 5 in network a. d) Shows the structure of JPU module that takes output of blocks 3, 4 and 5 of the encoder to perform a guided upsampling that exponentially expands the ERF without the computational cost of using dilated conv layers in the encoder path and produces a 4x upsampled output with ⅛ of the original input resolution. Tconv is short for transposed convolution.

DeepLab and its variants have been tested widely for RS applications. For instance, Chen et al. (*242*) modified DeepLab by adding batch normalization and incorporating a CRF as an RNN to monitor sea surface oil spills from SAR and VHR optical satellite imagery. Segal-Rozenhaimer et al. (*243*) used DeepLab to predict cloud and cloud shadow from only RGB and NIR bands of Sentinel-2 and WV-2 images, achieving comparable or better performance than conventional rule-based methods like F-mask. Guo et al. (*244*) demonstrated the effectiveness of the ASPP module in pixel-wise classification of high-resolution aerial imagery, and suggested using CRF-based post-processing to improve object boundary delineation in the output prediction. Multi-scale context aggregation is also utilized in other encoder-decoder designs. For instance, Liu et al. (*245*) integrated a ResNeXt-50 backbone with a squeeze-and-excitation attention module and an ASPP module in a custom U-Net architecture to identify and grade Maize crop type drought levels from UAV overhead imagery.

Several design strategies have been proposed to improve the pyramidal arrangement of dilated convolutions in the ASSP module. One such example is the Efficient Spatial Pyramid (ESP) module introduced by Mehta et al. (*246*) that reduces the computational cost of the ASSP module using convolution factorization and adoption of a reduce-split-transform-merge strategy to make the module suitable for use in resource-constrained edge devices (Figure 14a). ESP blocks are used in different stages of the encoder to learn scale-invariant representations and increase the receptive field (RF) of the convolution layer. ESPNet-V2 *(247)* improved both the computational cost and performance of ESPNet using a more efficient convolution factorization based upon group point-wise convolutions and depth-wise dilated separable convolutions. The Dense ASPP module *(248)* offers a different strategy to improve the RF of the ASSP module by serially arranging the dilated conv layers from the smallest to the largest dilation rate (Figure 14b). Serial arrangement achieves a much higher RF compared to the parallel arrangement of these layers in the original ASPP. Furthermore, the addition of dense connections between the cascaded layers ensures that the module learns a high diversity of scales and a denser sampling, as more pixels are involved in the convolution process. Another context aggregation module introduced by Wang et al. *(241)* is based on a sequence of three deformable convolutions with superior performance compared to ASPP and SPP on several benchmark datasets. Deformable convolution has more freedom in changing its receptive field thus can facilitate capturing objects of varied shapes and scales *(249)*, which may be harder to capture with standard convolutions, since these are applied to fixed locations within the input feature map and can thus lead to misalignment between the kernel and object features. A deformable convolutional layer *(97)* also has a fixed kernel size, but it allows the receptive field to be shifted and deformed within a local neighborhood, enabling it to better capture object features that are not aligned with the grid. Deformable convolution achieves this by introducing an additional set of learnable parameters called offsets, which determine the spatial shift of the kernel at each location in the input feature map. In addition, deformable convolution also includes an additional set of learnable parameters called modulation coefficients, which control the weighting of the input feature map at each location, allowing the model to better capture object features with varying scales.

Dilated convolutions can lead to grid-like patterns and artifacts (i.e. checkboarding effects) in the output feature maps, especially when they are used in a cascade with increasing dilation rates *(250)*. These artifacts are caused by the fact that the effective receptive fields (ERFs) of adjacent units in the output layer are independent of each other, and as the number of layers in the cascade increases, the number of neighboring units with non-overlapping receptive fields also grows larger. This problem can be exacerbated by using increasing dilation rates by a common factor in the cascaded layers, as it ensures that the ERF of neighboring units in successive activation maps never overlap. The idea of blind spots introduced through non-overlapping ERF in consecutive dilated convolution layers is illustrated in Figure 15. This problem can be intensified by using large dilation rates, where the distance between non-zero elements in the filter becomes too large, resulting in inconsistent feature maps. Additionally, when multiple dilated convolutions with the same rate are used in a cascade, a pixel in the top layer may only be affected by a small fraction of the pixels in its ERF from earlier layers, leading to a loss of local information and the appearance of a checkerboard pattern in the output *(251)*.

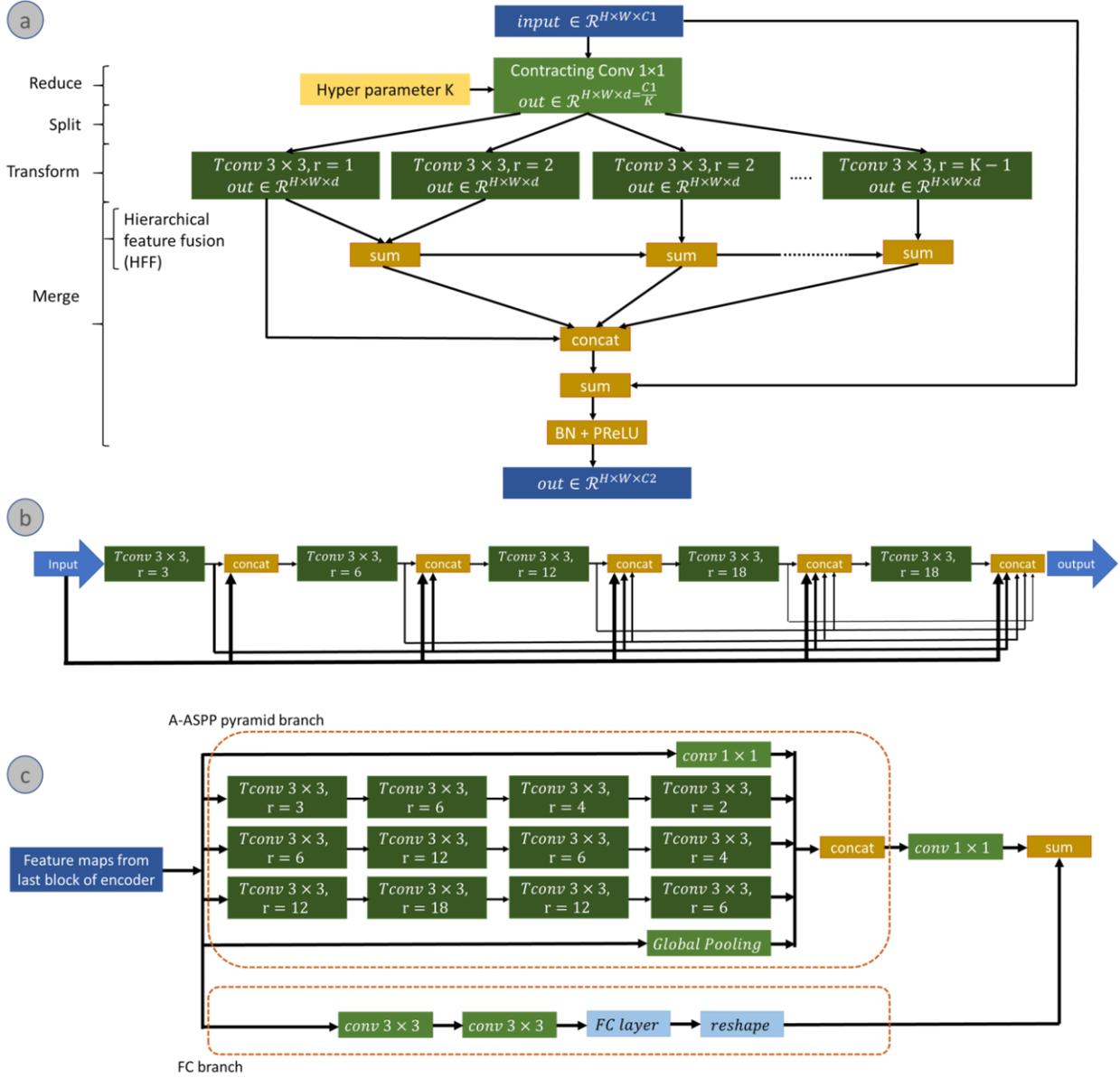

Figure 14. a) Illustration of the Efficient Spatial Pyramiding (ESP) block. First a 1×1 conv layer compresses the input over the channel dimension based on a compression factor (K) to produce $d = \frac{C}{K}$ feature maps, where C is the desired number of feature maps in the ESP module output. This reduced representation is then transformed through K independent branches of n×n dilated convolutions with different dilation rates of $2^{(k-1)}$ and different effective kernel size of $[(n-1)2^{k-1} + 1]^2$, where k increases with the branch index number from 1 to k. The output of all branches are then concatenated and summed up with the module's input to produce the ESP output. b) Dense ASPP module. Notice how the pairwise fusion strategy in HFF extends to dense connectivity. c) Augmented Atrous Spatial Pyramid Pooling (A-ASPP) module. Tconv is short for transposed convolution.

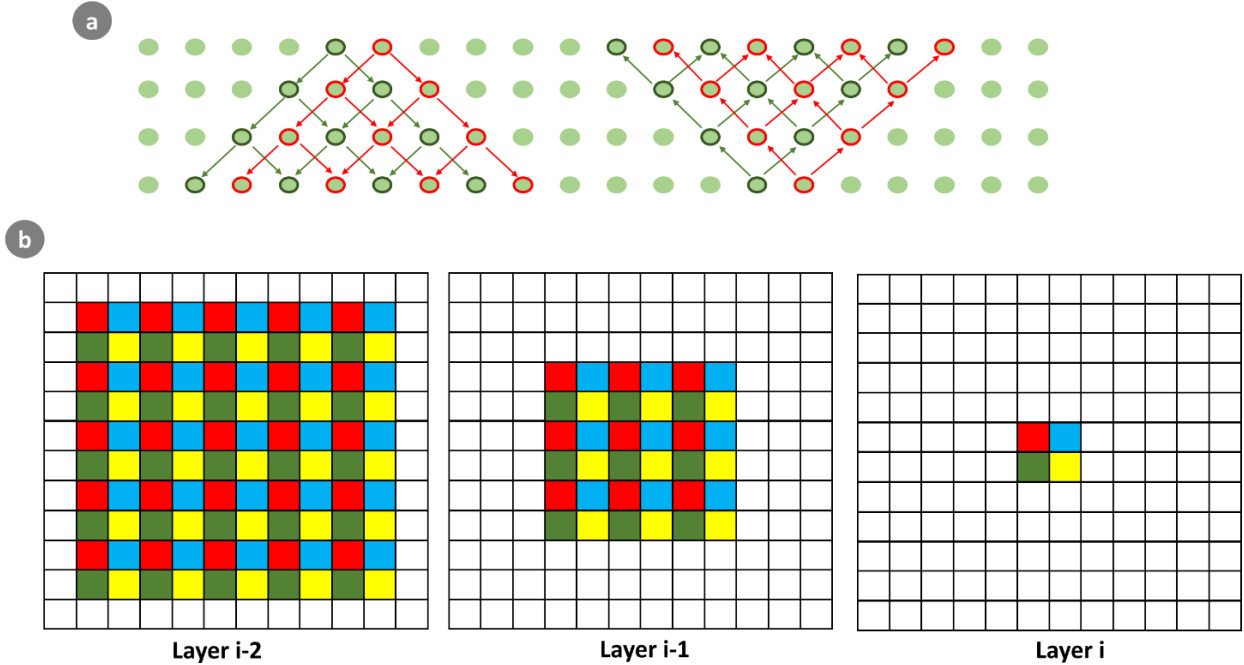

Figure 15. (a) Illustration of non-overlapping fields of view of two adjacent units in a 1D conv on the top-most network layer (left), and the non-overlapping areas of influence of the two adjacent units on the bottom-most network layer (right). (source: (56))  (b) An illustration of gridding artifacts in a 2D convolution. The operations between layers are dilated convolution with a kernel size of 3×3 and a dilation rate of r = 2. For four neighboring units in the layer i indicated by different colors, we mark their actual receptive fields in layer i - 1 and i - 2 using the same color, respectively. Clearly, their actual receptive fields are completely separate sets of units. (source: (250))

ESP (246, 247) proposes a solution to the gridding issue when layers have parallel arrangement, which involves adding the feature maps from a branch with a lower dilation rate to the adjacent branch that has a higher dilation rate (e.g., the merge strategy in Figure 14a). Hybrid Dilated Convolution (HDC) (251) is another approach to address the gridding issue in dilated convolutions. HDC divides dilated convolutions into groups of typically three and assigns different dilation rates to the convolutions in each group. There are two restrictions in HDC: 1) the maximum distance between two non-zero values for the second dilation layer in the group must be smaller than the layer's convolution kernel, and 2) the dilation rates in each group's convolutions cannot have a common factor. A-ASPP module (augmented-ASPP) (252), uses a sequence of increasing and decreasing dilation rates in each of its 3x3 dilated convolution branches, resulting in denser sampling and minimizing the gridding issue. Additionally, A-ASPP incorporates a fully connected (FC) fusion branch that captures and integrates global context information into the prediction output, as shown in Figure 14c. The authors reported that these enhancements lead to substantial improvements over DeepLab V3 in segmenting small objects.

The Large Kernel Pyramid Pooling module (LKPP) (253) aims to detect objects at different scales by fusing local and long-range dependencies. The LKPP module (Figure 16a), which sits between the encoder and decoder sub-networks, is designed to mitigate both the gridding issue and under-segmentation issue where large objects are fragmented into multiple parts. The LKPP module consists of three Hybrid Asymmetric Dilated Convolution (HADC) blocks that come in two configurations: parallel LKPP (Figure 16b) is preferred for indoor scenes with objects that tend to be irregular in shape, while cascaded LKPP (Figure 16c) with a more rapid and extensive expansion of the model's receptive field is recommended for detecting objects of regular size in outdoor scenery. A subsequent modification of the parallel HADC block (Figure 16d) replaced the paired layers with two asymmetric dilated convolution (ADC) layers, with complementary kernels of $k_1 \times k_2$ and $k_2 \times k_1$ arranged in parallel where $k_1 \neq k_2$ and $min(k_1, k_2) > 2$ (88). This modification yields cross-shaped receptive fields that help capture more context while discarding redundant information. The gridding issue in this revised configuration is minimized, by setting the dilation rate in the first sub-block for each kernel direction to 1. The dilation rates of the other two sub-blocks are chosen such that the maximum absolute rate difference of block 3 from block 2 or the rate in block 1 must be equal or smaller than the kernel size in both vertical and horizontal kernels (i.e. $max(|r_n^3 - 2r_n^2|, r_n^1) \leq k_n \ \forall n \in \{1,2\}$) where the superscript represents the sub-block and subscript represents the kernel direction).

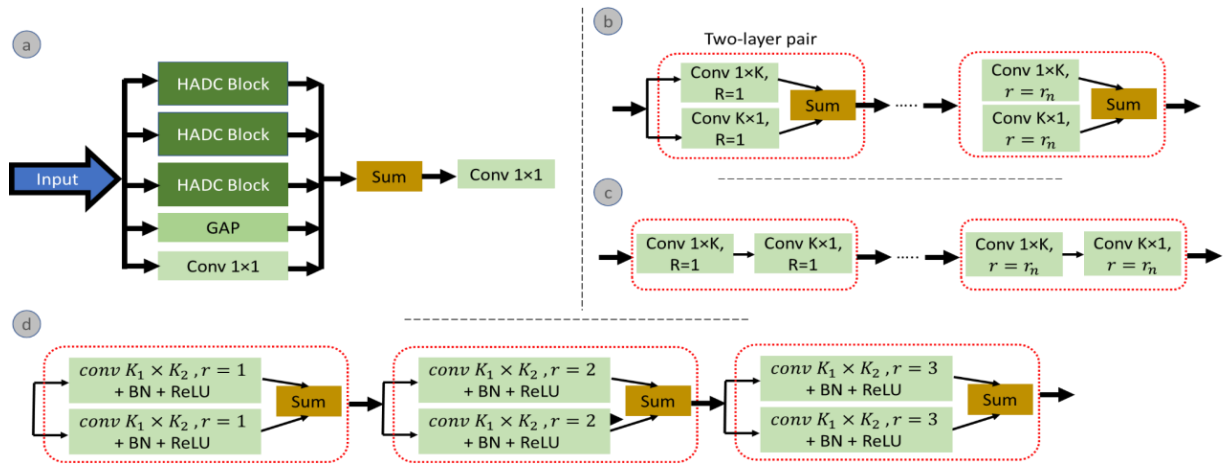

Figure 16. a) The Large Kernel Pyramid Pooling module (LKPP). The module is a multi-branch design that consists of a 1×1 conv, Global Average Pooling, and three Hybrid Asymmetric Dilated Convolution (HADC) blocks. Each HADC block consists of three sub-blocks containing a dilated conv decomposed into 1×k and k×1 layer pairs, with either cascaded or parallel arrangement where each sub-block has a different dilation rate. b) The HADC block in parallel LKPP. C) The HADC block in cascaded LKPP d) modified parallel HADC with complementary kernels. Having a cascade of multiple layer pairs has the effect of densely enlarging the receptive field.

Wang and Ji (250) proposed a novel approach by decomposing a dilated convolution layer into three stages (Figure 17). Firstly, a periodic subsampling with a factor equal to the desired dilation rate (r) is applied to the 3D input of size $H \times W \times C$, resulting in $Cr^d$ groups of feature maps with coarser resolution of $\frac{H}{r} \times \frac{W}{r}$, where d is the number of spatial dimensions (usually 2 or 3 for image inputs). Secondly, a shared-weights valid standard convolution is performed on each output feature map from the first stage, using the weights of the original dilated convolution after removing the inserted zeros. Finally, the resulting $r^d$ feature maps are interlaced to generate the final output, which is the same size as the output produced by the valid dilated convolution with zero-insertion but would not be identical. The decomposition method clarifies that the lack of inter-group dependency between the $r^d$ feature maps in the first stage and the second stage can cause each pixel in the intermediate layers to be separated from its neighboring pixels in the other d-1 groups. To overcome this issue, Wang and Ji (*250*) suggest adding dependencies either before the first stage of decomposition, by using a shared separable convolution before the dilated convolution, or to the output of the second stage, by incorporating a shared separable block-wise fully-connected layer after the dilated convolution to consider dependencies. Hamaguchi et al. (*56*) observed that cascading convolutions with increasing dilation rates can lead to the loss of small objects in dense predictions. To address this issue, they proposed a Local Feature Extraction (LFE) module that can be added to the end of a cascade of dilated convolutions. The LFE module consists of a block of dilated convolutions with decreasing dilation rates, which helps to correlate the ERF of neighboring units and recover local relationships.

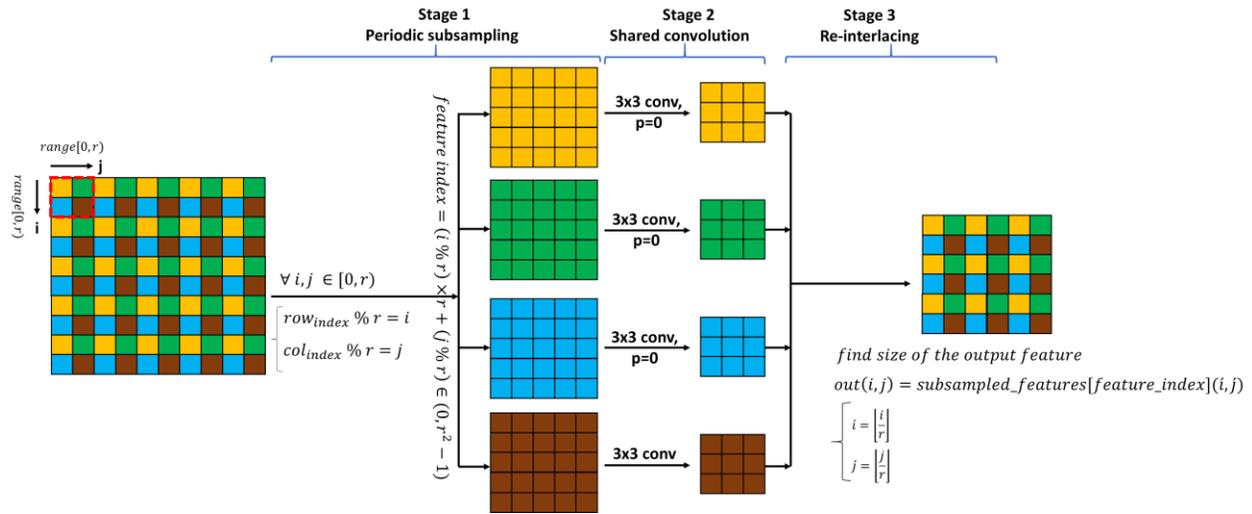

Figure 17. An example of the decomposition of a dilated convolution with a kernel size of 3 × 3 and a dilation rate of r = 2 on a 2-D input feature map. Starting from a 10×10 input, we get 4 groups of reduced features (5×5) as a result of periodic subsampling. At the second stage a shared convolution is applied to these intermediate features to produce four new groups of features (3×3 in size because the shared convolution is valid with no padding). Re-interlacing these features will produce a 6×6. (Source: (*250*))

**4.4. Comparison of different CNN architectures for semantic segmentation in remote sensing**

CNNs have shown remarkable success in semantic segmentation of remote sensing imagery and their usage is growing rapidly. The state-of-the-art methods in this field employ various architectures, ranging from classic models like U-Net, SegNet, and DeepLab to more advanced models such as DenseNet, PSPNet, and OCNet. These models have demonstrated impressive results on various types of remote sensing imagery, including aerial imagery, satellite imagery, and Unmanned Aerial Vehicle (UAV) imagery. One example that brings together and compares several of the strategies we described in this section is described by Singh et al. (*254*), who extensively investigated the performance of four popular dl-based models namely: DeepLab, UNet, SegNet and DenseNet for mapping river ice from VHR aerial imagery. They conclude that all DL models outperform conventional machine learning algorithms, such as SVM for the majority of ice categories, by significant margins. DeepLab achieves the highest quantitative results on the annotated test dataset, but DenseNet shows the best generalization power to unseen data. UNet, on the other hand, provides the best trade-off between generalization and accuracy. In a similar study, Pashaei et al. (*70*) compared the performance of a few common architectures, including SegNet, U-Net, FC-DenseNet, DeepLabV3+, PSPNet, and MobileU-Net, for land cover mapping of coastal wetlands from a small-sized Unmanned Aerial System (UAS) imagery dataset, finding the best performance using FC-DenseNet, followed closely by UNet. Erdem et al. (*255*) used an ensemble of five different types of UNet with majority voting with equal weight for all branches to extract shorelines on a global scale using only the blue, red, and NIR bands of Landsat 8 OLI imagery. Chowdhury et al. (*256*) investigated the performance of ENet, PSPNet and DeepLab V3+ in semantic segmentation of hard-to-identify and confusing disaster-related classes like debris and buildings with different levels of damage from very high resolution aerial imagery, where PSPNet substantially outperformed the other two models. Xia et al. (2021)(*257*) explored the performance of several carefully selected models, including FCN-8, PSPNet, U-Net, SegNet, DeepLabv3, DeepLabv3+, DenseASPP, DANet, and OCNet, for classifying pine trees infected with Pine wilt disease in UAV imagery. Among the models, DeepLabv3+ had the best overall performance, but U-Net achieved the best recall score. FCN-8, DANet and OCNet had more omission errors, while PSPNet, U-Net and SegNet were more prone to commission errors. The authors observed that increasing the depth of the ResNet backbone in these architectures reduced the recall but increased the precision, which they related to overfitting due to excess capacity provided by deeper backbones.

## 4.5 Designs with multiple branches and communication strategies

While we have discussed single-branch networks so far, there are scenarios where multi-branch networks are more suitable. For example, models using multi-task learning, geometry-aware models, or those designed to work with multi-source inputs may require multiple branches. The latter case is particularly important because of the increasing number of Earth observation satellites in orbit (*258*), which carry a range of sensor types, such as passive (e.g., optical, thermal) and active (e.g., synthetic aperture radar (SAR), LiDAR), which collect data at various spatial scales and within different electromagnetic spectra. These complementary data provide valuable information to differentiate semantic objects (*249*). For instance, SAR captures information on the three-dimensional structure of surface features, and is increasingly combined with optical imagery to improve land cover maps, particularly in cloudy tropical regions (*259–261*).

DL models that are designed to handle multi-modal RS data often have multiple branches and typically follow three approaches to fuse these branches. These approaches can be used with both Convolutional Neural Networks (CNNs) and Recurrent Neural Networks (RNNs), as shown in Figure 18. The three approaches are:

1) *Dataset-level fusion*, where the multi-modal data are fused at the dataset level before being fed into the model for training. This strategy assumes that the model learns to extract the useful features from the joint distribution of the multi-modal dataset through a single branch. Langford et al. (*262*) and Elamin and El-Rabbany (*263*) investigated dataset-level fusion for land cover and land use mapping and found that it improved segmentation accuracy. However, multi-modal data can have different physical and numerical properties, such as optical and radar data, or are encoded using a different data structure, as in 2D multi-spectral imagery and 3D Lidar point clouds (*264*).

2) *Early* and *Late fusion* strategies process multi-modal data independently within a specific branch for each data type, combining the branches at least in a single point in the network before the classifier, either in the earlier or later layers. A good example of early fusion is Chen et al. (*265*) using DeepLab v3+ with two separate ResNet-50 encoders, one branch for optical imagery input and the other dedicated to ancillary raster data (DSM/NDSM/NDVI). Encoded representations at the end of the encoder branches are merged through summation and pass through a shared decoder. They also add a weighted cosine similarity term to the CE loss to act both as a regularizer and constrain the model to learn complementary information from the two input branches. An example of the late fusion is Audebert et al. (*213, 266*), who use a design with two separate SegNets to process elevation and multi-spectral imagery separately up to the point where raw score maps at the end of the decoder are generated. Scores from the two branches are concatenated and passed through a residual block as a new third branch and their output is passed to the softmax classifier. Other designs mix early and late fusion, such as the multi-scale fusion strategy used by FuseNet (*267*) or DP-DCN (*188*), which aims to better integrate modalities through more frequent connections between branches. Chen et al. (*268*) introduce a single branch network with early fusion that can imitate a late fusion strategy, using a mixture of group and standard convolution layers to imitate modality fusion at different scale levels. Other designs, such as C3Net (*269*) and (MS)2Net (*270*), use specialized modules with attention mechanisms designed to minimize multi-modal noise between optical and depth (e.g. US3D dataset), and more effectively integrate the complementary features from these different modalities.

3) *Decision-level fusion* fuses probability maps at the prediction level, through a variety of techniques, ranging from averaging (*213*), summation (*271*) and voting, to more complex fusion modules that use residual corrections, attention mechanisms, or CRFs (*272*). Decision-level fusion is most beneficial when each branch makes separate high confidence predictions. If the confidence level of predictions from one branch is much lower, then earlier, within-network fusion might be more suitable, as it can incorporate weakly ancillary data into a joint-learning pipeline (*213*). A good discussion of decision-level fusion is found in Ge et al. (*273*), where a weighted linear combination of prediction probability vectors from different sources is used to reconstruct the prediction probabilities in the final output.

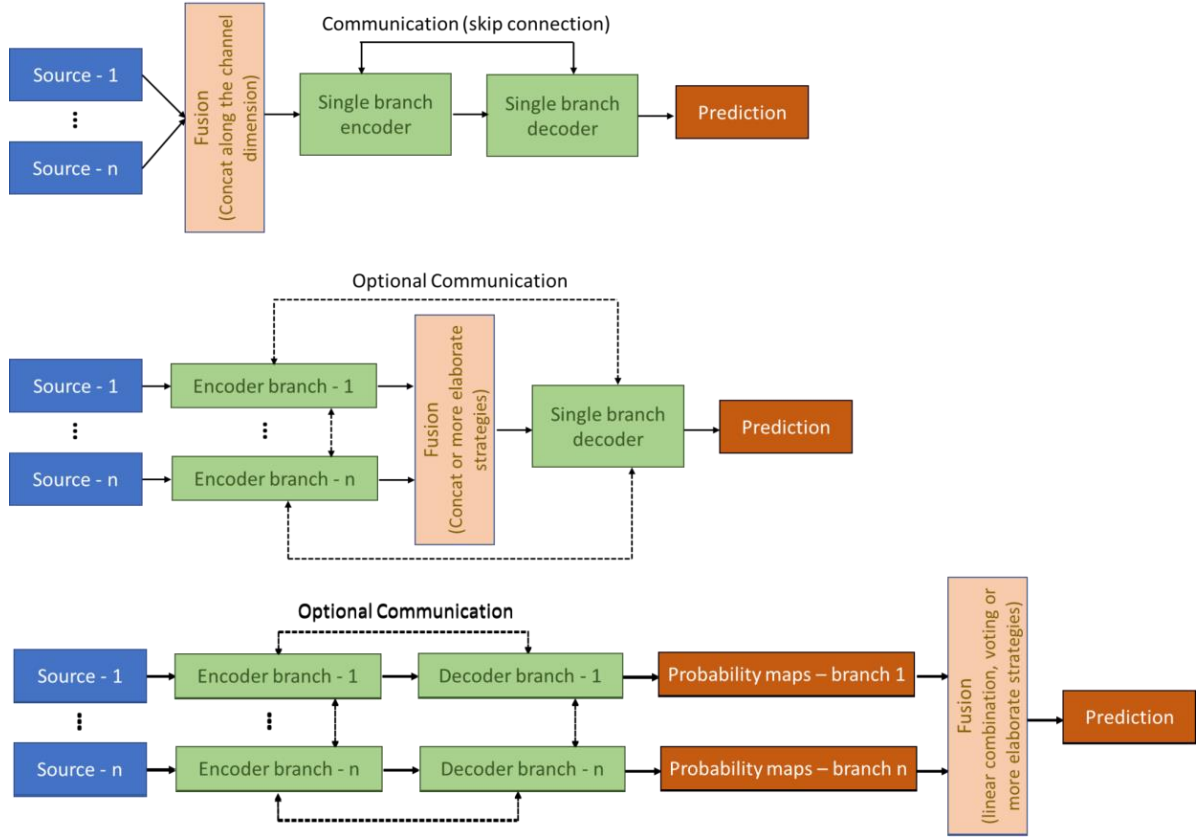

Figure 18. Generic structure of different fusion strategies to handle multi-modal input. From top to bottom: dataset-level fusion, late fusion and decision-level fusion.

### 4.6 Boundary-aware Models

Delineating boundaries between objects of different classes is crucial for successful semantic segmentation. However, two challenges make boundary delineation complicated. The first challenge is related to separating objects despite inter-class similarity (i.e. adjacent objects of different classes with similar spatial or spectral characteristics) and the second is related to intra-class dissimilarity (i.e. adjacent areas of the same class with different spectral characteristics) (*274*). These challenges are magnified with VHR imagery, where large compound objects such as rooftops may be broken into multiple objects of different classes, exacerbated by the limited number of broad bands available, resulting in an increasing chance of spectral confusion. Intra-class dissimilarity typically arises from limitations imposed by the model architecture, such as downsampling operations to increase the model's receptive field or model's inability to learn efficiently at multiple scales. Common techniques to deal with this second challenge include multi-scale feature fusion methods (e.g. Exfuse (*216*) and receptive field enlargement techniques such as using larger kernels (e.g. GCN (217) or the multi-receptive field module (*275*). In Chen et al. (*276*), a boundary refinement module was developed to improve the delineation of urban green spaces in Sentinel-2 imagery. This module consisted of a residual block with three dilated convolutional layers of different kernel sizes arranged serially.

Although these techniques are mostly successful in mitigating intra-class inconsistency, most misclassified pixels in popular DL models (e.g. DeepLabV3, HRNet, and Gated-SCNN *(277)* are located within 5 pixels of object boundaries *(275)*. These errors arise because the model architectures do not consider the geometric prior relationships between the interior and boundaries of objects *(278)*.

Model architectures for refining boundaries can be classified into two main categories. The first category incorporates relational dependencies either during end-to-end training or through a second stage post-processing step, such as using Conditional Random Fields (CRF) *(235, 279–281)*, Markovian Random Field *(282)*, or attention mechanisms *(233)*. The second group, which is the focus of this section, consists of models that explicitly incorporate geometric information during the training process to address inter-class similarities. The model designs in this group differ in terms of 1) how geometric information is incorporated in the learning process (e.g. dual branch vs single branch; single task vs multi-task), 2) how they obtain boundary information (edge detection filters or morphological operations), and 3) formulation of the loss function.

Specific boundary-aware loss functions are required because the Cross Entropy loss, the most commonly used loss function in semantic segmentation, assumes equal importance for all samples regardless of their proximity to object boundaries *(283)*. CE measures the dissimilarity between the predicted probability distribution and the true distribution. In contrast, area-based loss functions like IOU and Dice measure the similarity between the predicted and true segmentation masks in terms of the overlap between their areas. Both IOU and Dice are differentiable surrogates of evaluation metrics with the same names. IoU (also known as Jaccard index) is defined as the intersection of the predicted and ground truth segmentation areas divided by their union (i.e. $\frac{TP}{TP+FP+FN}$). Dice loss is defined as twice the intersection of the predicted and ground truth segmentation areas divided by the sum of their areas (i.e. $\frac{2TP}{2TP+FP+FN}$). Both losses take into account the number of true positive (TP), false positive (FP), and false negative (FN) pixels, which allow them to penalize false positives and false negatives more heavily than CE loss. For more information on boundary-aware loss functions, refer to *(284–286)*.

**4.6.1 Single-branch boundary-aware models**

One straightforward approach to crafting single-branch, boundary-aware models is to classify object borders as a distinct category, with the goal of more precisely outlining the separately annotated object interiors. However, a potentially more fruitful strategy for single-branch design involves the use of a compound loss function that introduces an extra complementary term or imposes a weighting scheme based on, for example, each pixel's position in a boundary distance map. By penalizing misclassifications based on the inverse distance to boundaries, this technique can bolster the model's capacity for accurate object boundary demarcation (e.g. *(287)*). Kervadec et al. (2019)*(288)* developed a boundary loss function that calculates the Euclidean distance of semantic contours. When combined with traditional loss functions such as cross entropy or Dice, this boundary loss produces distance-to-boundary information that enhances evaluation metrics, results in smoother loss curves, and improves the representation of small objects. The best results were observed when the boundary loss was assigned low importance weights (e.g. 0.01) in the initial epochs, which were then gradually increased throughout subsequent epochs. NeighborLoss *(289)* utilizes the spatial correlation of pixels in the prediction output by weighting each pixel based on its prediction similarity with its eight neighbors, with weights increasing proportionally with dissimilarity. This approach achieves more consistent results than CE, although its efficacy largely hinges on class balance. Another approach (*290*), used for extracting crop parcels from PolSAR imagery, involves a single branch PSP-Net with added skip connections, attention modules, and a compound loss function that combines CE loss for object interiors with dice loss for boundaries. To extract boundaries for the dice loss, the model applies the Canny operator on the prediction and reference labels. Jin et al. (2021)(*291*) also use a single branch and directly generate boundaries using a Laplacian operator from both the model prediction score and the annotated layer. They also introduce a boundary aware loss function that combines three terms: 1) CE loss for the dense prediction, 2) the Online Hard Example Mining (OHEM) loss for boundaries, and 3) an extra term made up of an auxiliary CE loss from an intermediate representation in the encoder. Edge-Enhanced RefineNet (ERN) (*292*) uses a single-branch encoder-decoder structure based on HSNet, with a compound edge-reinforced loss function to mitigate semantic ambiguity. The loss function linearly combines three weighted CE loss terms: one for the semantic task, and separate edge losses for the encoder and decoder subnetworks. The edge losses are calculated by applying a Sobel filter to the ground truth segmentation masks and the predicted output and then computing the weighted binary cross-entropy loss. Each edge loss has an importance weight that should be orders of magnitude larger than the semantic term weight, which helps to reinforce the edges and better define object boundaries. Edge-attention Network (EaNet) (*88*) adopts an encoder-decoder design equipped with modified LKPP module and a compound loss consisting of CE for dense prediction and Dice-based edge loss, which simultaneously learns the semantic category and edge structure to refine object boundaries.

### 4.6.2. Multi-branch/multi-task boundary-aware models

Multi-branch boundary-aware models typically adopt a multi-task learning (MTL) approach, where the model is trained on multiple interrelated tasks (e.g. identifying object semantics and geometry), which provide complementary information that boosts model performance for each individual task (*75*). Although each task can have equal importance, MTL can be designed as a set of primary and auxiliary tasks. In this case, the auxiliary task, such as boundary detection, can enhance the performance of the primary task, such as classifying the object interior, by regulating (e.g. bounding) the parameter space during optimization (*293*). Depending on the design of the model, MTL tasks can be executed simultaneously and independently or conditionally (*294*). Design of an MTL requires a strategy to fuse branches/tasks and a suitable loss function for each branch/task. For instance, Marmanis et al. (*295*) explicitly encoded class boundaries in an ensemble of multi-branch networks using a multi-stage training strategy. Two SegNet branches classify different inputs (RGB imagery and a DSM), while two Holistically-Nested Edge Detection (HED) (*296*) branches extract boundaries from the same inputs and the final prediction is the result of the decision-level fusion of all four branches. Shen et al. (*297*) built an FCN network with a shared encoder and two separate decoders for boundary and region tasks respectively. They applied a late fusion strategy that merges branches through concatenation of the latent features before the classifier. They also employed deep supervision for each individual branch/task and the total loss is the sum of each branch's CE loss. The Discriminative Feature Network (DFN) (*274*) also uses a shared encoder (e.g., ResNet) and dual decoder branches, with a different flow direction for each decoder branch. In the segmentation branch, the flow is from bottom (late layers) to top (early layers), gradually infusing details into high-level semantic features. In contrast, the boundary branch moves from top to bottom, adding semantics to low-level features. The semantic branch is followed by CE and the boundary branch with the focal loss, and there is deep supervision on each level of decoder in both branches during training. SegFix (*298*) uses HRNet as a shared encoder with two dedicated decoder branches for boundary and direction. Their data-agnostic boundary refinement module uses an offset map to reclassify predicted boundary pixels to the category of interior pixels. The refined boundary map is then used to further refine the segmentation mask in the direction branch. The direction branch takes the output of the encoder and the refined boundary map and applies a refinement module that uses a direction map to assign each pixel to a specific building. The boundary refinement module is trained on a separate dataset of boundary maps and offset maps, which allows it to generalize to new datasets without the need for re-training or fine-tuning. Bischke et al. (*299*) adapted SegNet for multi-task learning to map building footprints in VHR aerial imagery. Their model branches at the last decoder layer to separately predict the segmentation mask and the distance to building borders using a compound Uncertainty Based Multi-Task Loss function controlled by a hyper-parameter to assign the relative importance of each task. EANet (*300*) propose an encoder-decoder structure with multi-task learning for building region and edge segmentation, where the edge branch is obtained by lateral extension at each level of the decoder, with a BCE loss for each level. Edge labels are generated using a morphological erosion operation on the edge prediction score and corresponding reference annotations. Waldner and Diakogiannis (*301*) used ResUNet for MTL with four tasks: 1) reconstruction of the input image in HSV color space; 2) estimating the distance to the nearest boundary; 3) identifying the field boundaries; 4) classifying agricultural fields in which the order of tasks 2-4 is determined by the preceding task that was used to construct the logits for the next task.

A good example of a multi-branch structure without MTL is BT-RoadNet (*7*), which aims to extract roads from VHR imagery while considering both boundary accuracy and topological connectivity of the extracted roads. The design involves serial arrangement of two UNets, with the first one being a deeper 6-level UNet with residual blocks that generates a coarse road probability map. The second UNet is shallower, with 4 levels and standard convolutional blocks, and it refines the road boundaries by learning the residual between the coarse prediction and the reference layer. BT-RoadNet has a compound loss function that sums Binary CE loss for pixel level prediction, Intersection over Union (IoU) loss for image-level dissimilarities, and Structural Similarity Index Metric (SSIM) to assess the local structural similarity of roads.

### 4.7. Attention Mechanisms

Attention mechanisms were inspired by the ability of the biological brain to selectively focus on the parts of an input signal that are most relevant for a recognition task (*302*). Originally used in Natural Language Processing (NLP) to handle relationships in sequential input (*303*, *304*), attention mechanisms have been adopted in computer vision to identify long-range relationships between features in imagery while filtering out irrelevant information (*305*, *306*). There are two main branches of attention: hard and soft. Hard attention is deterministic and not differentiable, and needs complex procedures like Monte Carlo

Sampling (*307*) or reinforcement learning for training (*308*). Soft attention is probabilistic and differentiable, and can be trained with a backpropagation algorithm, making it easier to integrate it into existing semantic segmentation models. Here we focus only on soft attention approaches.

Soft attention methods can be categorized based on the nature of the inputs to the attention module. When there is only one input, the mechanism is referred to as self-attention, and dependencies are established among the elements (e.g., pixels) within the same source. In contrast, co-attention applies when there are two or more inputs from different sources, and long-range dependencies are obtained between any two pairs of elements from different sources. Attention mechanisms can also be classified based on the dimension of the input to which the attention is applied, which can include spectral (what features?), spatial (where in 2D space?), or temporal (when in image temporal cubes or geo-tagged video frames?) dimensions, or arbitrarily defined dimensions such as scale (*309*).

Attention modules are used in various positions within CNN or RNN networks, and multiple attention modules can be utilized, each focusing on different dimensions of the input or different levels of the network. The integration position of these modules depends on the intended use of the mechanism, the attention dimension, and the computational overhead of the module. Self-attention mechanisms are commonly used in semantic segmentation to capture long-range dependencies and incorporate global context into the encoder-decoder structure, making the top layer of the decoder before the classifier, or the end of the encoder ideal positions to attach these modules (*310, 311*). Co-attention mechanisms are often placed with lateral skip connection layers to improve the efficiency of multi-scale feature fusion in encoder-decoder designs (*312–314*), or to fuse learned features from multi-source branches (*302*). More elaborate structures like gated convolution layers have also been developed to enable joint multi-task learning of shape and semantics (*277*).

The attention module is typically implemented as a multi-branch structure. For instance, self-attention is often defined as a convex function with three branches known as query (Q), key (K), and value (V), with the naming convention inherited from relational database retrieval procedures. In computer vision, these three branches are typically transformations of the same input, which are obtained either by matrix multiplication of the input and separate learnable weights or by passing the input through a convolution layer that matches the input shape (e.g., 1D signal, 2D grid) and the dimensions in which the attention mechanism is applied (e.g., spatial, spectral) (*305*). The query-key-value triplet attention is defined as:

$$Att(Q, K, V) := similarity(Q, K) \cdot V$$

The attention module generates an importance score map by finding similarities between the query (Q) and key (K) branches. The resulting map is then used to weight the value (V) vector, which produces the recalibrated output. The choice of score function used to measure similarity between Q and K is important and can vary depending on the application. Some common similarity functions used in attention mechanisms include feedforward neural networks (*303, 315*), dot product (*316*), cosine similarity (*317, 318*). In dot product self-attention the similarity matrix is usually scaled by the square-root of the feature size of the K branch to avoid numerical instability (*319*).

Co-attention can also be formulated as QKV branches in which K and V come from a different source than the Q vector. For instance, Attention Unet (*313*) uses a spatially-focused co-attention module at each scale level of the decoder before each skip connection. The module applies additive gating between the feature maps from the decoder (Q) and encoder (K, V) paths to preserve the relevant activations while suppressing irrelevant background features. Figure 19 shows an example of this gating mechanism. Although the element-wise summation used in the module formulation acts as a similarity measure and strengthens aligned activations with little computational cost, it cannot capture the full contextual information because it ignores between-pixel correlations. Summation attention is further developed for 3D convolutional networks by Schlemper et al. (*314*). Coupled CNN and Transformer Network (CCTNet) (*320*) is a hybrid model designed for crop mapping. The architecture has two encoders: A ResNet-50 CNN and a CWin transformer branch with a shared simple decoder. The idea is CNN and VT are capturing different patterns due to their different structure so they wanted a fusion strategy that can generate attention maps that can reweight the features of each branch based on the patterns of the other branch. This is achieved by repeating the self-attention but changing the branches that make up the QKV triplet as shown in figure 19b. The Gated-Feedback Refinement Network (G-FRN) (*312*) enhances the integration of multi-level features in FRN-type encoder-decoder networks. This is achieved by incorporating semantics from deeper features with higher receptive field (RF) into lower semantic layers, using a multiplicative gating. The gating unit's output and prediction features are then passed through an aggregation unit to improve the predictions at each level. The gating unit's output and prediction features are then jointly passed through an aggregation unit to improve the predictions at each level (Figure 20).

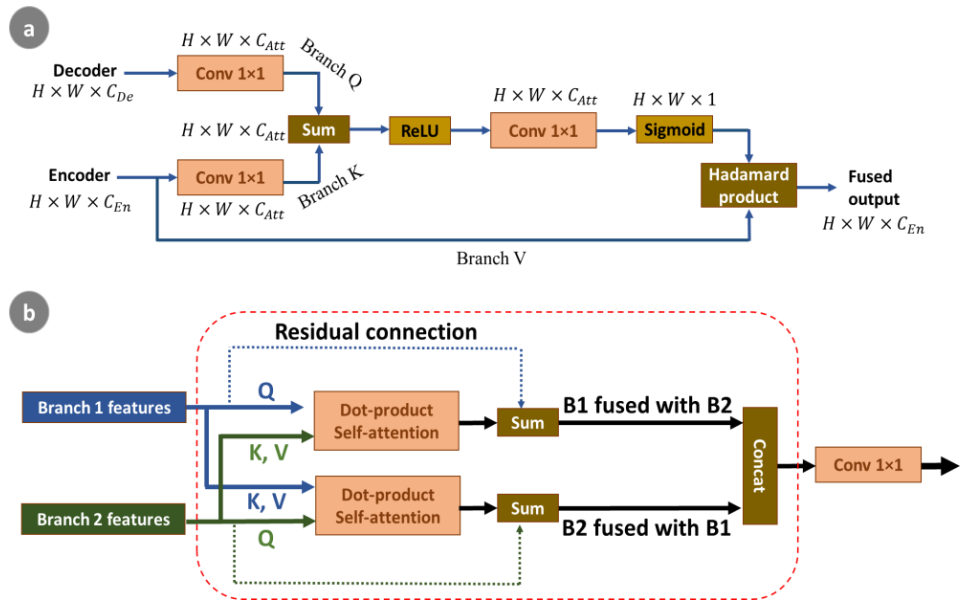

Figure 19. Two formulations of co-attention. a) additive gating strategy modified with an upsampling operation (*313*), b) co-attention strategy for fusing multi-branch information while preserving the information of each branch as introduced in Wang et al. (*320*).

Attention mechanisms can also be used to improve feature fusion across multiple scales. For example, Chen et al. (*321*) used attention to fuse predictions from a set of input images with different scales (e.g. image pyramid input scheme). In this case, the attention mechanism weights the scores at each pixel location for each scale, and uses the weighted sum of all the score maps to generate the final dense prediction.

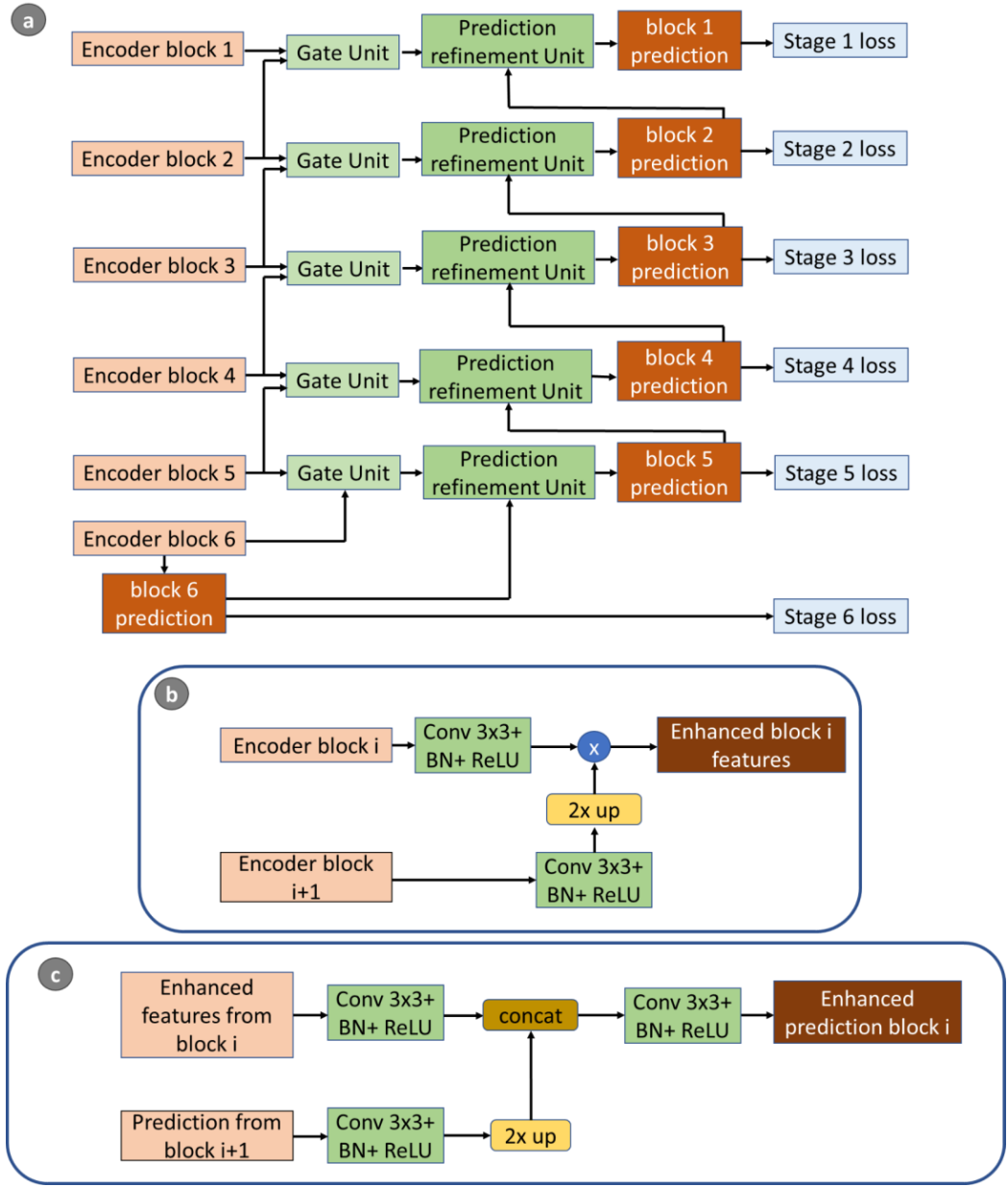

Figure 20. a) Structure of G-FRN b) a gate unit; c) prediction refinement unit

Dot product similarity is a powerful approach for capturing pixel-level correlations between pairs of pixels, which can lead to stronger dependencies compared to additive approaches. One example of this is DANet *(316)*, which uses self-attention through separate modules to capture dependencies in both the spatial and spectral dimensions. In DANet, the QKV triplet branches of the spatial attention module are parameterized using a 1x1 convolution without weight sharing in each branch. The similarity matrix is obtained as the dot product of feature vectors for all positions in the Q and transposed K branches, resulting in a score map of size (HW) × (HW) where each row captures the correlation of each query position and every key position. The score map is then normalized row-wise using softmax. Matrix multiplication between the transposed normalized similarity matrix and branch V provides the attention map, where each position is the weighted sum of an attention weight of size (HxW) with every feature in the V branch. This results in an updated relation-aware V branch that is further scaled by a learnable weight ($\alpha$). The scaled attention is then summed with the original input features to produce an output of the same dimensionality with more aligned activations (see Figure 21a). This approach helps ensure semantic consistency by encouraging adaptive multi-scale fusion and improving the intra-class compactness. In a similar way, interdependencies between channels can be formulated to improve the discrimination of semantic representations by focusing on the most relevant feature maps along the channel dimension (see Figure 21b).

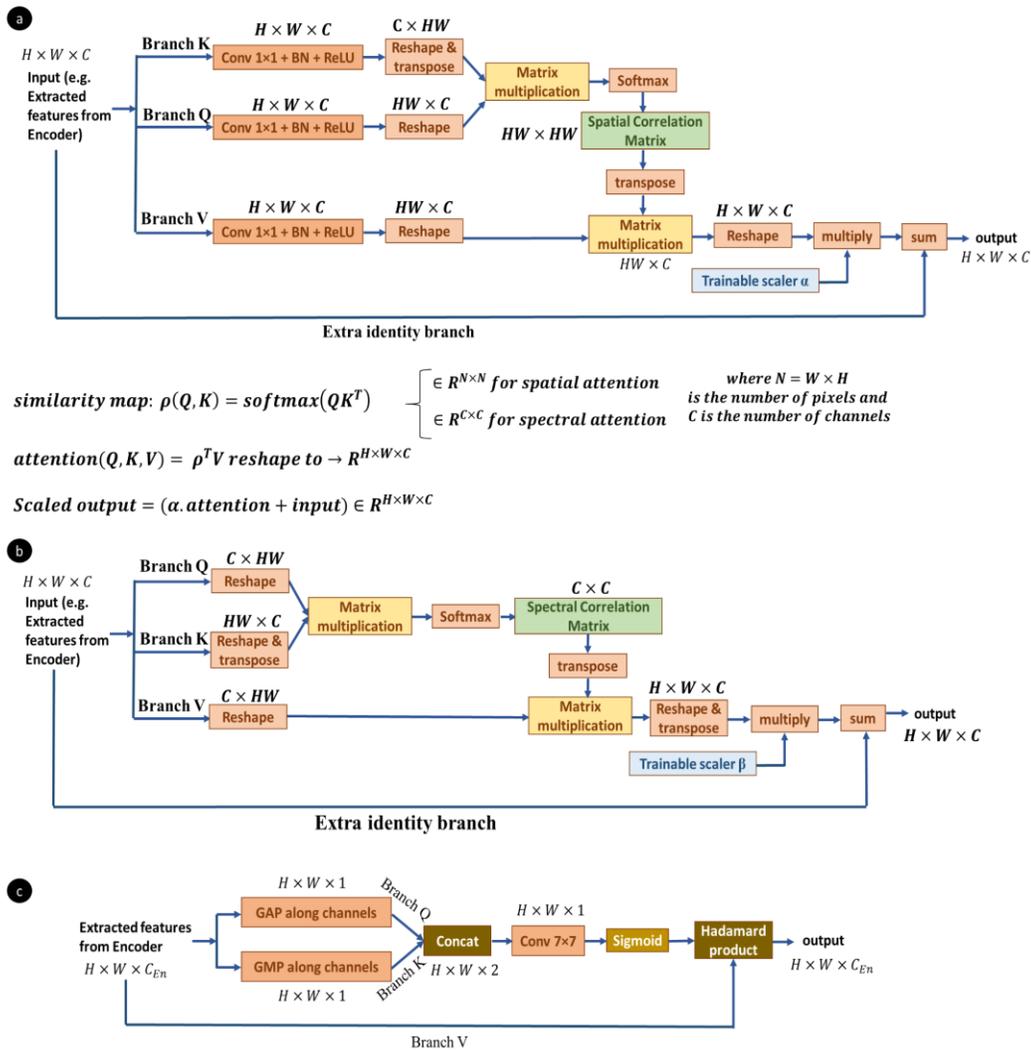

Figure 21. Dot product similarity in attention mechanism used in DANet (*316*). a) Spatial attention module. b) Spectral attention module and c) is a spatial self-attention mechanism (*311*) that also sits between the encoder and decoder branches. The spatial attention module takes the output of the encoder as input, and applies global max pooling (GMP) and global average pooling (GAP) on the channel dimension of the input in parallel. Outputs from the pooling branches are concatenated and passed through a convolution with a large kernel (e.g. 7×7) and sigmoid nonlinearity to make the attention weight, which is multiplied by the original input to generate new feature maps. Notice, when we talk about scaled dot-product attention, we're referring to scaling the similarity map by the factor $\sqrt{d_k}$, where $d_k$ is the dimension of the queries and keys. The scaling is important to prevent the dot product to become too large, causing the softmax to have extremely small gradients which makes the training slow and unstable.

A limitation of dot product similarity is that its computational and memory complexity scales quadratically with the input size, leading to memory and computational complexity of $O(N^2)$. To address this issue, researchers have developed several techniques to reduce complexity, such as sampling the receptive field and the most relevant features or rewriting the similarity function. For example, Fu et al. (*201*) used a pyramid pooling module to sample features at different scales (e.g. spatial bins) into a compact representation, which was then used to build spatial correlations. They also used dimensionality reduction through a 1x1 convolution to make channel attention more efficient. In a similar way, ICENET (*322*) used a two-branch design to classify river ice in UAV images, with one branch gathering spectral relationships across scales using three channel attention modules attached to the last three blocks of ResNet-101, while the second, shorter branch uses two strided convolutions to reduce the spatial dimensionality to 1/8 for input to a spatial attention module. CCNet (*323*) reduces the dot-product computational complexity by restricting the receptive field to the axial direction in the attention module using convolution layers with asymmetric kernels. The CCNet attention module consists of two criss-cross attention modules that share parameters and are arranged serially to capture dense contextual information. By finding the dependencies (e.g. correlation) through an affinity operation on the Q and K branches, in which the dot-product between each spatial position in Q and those positions in K that share the same row and column index as the pixel position in Q is calculated, which generates a compact spatial correlation map with the size (H+W-1)×(H×W). Once the spatial correlation map is computed, it is normalized using channel-wise softmax to obtain an attention map that captures the importance of each position in the feature map. This attention map is then applied to the value (V) feature map using an aggregation operation, such as matrix multiplication, to obtain the final enhanced feature map. Li et al. (*324*) uses first-order Tylor approximation and L2 normalization of the Q and K to reduce the complexity of dot product similarity to O(N). MANet (*325*), formulates the similarity function using a kernel strategy, which calculates the dot product between the softplus of Q and K with linear complexity. CANet (*326*) utilized dot product similarity in a co-attention setting to capture dependencies between color and depth for optical and depth input modalities.

Squeeze-and-Excitation (SE) spectral attention is another attention mechanism introduced in SENet (*327*) that can be easily integrated into existing CNN architectures like ResNet, DenseNet, or Inception variants. The SE module consists of two parts: the squeeze and excitation stages. In the squeeze stage, global average pooling is used to generate a descriptor vector for each channel, which captures its relevance and embeds global contextual information. The excitation stage involves two dense (FC) layers followed by ReLU and Sigmoid activation functions, respectively. These layers re-calibrate the input feature channels using the descriptor vector from the squeeze stage (Figure 22a). The first FC layer projects the descriptor vector to a lower dimension, and the second projects it back to the original dimension and outputs the final weights, which are used to scale the input signal. This formulation of the excitation component captures the non-linear interrelationships among channels, and ensures that the learned relationships are not mutually exclusive by allowing emphasis on multiple channels. The Gather-Excite mechanism (GE) (*328*) is a more general extension of the SE module. In this mechanism, the gather (e.g. squeeze) stage occurs on large local spatial neighborhoods through a parameter-free (e.g. average pooling) or learnable (e.g. strided convolution layer operation) process. There have been other efforts to improve the SE module, such as choosing a more descriptive global aggregator in the squeeze stage through the use of second-order pooling (*151*), or combining global average pooling with global standard deviation pooling (*329*). There have also been efforts to create a more efficient excitation mechanism, such as ECANet (*330*), that uses 1D convolution layers instead of FC layers and examines each channel's relationship with its k nearest neighbors to reduce the computational burden.

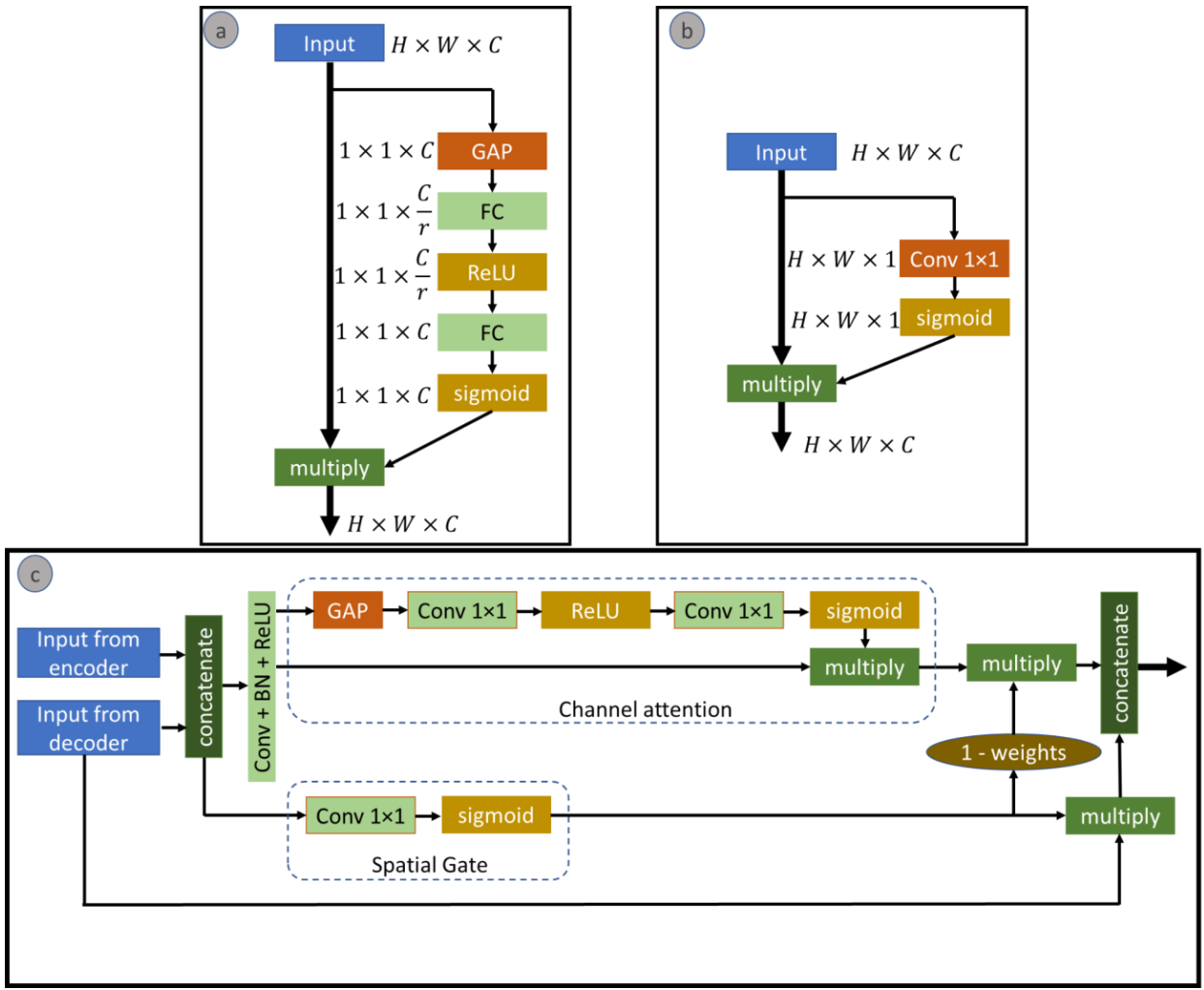

Figure 22. Variations of squeeze and excitation attention module: a) Channel attention where the squeeze happens in the spatial dimension and excitement on the channel dimension. b) Spatial attention with squeeze on the channel dimension and excitement on the spatial dimension. c) Parallel arrangement of channel and spatial SE attentions modified in the form of co-attention.

The Squeeze-and-Excitation (SE) module uses Global Average Pooling (GAP) as a contextual aggregator, but the effectiveness of this approach at large spatial extents is unclear. For example,

Panboonyuen et al. (*331*) used a SE-like spectral self-attention to strengthen the signal in the decoder branch of a global convolution network (GCN) by employing global pooling followed by 1×1 convolution layers and a sigmoid layer to generate importance scores. In this case, global pooling aggregated context at the patch level. Recent experiments by Islam et al. (*332*) on street-view images showed that GAP can encode coordinate-based positional information along the channel dimension based on the ordering of the channels. However, Ding et al. (*333*) argue that scene-level contextual information is vague for remote sensing imagery, especially those with a large swath width. To address the lower effectiveness of global aggregators over larger areas, the authors used patch-based average pooling in a self-attention module to extract meaningful local context descriptors and a second patch-wise co-attention mechanism to fuse low- and high-level features.

The SE concept has also been extended to the spatial dimension by incorporating channel squeeze and spatial excitation (see Figure 22b), which is a better fit for semantic segmentation tasks (*334*) and already been integrated in several networks designed specifically for satellite imagery (*181, 335, 336*). For instance, the RSSAN model (*337*) employs a sequential arrangement of spectral and spatial SE modules, where the output of channel attention is used as the input of the spatial attention module. These modules are positioned at the first stage of the network, where they receive small cubes of hyperspectral imagery as input. The spectral attention module in RSSAN is modified to have two branches, one with GAP and the other with global max pooling. The two branches provide complementary context aggregators that more efficiently use spatial information than either individual operation to weight the channel dimension. The descriptive vectors from each branch pass through two FC layers with shared parameters and are then summed to produce the final attention weights. In Jin et al. (*291*), the attention mechanism is designed as a parallel arrangement of channel and spatial SE, as shown in Figure 22c. The attention module is placed on all lateral skip connections between the encoder and decoder paths.

Another example of combining spatial and spectral attention is seen in the work of Mou et al. (*338*), where the effect of attention mechanisms on the performance of Fully Convolutional Networks (FCN) for classifying very high-resolution aerial imagery was examined. In this approach, a dot product is used in both spatial and spectral dimensions, but the spatial attention module applies ReLU instead of softmax on the similarity score, and directly concatenates it with the original input. The spectral attention is a hybrid inspired by both DANet and SENet. Similar to previous examples, attention modules are arranged in a serial manner, where the output of spectral attention provides input to the spatial attention module. Yang et al. (*302*) introduced a co-attention mechanism that fuses features at each level of a dual-branch encoder to better integrate multi-source input. Their shared decoder uses a different attention module in each level of upsampling to efficiently fuse multi-scale features. Similarly, Dong et al. (*339*) used a customized multiplicative spatial attention module between low-level features of ResNet-101 and a channel attention module on the ensemble features from different blocks of ResNet, along with a separate spatial branch. Jin et al. (*291*) proposed the Gated-Attention Refined Fusion Unit (GARFU), which combines spatial gating and channel attention and is added to each skip connection to fuse multi-level features.

In Li et al. (*340*), the authors argue that many global aggregation (GA) modules, such as those formulated as pyramid pooling, ASSP, or attention mechanisms, tend to miss small objects or smooth their boundaries too much, as global statistics are biased towards larger, more abundant objects. This means that pixels from small objects, particularly those near edges, may capture contextual information from pixels belonging to other objects. In a diverse and unbalanced scene, pixels from small objects may change categories, increasing the miss rate of small objects. To address this issue, the authors propose GALD, which adds a local distribution (LD) module after the GA. The LD module weights each position in the GA module's output by recalculating the spatial extent of that position, which changes the distribution of features in GA and produces a finer-grained output. Jin et al. *(291)* integrate the long-range contextual module (e.g. SPP, ASPP or attention mechanism), using semantic level contextual information to provide a more efficient and effective context aggregation, which establishes local dependencies within individual category regions. Additionally, Ghaffarian et al. *(341)* provide a systematic review of attention mechanisms and their use in different RS application domains, as well as the effects of such mechanisms on performance improvement.

## 5. Semantic segmentation with sequential data

A unique characteristic of satellite imagery is its ability to provide high-frequency observations of the same geographical area through time, offering valuable additional information that can help with discrimination of objects on the Earth's surface. However, these sequences exhibit statistical dependencies, necessitating models that are explicitly designed to understand and utilize these dependencies. Satellite image time series (SITS) are the most common form of sequential data in remote sensing, and they can be harnessed to train models that analyze data across temporal-spectral dimensions (e.g., per-pixel analysis) or across temporal-spectral-spatial dimensions (e.g. image patch cubes) by extracting discriminative features for categories that have distinct temporal profiles *(342)*, such as crop types with distinctive growth cycle stages *(343, 344)*. The strategic use of sequential data can effectively mitigate the impact of atmospheric noise or fluctuating illumination *(345, 346)*, particularly useful when classifying imagery with moderate spatial resolution, such as Landsat and Sentinel *(347)*. Besides time series, sequential inputs can also come from other aspects of data, such as the spectral bands in hyperspectral imagery (HSI) *(348)*, or image pyramids of different scales *(349)*.

The process of structuring SITS typically involves separating the temporal from the spatio-spectral dimensions, leading to a sequence of multi-spectral image snapshots. However, the multidimensional arrays representing hyperspectral imagery (HSI), SITS, or other sequential data from various sensors can be reshaped to align with the requirements of different dimensionalities in Recurrent Neural Networks (RNNs) or CNNs. From a modeling perspective, how the axes of the array are arranged determines the types of patterns the model can learn, whether they're rooted in spectral-spatial, spectral-temporal, spectral-spatial-temporal domains, or others.

### 5.1 1D convolution networks

One-dimensional convolutional neural networks (1D CNNs) are commonly used for classifying both HSI and SITS. In its generic form, a 1D convolution takes a single data cube $x_{in} \in R^{L_{in} \times C_{in}}$ as input, applies a weight matrix $W \in R^{C_{in} \times k \times C_{out}}$ and bias vector of $b \in R^{C_{out}}$, resulting in the output $x_{out} \in$ $R^{L_{out} \times C_{out}}$, where L represents the sequence length (e.g. temporal dimension), C is the number of measurement features (e.g. spectral bands) and k stands for the kernel size. As an example, C equal to 1 represents a univariate time series, such as an NDVI profile, while C > 1 indicates a multivariate time series, such as multiple bands from Sentinel-2.

In general, 1D CNNs perform optimally on sequences with equal time intervals, although this isn't a mandatory requirement. While per-pixel classifiers such as 1D CNNs that don't consider spatial relationships are susceptible to salt-and-pepper artifacts, they generally outperform conventional machine learning classifiers like Support Vector Machines (SVMs) *(350, 351)*. A recent study by Zhao et al. *(352)* indicates that 1D CNNs are less sensitive to missing data than RNNs, and they rank second in accuracy after hybrid CNN-RNN models, provided the missing data doesn't significantly affect the sequence's relevant timestamps. One strategy for managing SITS with non-uniform intervals involves explicitly encoding each time-point's acquisition date as a new band for example, in the form of the day of the year *(259)*. Alternatively, one can encode the time difference between each time-point and the preceding one to highlight gaps in the time series *(353)*.

One-dimensional (1D) networks used for SITS analysis typically have a simpler, shallower structure compared to 2D models mainly due to the smaller size of SITS datasets, which increases the chance of overfitting as model complexity increases *(354, 355)*. These networks often consist of several Conv1D layers, followed by optional downsampling. The output is then flattened and processed through a series of dense layers. The final layer, equipped with softmax activation, serves as a classifier, providing output class probabilities *(356–360)*.

Temporal patterns in SITS often manifest at varying scales especially if the dataset covers a large geography. Extracting these multi-scale features requires a large and varied RF in the network. Using downsampling, larger kernels or deeper models are common strategies to enlarge the RF. Yet, there isn't a clear consensus on the use of pooling in 1D networks. Some modern architectures such as 1DVGG *(361)* completely drop all pooling layers while others like 1DResNet *(362)* or InceptionTime *(354)* avoid using pooling between the conv layers. Instead, these architectures utilize a Global Average Pooling (GAP) layer to average the learned features over the entire temporal extent prior to transmitting the signal to the dense layers. Contrarily, TempCNN *(363)*, a revised version of 1DFCN developed for crop type mapping from Formosat-2 image time series, eliminates the GAP layer due to its perceived negative impact on performance. This model comprises three triplet layers composed of conv, BN, and ReLU layers, followed by dense and softmax layers with 128, 256, and 128 features for the conv layers, respectively. Similarly, Ma et al. *(364)* developed a modified 1DResNet with three blocks and no pooling or GAP layer to perform per-pixel cloud classification from different RS sensors.

In contrast to 2D networks, breaking a conv layer with a large kernel into a series of conv layers of smaller kernels will not reduce the number of parameters in a 1D network. Furthermore, the need to enlarge the RF in shallower depth 1D networks usually lead to the usage of large kernel sizes. However, the choice of kernel size is also positively related to sequence length and the nature of patterns that exist, where a larger kernel is needed to detect longer patterns *(354, 363, 365)*.

To optimize feature extraction for time series classification (TSC) tasks, the best feature extraction scales often vary between datasets due to the natural composition of multiple signals on different scales within time series data. Augmentation or multi-branching with various kernel sizes can be used as solutions to scale selection. For instance, Multi-scale Convolutional Neural Network (MCNN) *(366)* is a designed architecture that extracts features from different time scales in equal interval time-series. MCNN has three stages: the first stage transforms the time-series with smoothing and downsampling augmentation, resulting in two transformed time-series of different lengths beside the original. Each time-series is passed as input to the next stage. The local convolution stage in MCNN consists of three independents parallel 1D convolutional layers with the same kernel size, which are designed to capture different temporal scales from the varied-length input time-series. In each branch, convolutional layers are followed by max-pooling with multiple kernel sizes, controlled by a pooling factor that decides on the output length, and usually set to values like {2,3,5}. Features from all branches are then concatenated vertically and fed to the full convolution stage, which is designed to capture the overall temporal structure of the input time-series. This stage consists of a 1D convolution followed by a dense layer and a softmax classifier to make the final prediction. A novel approach to address the challenge of scale is proposed by Tang et al. *(367, 368)*. They suggest that while the receptive field (RF) size is important, 1D-CNNs are not highly sensitive to specific kernel size configurations that compose the RF size. Instead, the performance of 1D-CNNs is primarily determined by the optimal RF size chosen in the network. To capture features at all scales, the authors propose a kernel size configuration inspired by Goldbach's conjecture, which states that every even integer greater than 2 can be expressed as the sum of two prime numbers. This approach involves using a set of prime number kernel sizes in an OS block, enabling a 1D-CNN to capture RFs of all scales. The resulting OS-CNN architecture comprises an OS block with a global average pooling (GAP) layer and a fully connected (FC) layer for classification. The approach is flexible and can be extended to include additional layers.

Classifying imagery based solely on the pixel's spectral characteristics is also done for hyperspectral imagery *(369)* or when the annotated labels are gathered pointwise *(370)*. In these cases, an image is sampled into pixel cubes of shape $x_{in} \in R^{C \times 1}$, where $C$ is the number of channels in the sequential dimension. There are also more elaborate schemes to shape the input, such as using PCA on the target pixel and its surrounding pixels, then concatenating the first Q principle components to the original input bands, in order to encode a more stable signal *(371)*. From the design perspective, these networks are mostly shallow but, unlike 1D networks used for SITS, they usually utilize pooling layers *(356, 364, 372–374)*. Another common design choice is in the form of a Dual branch CNN where one branch consisting of a 1D network is responsible to learn the spectral features and the other 2D or 3D branch is dedicated to learn the spatial features *(357, 369, 375–377)*. The input fed to the 1D branch is usually a single pixel with all the bands, but the 2D network receives a small patch centered on the target pixel with bands reduced through similarity or dimensionality reduction techniques *(369)*. As mentioned previously (section 4.4), the fusion strategy between branches is an important design decision in such models. For instance, Zhang et al. *(357)*, utilized decision-level fusion to learn the joint spectral-spatial features, while, Yang et al. *(375)* used a deep fusion approach by concatenating features learned from each branch and combining them with dense and softmax layers to make class probabilities. More elaborate multi-point deep fusion procedures are also possible, such as those that employ a specialized module for extracting and fusing the features from the dual spatial and spectral branches *(377)*. In a different setting, Kussul et al. *(378)* stacked multi-sensor multitemporal data along the channel dimension. Specifically, 4 timepoints of Landsat 8 each with 6 bands and 15 Sentinel-1 each with 2 bands were stacked. Each pixel of the resulting 54 channels was then fed into a simple 1D network consisting of a pair of conv layers followed by Maxpooling to classify summer crop types in Ukraine.

**5.2 2D convolution networks modification to handle SITS**

CNNs based on conv2D can also be used to handle the temporal dimension of a spatiotemporal dataset (e.g. SITS) by feeding it to the model as 3D patches with $H \times W$ spatial dimension and $C \times T$ channels, where $C$ is the number of spectral bands and $T$ is the number of time points *(379)*. This dataset-level fusion encodes the temporal dimension as part of the spectral variability. However, despite its simplicity, this approach has several drawbacks, primarily because it doesn't take into account the temporal order of the images and the variable time intervals between images. It's also not shift invariant, meaning the model wouldn't recognize the same event happening at different times in the sequence as being the same.

Early/late fusion is an alternative method that processes each time point in a separate branch using Conv2D, with resulting feature maps fused before the classifier learns the spectral-spatial features along the temporal dimension. For instance, El Mendili et al. *(380)* proposed the temporal Unet, where each timestamp undergoes a separate dedicated encoder. The outputs of these branches are concatenated and passed through a shared decoder. However, such models are complex and require large computational resources, with complexity increasing with sequence length.

Another strategy for processing SITS is to treat each pixel as a 2D image with a single band and feed it into a classification CNN. This approach jointly learns the spectral and temporal features of the SITS. Debella-Gilo and Gjertsen *(365)* applied this strategy to map three agricultural land use classes using Sentinel-2 time-series. However, compared to 1D CNNs, this method tends to have lower generalization power. In 1D CNNs, the optimum number of filters and filter size is usually larger, while the optimum dropout rate shows the opposite trend.

**5.3 3D Convolution networks**

The use of 3D convolutional layers in SITS architectures requires higher memory and computing resources than 2D convolutional layers, but the extra dimension allows 3D convolution networks to learn either spatial-spectral features along the temporal dimension (input shape: T×C×H×W) or spatial-temporal features along the spectral dimension (input shape: C×T×H×W), where spectral and temporal dimensions are independent axis dimensions *(381)*. The latter configuration has been found to be more effective for crop classification using GF1 & 2 imagery, and a standard 3D VGG network that classifies each pixel independently. A later paper by Ji et al. *(382)* introduced a 3D symmetric encoder-decoder architecture (figure 23) with a global aggregation module between the encoder and decoder parts and 3D channel attention attached to each between-the-scale-level skip connections for improved feature fusion. Similarly, in both models, the downsampling and upsampling operations are designed to preserve the temporal information of the growth cycle, and the sequence length is kept intact as the temporal Gaofen-2 dataset they used was sparse with only 4 and 7 timepoints for years 2015 and 2017 respectively. Despite the success of the spatial-temporal approach, the spatial-spectral configuration is also commonly used (383, 8). A 3D Unet with dilated convolutions using spatial-spectral inputs has also been reported (384). Additionally, many 2D architectures are extended to their 3D designs, such as extending residual design (385) or multi-kernel Inception blocks with variable temporal convolution kernel depths embedded into a 3D extension of DenseNet (386). Bhimra et al. (383) used a 3D-ResNet-34 model to classify four anthropogenic transitions (e.g. construction, deconstruction, cultivation, de-cultivation) from a sparse multi-year SITS dataset. Han et al. (387) used a shallow network consisting of three 3D convolutional layers, each followed by a max pooling, and the last 3D convolution is equipped with Squeeze-and-Excitation attention to extract features from EO-1 and Landsat-8. Learned features are fed into an SVM classifier to output sea ice dense prediction maps. The STEC-Net (388) processes image sequence chips of HxWxF where F is the sequence length using cascaded 3D convolution-pooling networks, simultaneously extracting spatial and temporal features. To ensure sequence causality in temporal convolution — meaning current outputs depend only on current inputs and past data — padding with zero images is applied at the beginning of the temporal dimension for 3D convolution, specifically padding of $(temporal\ dilation\ rate - 1) \times (sequence\ length - 1)$. Temporal dilation rates increase across layers to optimize information propagation over sequence gaps while Spatial dilation rate decrease to prevent the receptive field from exceeding image boundaries. A compact connection is introduced, replacing computationally expensive FC layers with 3D convolution with kernel size mirroring the input feature tensor size which are then flattened, and their last elements are concatenated for the output vector. Due to the causality of the dilated 3D convolution, the last time slice accumulates all sequence information, enhancing classification performance. The model was tested on the MSTAR dataset, comprising timeseries of single-band SAR imagery. There are also hybrid architectures to mitigate high cost of 3D CNN models. For instance, Saralioglu and Gungor (389) demonstrated that incorporating just one 3D convolution, followed by 2D convolutions, can boost the performance of a CNN encoder. This enhancement proved effective in classifying land cover classes with similar spectral responses, especially in the very high-resolution imagery they examined, including Worldview-2, IKONOS, Pleiades, and Deimos-2.

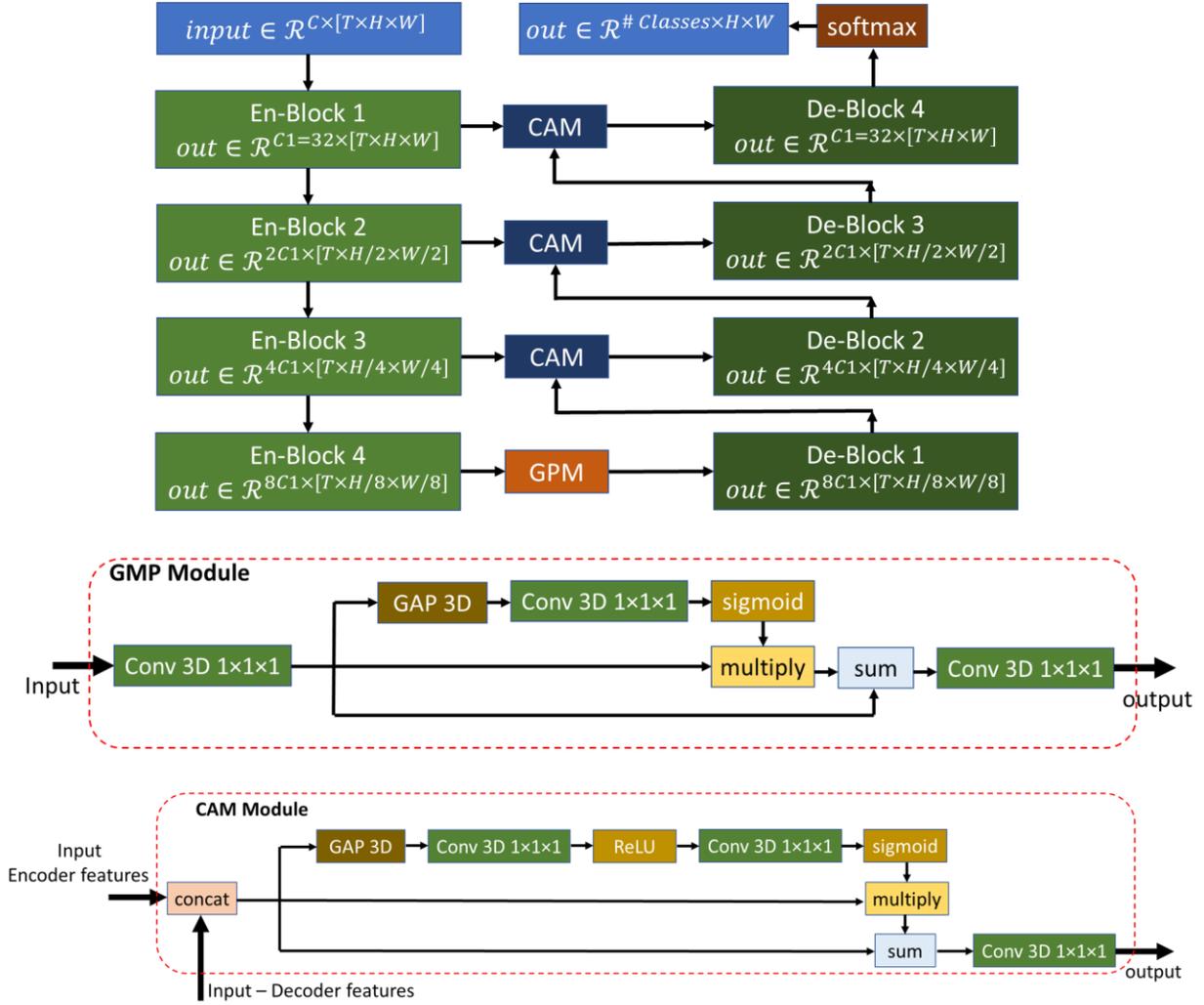

Figure 23. Example of a symmetric encoder-decoder with 3D conv layers augmented with a global pooling module and channel attention mechanism for better feature fusion *(382)*. Notice the GAP in the GMP module is applied over channel dimension while the GAP in the CAM module is applied over the spatial dimensions.

3D CNNs are also popular for HSI image classification, given their ability to extract discriminative features from the densely correlated spectral bands of HSI. The 3D convolutions preserve both the spectral and spatial information throughout the network, resulting in improved performance. However, the high correlation between HSI bands and the additional parameters required by 3D conv layers demand larger training datasets than those required for networks based on 2D convolutions. To control the number of model parameters, hybrid networks combining both 3D and 2D conv layers have been proposed, and they have outperformed their purely 2D or 3D equivalent structures in all the benchmark datasets investigated. Hybrid structures have shown promising performance in several studies, including *(390–394)*.

## 5.4 RNNs

The Recurrent Neural Networks (RNNs) are a type of ANN specifically designed to process sequential data inputs. They are capable of forming feedback loops within the network, which allows the model to behave as a non-linear dynamic system. A basic RNN architecture comprises an input layer, hidden layer, and output layer for each element of the sequence. However, deep RNNs with multiple hidden layers are also common, in which the hidden layers are stacked and each layer passes its activations to the next higher-level hidden layer for each element of the sequence.

Training RNNs mostly involves the backpropagation through time (BPTT) algorithm. BPTT requires the model to simulate a feedforward network by unrolling over the sequence dimension *(395)*. This allows BPTT to calculate the gradient along the sequence length instead of the layer-wise calculations performed in standard backpropagation. In the case of SITS, a standard RNN takes 2D input which means that for each pixel in the imagery, the time points are held in rows, and the band values in columns *(396)*. Another approach is to use the aggregated spectral response from the pixels comprising a specific object, such as an individual crop field *(397)*.

RNNs consist of three sets of trainable parameters: weight matrices for the input-to-hidden ($W_{xh}$), hidden-to-hidden ($W_{hh}^l$), and hidden-to-output ($W_{h\hat{y}}$) connections, as well as bias vectors $b_h^l$ and $b_{\hat{y}}$ for the respective hidden and output layers. These parameters are shared across each sample in the sequence as the network unrolls (as shown in Figure 23), with the weights being shared within each layer but not between them *(398)*. RNNs differ in how they handle the feedback cycle and input-to-output structure. There are many-to-one (the most common) and many-to-many designs, with the latter being used for tasks like mapping the rotational sequence of crop types within a single or over multiple years, where predictions need to be made for each timestamp *(399)*.

RNNs can handle input sequences of varying length, but training lengthy sequences using BPTT can be problematic, as the gradients tend to become exponentially smaller or larger as a function of sequence length *(400)*. To reduce the exploding gradient problem, an upper bound is put on the norm of the gradient vector, a procedure known as gradient clipping. However, the most popular strategy to overcome both exploding and vanishing gradients is to use gating mechanisms, which provide better control over the internal dynamics of the recurrent model by enabling each recurrent node to dynamically filter its input *(401)*. Long short-term memory (LSTM) *(402)* and gated recurrent unit (GRU) *(403)* are good examples of such sophisticated computational units.

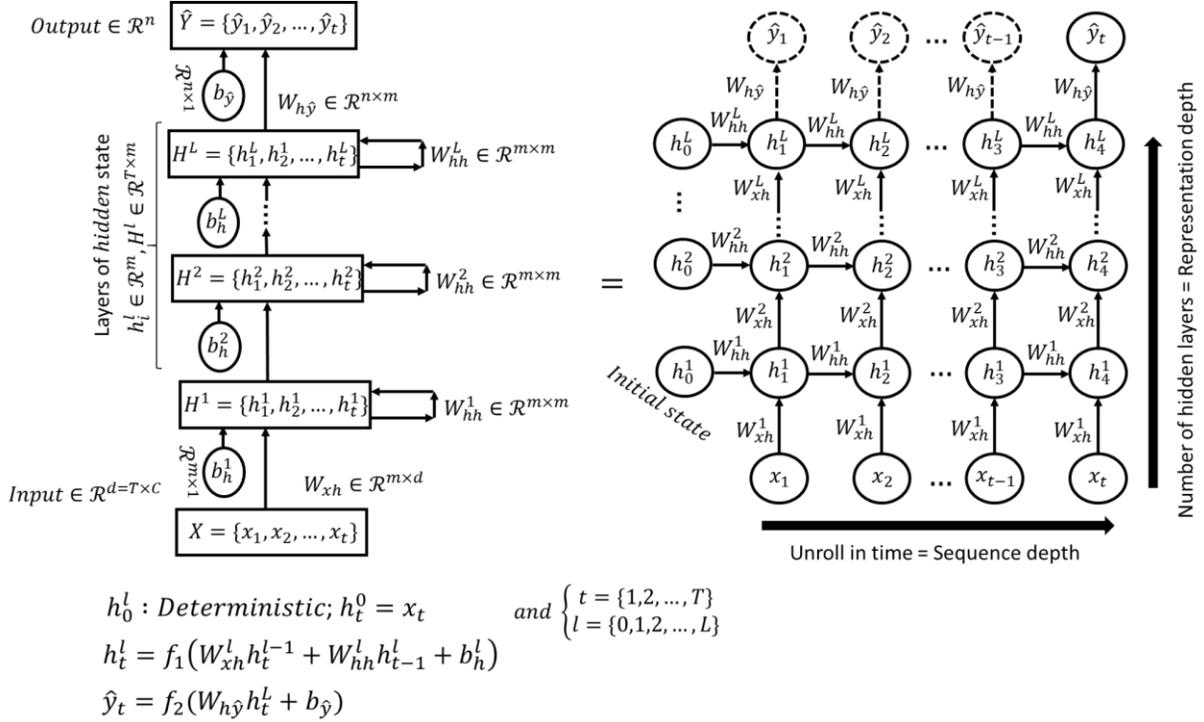

$h_0^l$ : Deterministic; $h_t^0 = x_t$ and $\begin{cases} t = \{1,2,...,T\} \\ l = \{0,1,2,...,L\} \end{cases}$

$h_t^l = f_1(W_{xh}^l h_t^{l-1} + W_{hh}^l h_{t-1}^l + b_h^l)$

$\hat{y}_t = f_2(W_{h\hat{y}} h_t^L + b_{\hat{y}})$

Figure 23. Generic RNN with multiple hidden layers on the left, and the same model unrolled over time on the right. Hidden state is initialized using a deterministic function (in practice set to zero) and gets updated in the forward pass based upon the previous state and the external input going through a non-linearity ($f_i$) over the sequence with tanh and sigmoid as common activation functions for hidden and output layers respectively. $W_{ij}^k$ is the weight matrix transforming layer i to j with k indexing the transition between hidden layers. Weight parameters are usually initialized through orthogonal and identity matrices and bias terms are initialized with zero. The unrolled RNN network on the right creates a grid where gradients are flowing from left to right along the sequence dimension and bottom to top along the layers (e.g. network depth).

A standard LSTM cell applied to each sequence element consists of a cell state, hidden state, and three gates (figure 24). The forget gate determines which information to keep from the previous cell state, while the input gate decides which part of the input should update the current cell state. The candidate cell state rescales the input and previous hidden state to update the current cell state, and the output gate controls which information from the current cell state should be passed to the next cell. In a multi-layer LSTM network, the output of the LSTM layer is a sequence of hidden and cell state vectors, one for each time step in the input sequence. The output of the LSTM layer can be further processed by subsequent layers, or it can be used as the final output of the network, depending on the task being performed. There is also a variant of the standard LSTM that allows the gates to have additional peephole connections to the cell state which might improve the model's ability to capture long-term dependencies.

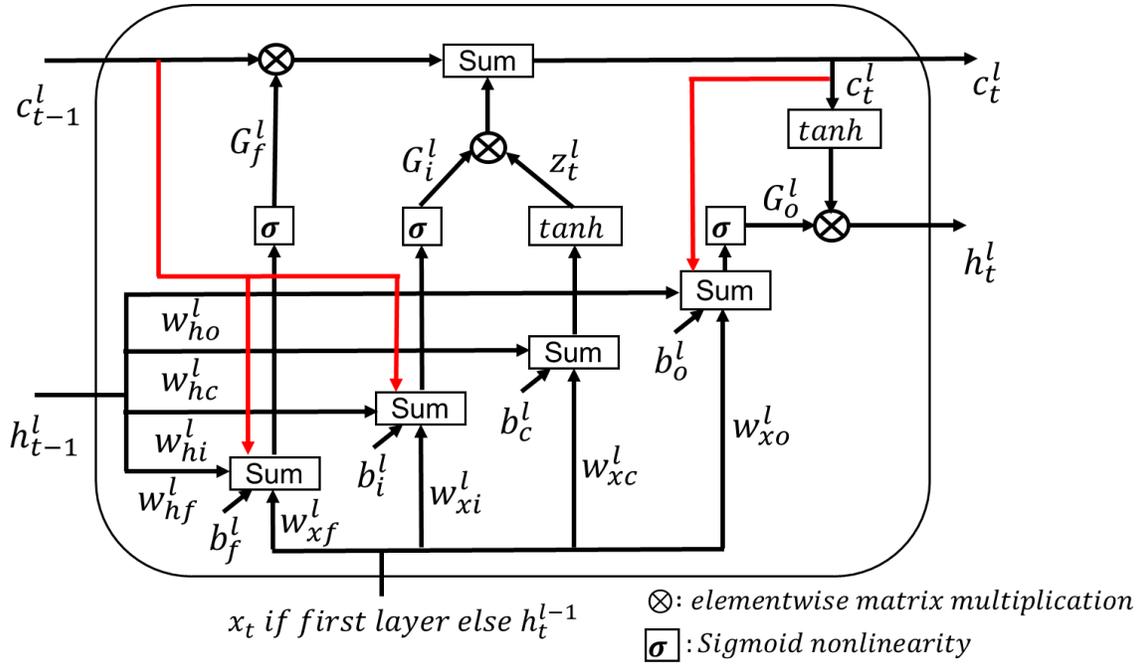

$Forget\ Gate: G_f^l = \sigma\left(w_{xf}^l h_t^{l-1} + w_{hf}^l h_{t-1}^l + w_{cf}^l c_{t-1}^l + b_f^l\right)$

$Input\ Gate: G_i^l = \sigma\left(w_{xi}^l h_t^{l-1} + w_{hi}^l h_{t-1}^l + w_{ci}^l c_{t-1}^l + b_i^l\right)$

$Candidate\ Cell\ State: z_t^l = tanh\left(w_{xc}^l h_t^{l-1} + w_{hc}^l h_{t-1}^l + b_c^l\right)$

$Current\ Cell\ state: c_t^l = G_f^l \otimes c_{t-1}^l + G_i^l \otimes z_t^l$

$Output\ Gate: G_o^l = \sigma\left(w_{xo}^l h_t^{l-1} + w_{ho}^l h_{t-1}^l + w_{co}^l c_t^l + b_o^l\right)$

$Cell\ Hiden\ State: h_t^l = G_o^l \otimes tanh(c_t^l)$

Figure 24. Detailed structure of a single standard LSTM cell in a multilayer structure. Peephole connections in both schematics and formula are shown in red.

A GRU, on the other hand, is a simplified version of LSTM with fewer trainable parameters (Figure 25). It has no cell state and only two gates: an update gate and a reset (or relevance) gate. The update gate decides how much of the previous hidden state and the current input (e.g. sequence element) should be kept, while the reset gate combines forget and input gates and determines how much of the previous inputs should be added to the cell's hidden state. Both LSTMs and GRUs use Tanh and ReLU functions as effective hidden state activations (404). Cells can stack together to make larger models. The last hidden state from the last LSTM or GRU cell or each cell is fed into a FC layer with SoftMax activation, with as many neurons as the number of classes (342).

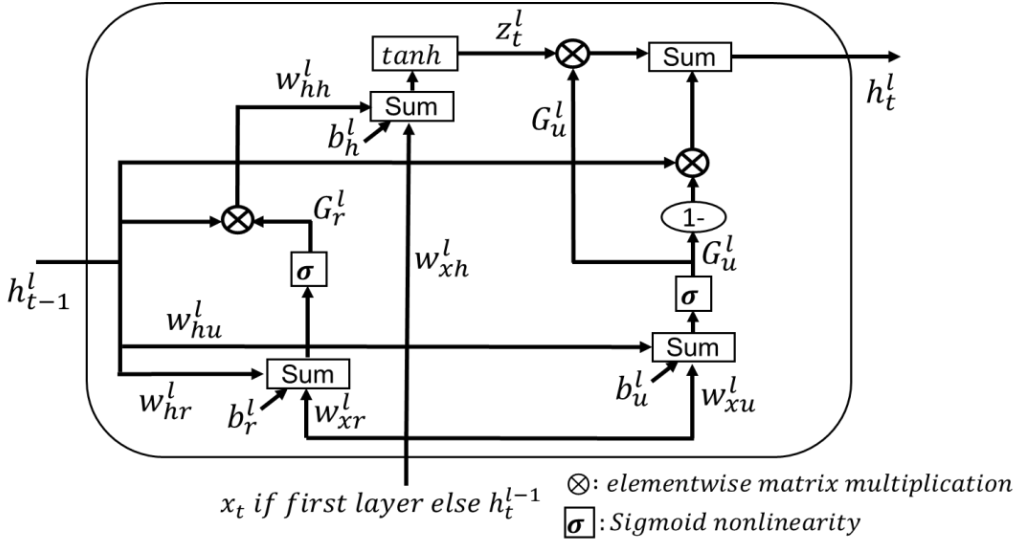

$Reset\ Gate: G_r^l = \sigma(w_{xr}^l h_t^{l-1} + w_{hr}^l h_{t-1}^l + b_r^l)$

$Update\ Gate: G_u^l = \sigma(w_{xu}^l h_t^{l-1} + w_{hu}^l h_{t-1}^l + b_u^l)$

$Candidate\ Hidden\ State: z_t^l = tanh(w_{xh}^l h_{t-1}^l + (G_r^l \otimes h_{t-1}^l)w_{hh}^l + b_h^l)$

$Cell\ Hiden\ State: h_t^l = (1 - G_u^l) \otimes h_{t-1}^l + G_u^l \otimes z_t^l$

Figure 25. Detailed structure of a single standard GRU cell.

The LSTM and GRU cells can be modified into a bi-directional form that processes sequences in both forward and backward directions *(405)*. This architecture allows the model to take advantage of both past and future context in predicting the current output. The bi-directional cell is composed of two layers: a forward layer that processes the input sequence from the beginning to the end and a backward layer that processes the input sequence from the end to the beginning. The output of a single bi-directional cell is the concatenation of the outputs of the forward and backward cells at a given time step.

Although LSTMs and GRUs can help with vanishing gradients along the sequence dimension, the number of layers in a multi-layer RNN is generally much smaller than in CNN models. This is partly due to the much smaller size of publicly available annotated sequential remote sensing datasets for training the recurrent models but mainly because these models cannot control the scale of the gradient magnitude flowing between layers. To address this issue, a new gating cell called STAckable Recurrent (STAR) unit *(398)* has been proposed. The STAR unit can keep the magnitude stable in the gradient lattice, making it possible to stack more recurrent cells and develop deeper models while requiring fewer parameters than other common gating cells (Figure 26).

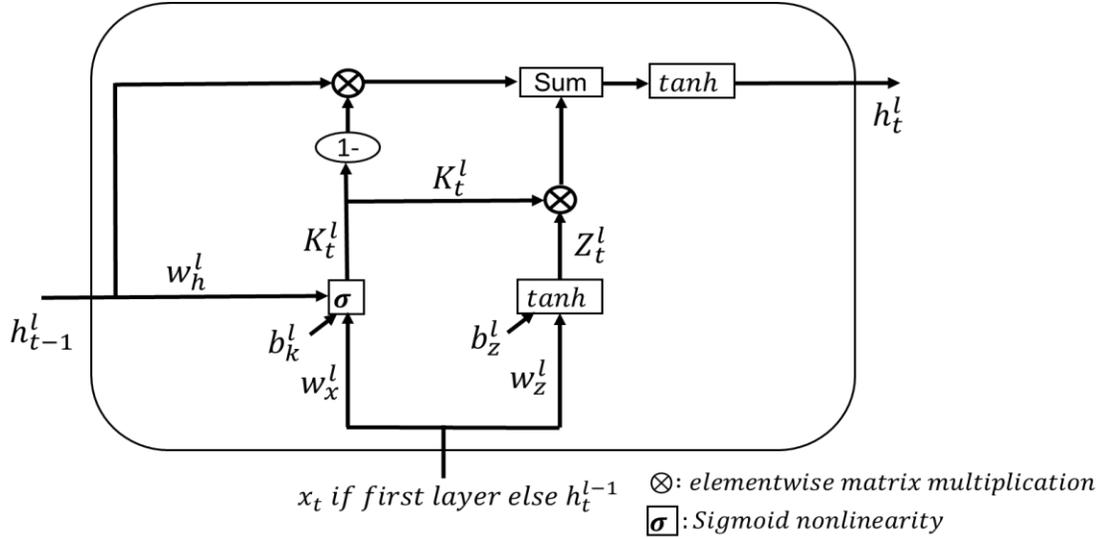

Input Gate: $Z_t^l = tanh(w_z^l h_t^{l-1} + b_z^l)$

Forget Gate: $K_t^l = \sigma(w_x^l h_t^{l-1} + w_h^l h_{t-1}^l + b_u^l)$

Cell Hiden State: $h_t^l = \tanh((1 - K_t^l) \otimes h_{t-1}^l + K_t^l \otimes Z_t^l)$

Figure 26. Detailed structure of a single STAR cell.

    The LSTM and GRU cells rely on matrix multiplication through an FC layer to form gates and internal states, which require the input sequences to be flattened into vectors. This makes it difficult for these models to handle the spatial dimension of remote sensing data. To overcome this limitation, hybrid RNN/CNN models have been developed that can leverage both the spatial and temporal information in SITS (Figure 27). One example is the Recurrent Residual Network (Re-ResNet) *(379)*, which maps temporal land cover changes over multiple seasons using Sentinel-2 imagery. In this model, the imagery for each date is processed through a 2D CNN (ResNet) to learn the spectral-spatial features, and the representation is then fed into a global average pooling before passing it to an LSTM cell connected to the next date in the sequence. The ARCNN architecture *(406)* uses a 2D CNN encoder composed of four conv layers and four deformable conv blocks to extract feature representations that capture the shape and scale variations within complex agricultural landscapes at each time point, which are then fed to bi-directional LSTM cells to learn temporal relationships. An attention mechanism re-weights the temporal features to mitigate noise before passing them to an FC layer with softmax activation to make class assignments. DuPLO architecture *(407)* combines CNN and RNN networks to take fused multi-temporal Sentinel-2 as input. A shallow 2D CNN branch processes small patches of 5x5, while the other branch uses a shallow CNN to generate a 1D aggregated feature sequence that is passed on to a GRU network. The learned features of each timepoint are then passed through a temporal attention, and the output feature vector is concatenated with the first branch. All branches are supervised through a compound loss with three cross-entropy terms.

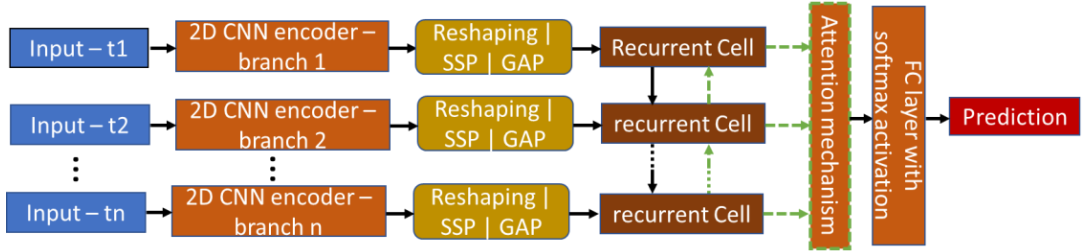

Figure 27. Generic structure of hybrid models integrating CNN and RNNs. Green dashed lines represent optional modifications including the bidirectional setting between RNN cells and attention head.

Conv-LSTM and conv-GRU architectures provide the opportunity for a simultaneous learning within the spatial and sequential dimensions, with input, output, and internal representations of a cell for each sample as 3D tensors. The weight tensors have the form $R^{k \times k \times c \times r}$ for input-to-state weights and $R^{k \times k \times r \times r}$ for state-to-state weights, where k is the convolution kernel size and c and r are the respective numbers of the input channels and the hidden state features (408). The internal matrix multiplication is replaced by a convolution operator in both the input-to-state and state-to-state transition functions of the cell, thereby serving to determine the state of a current cell for each spatial position in the image based on information from current input (e.g. sequence element) plus the prior state in each local neighborhood (409). These convolutional recurrent cells can be directly applied to the original SITS images or placed at the top of a CNN-based encoder and take deeper spectral-spatial features as input. For instance, Teyou et al. (346) uses U-Net as the spatial encoder for each timepoint, with convLSTM cells serving as recurrent layers. Teimouri et al. (410) added two conv-LSTM layers on top of FCN-8 to classify crop types using Sentinel-1 time series, structuring the inputs as mini-batches where a single temporal sample is expanded to a mini-batch with batch size equal to the number of timestamps. Chamorro et al. (399) developed an encoder-decoder architecture with a bi-directional conv-LSTM providing a bottleneck between the two stages. The encoder and decoder paths are both shallow, consisting of dense and pooling blocks that are downsampled by a factor of 4, followed by a transposed convolution for upsampling. Applied to Sentinel-1, this approach generates smoother crop predictions with less obvious salt and pepper artifacts than a bi-directional conv-LSTM.

Recurrent networks are also used to classify HSI imagery. For instance, Mei et al. (411) used a two-branch network to classify hyperspectral imagery. The first branch was a bidirectional GRU with an attention head on the spectral dimension that learned inner spectral correlations and weighted the importance of each band for classifying pixels. The second branch was a CNN with spatial attention applied to the HSI input, compressed using principle component analysis, to capture global spatial relationships.

## 6. Vision Transformers (VTs)

The transformer architecture was originally developed for neural machine translation, and has become popular in natural language processing due to its ability to process long sequences using self-attention mechanisms (412). More recently, the transformer has been applied to computer vision tasks, such as image generation (413, 414) and object detection (415–417). While CNN-based models have been very successful in semantic segmentation of satellite imagery, VTs offer several benefits that make them an attractive alternative for processing RS imagery. Unlike traditional CNN-based models that use fixed-size windows with restricted receptive fields, vision transformers process the input image in small, non-overlapping patches, allowing for more efficient processing of large images without losing resolution or alignment *(418)*, and the use of self-attention mechanisms enables VTs to better capture global dependencies, even in the early layers of the network *(419)*. VT models have better generalization performance than CNN-based models when trained on large datasets, particularly in scenarios where there is significant domain shift between the training and test datasets *(117)*. This is likely due to the structural design of VTs, especially the multi-head self-attention mechanism in each transformer layer in the encoder, which can model complex relationships between input features *(117)*. Naseer et al. *(420)* demonstrate that CNNs are have a tendency to make decisions based on texture while VTs are more tuned towards the object shapes. Dai et al. *(421)* shows that Transformers tend to have larger model capacity, although their generalization can be worse than CNNs when trained on limited data due to the lack of the right inductive bias[3]. Tuli et al. *(422)* shows that compared to CNNs, VTs tend to learn more from shape clues than texture in the learning process, and the model decisions are more consistent with the type of mistakes that the human operators tend to make when interpreting images. Raghu et al. *(423)* found that internal representation of VTs and CNNs differed for several image classification benchmark datasets, with greater similarity between representations in lower and higher network layers in VTs. They observed that CNNs, such as ResNet, required more layers to capture the same representation compared to VTs, which exhibit a mix of local and global relationships even in early layers. And Cordonnier et al. *(424)* showed that multi-head self-attention with sufficient number of heads and proper positional encoding has the representational capacity to express any function, including a CNN layer. However, transformers typically have higher computational and memory costs than CNNs, and require larger training datasets.

Pure Vision Transformers (VT) reformulate the visual task (e.g. classification/segmentation tasks) into a sequence-to-sequence mapping through tokenization and use positional encoding to introduce inductive bias. Through the process of tokenization (figure 28), the input satellite image of shape $\mathbb{R}^{H \times W \times C}$ is first divided into a grid of typically non-overlapping patches with the shape $\mathbb{R}^{N \times (p^2 C)}$, where each patch has a fixed size of $(p \times p)$, and $N = \frac{H \times W}{p^2}$ represents the number of generated patches. Decomposing the image into a larger number of patches typically contributes to higher prediction accuracy especially in semantic the segmentation task without an increase in the parameters, but with an increase in the computational cost as the length of the sequence fed to the attention mechanism increases *(425, 426)*. For semantic segmentation tasks, the patch size is usually fine-grained (e.g. 4x4 pixels) *(427)*. Each patch is flattened into a 1-dimensional vector, and goes through a linear projection (linear embedding layer) to make a vector of fixed size, which can be thought of as a K-dimensional feature space that represents the content of the patch in a compact form. Repeating this process for all the patches in the image results in a sequence of feature embedding vectors that get fed into the transformer encoder. The VT architecture for semantic segmentation is typically formulated as an encoder-decoder design and like CNN counterparts can use feature extractor VTs developed for image classification as their backbone. Each layer in the encoder typically consists of two main components. The first is a multi-head self-attention mechanism (MHSA) that allows the model to capture long-range relationships between different parts of the image (i.e. tokens), in order to compute a representation of the sequence. The second is a feedforward network (e.g. MLP) that applies non-linear transformations to the latent embeddings. A normalization layer (typically layer normalization (LN)), is placed either before or after each component (i.e. FFN, MHSA) with residual connection between the input and output of each component. There are other normalization choices besides LN. For example, Shao et al. *(428)* proposes a new normalizer called Dynamic Token Normalization (DTN) that can be used with various vision transformers, such as ViT, Swin, PVT to improve the performance. The authors found that because the LN only normalizes the embedding within each token, the magnitude of tokens at different positions becomes similar, which makes it difficult for VTs to capture inductive bias such as the positional context in an image. DTN performs normalization both within each token (intra-token) and across different tokens (inter-token), capturing both the global contextual information and the local positional context.

---

[3]Inductive bias in the context of artificial neural networks refers to the set of assumptions, constraints, or prior knowledge embedded within the architecture or learning algorithm, which guide the network's learning process. CNNs are specifically designed with a strong inductive bias for processing grid-like input data due to the local sliding window strategy with shared weights, which enables CNNs to leverage properties like local spatial coherence and translation invariance to effectively capture visual patterns.

As the attention mechanism in VTs is permutation-invariant, ignoring the ordering of the input sequence *(429)*, transformers also include a positional encoding (PE) layer to preserve the absolute and/or relative position of each patch within the sequence, which is necessary for position-dependent tasks like semantic segmentation *(93)*. Absolute positional encoding assigns a unique encoding to each of these tokens based on its position in the original image, resulting in a tensor of the same size as the image embedding and gets added to it before getting fed to the encoder. In contrast, relative positional encoding only considers the spatial relationships between each patch and its neighbors to generate a position encoding for each patch *(430)*.

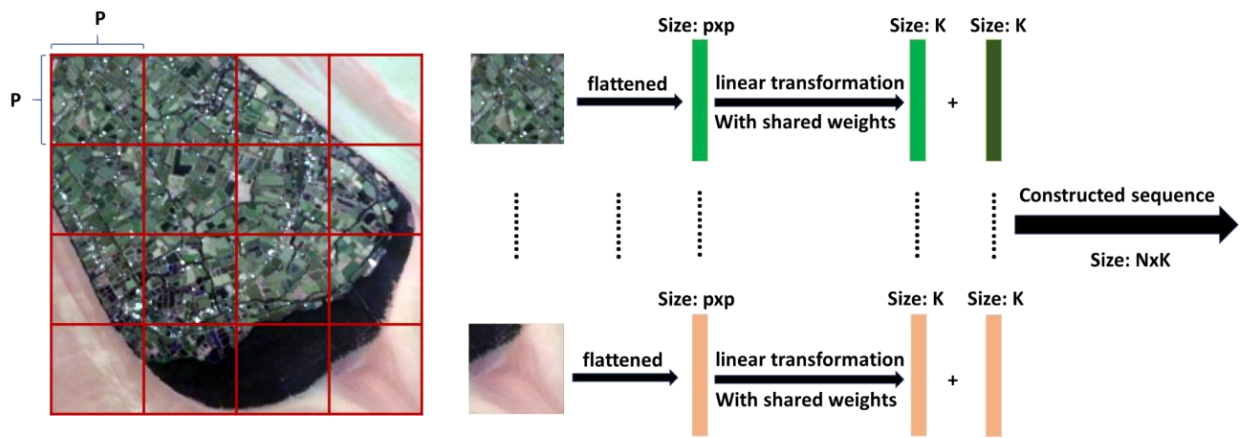

Figure 28. Patch embedding and positional encoding in ViT. In many modern VTs the positional encoding is implemented differently or dropped altogether but the patch embedding procedure is mostly intact.

## 6.1 Transformer backbones

The Vision Transformer (ViT) *(431)* is an example of a pure attention-based backbone with 12 to 32 layers developed for image classification tasks. In ViT, tokens along with their positional encodings make up the input tensor of shape $\mathbb{R}^{N \times K}$, which gets transformed to the shape $\mathbb{R}^{N \times D}$, where D is the dimensionality of the projected space. The MHSA module splits this input along the feature dimension to generate an H set of (Q, K, V) triplets, one for each attention head, with an equal number of features in each head. Each of the H heads has a separate set of learned parameters, and processes its input independently from other heads, enabling the MHSA module to capture a more diverse set of patterns within a global receptive field. Each head is a scaled self-attention mechanism[4] similar to the self-attention mechanism described in section 4.6. The output vectors from all heads are concatenated along the last dimension, resulting in an output tensor of shape $\mathbb{R}^{N \times (DH)}$. This output tensor is then passed through another linear transformation, which maps the DH-dimensional output space back to the original K dimensions. The resulting tensor is typically added to the input tensor via a residual connection, and then layer-normalized. The output is passed through the FFN component, typically consisting of two dense layers with a ReLU family activation (e.g. GELU) to fuse the features from different heads (Figure 29a). Both the number of heads and the number of features per head can substantially affect results; increasing the former improves expressivity of the layer, but reducing the latter has a negative impact on the accuracy *(432)*. ViT uses a learnable absolute PE with the same length as the input embedding.

There are a few concerns in adopting the ViT architecture especially in downstream semantic segmentation tasks. All layers in ViT maintain the same scale because the tokens, representing fixed-size patches, are of the same length across the model. Such a fixed scale, can cause issues in encoding objects that vary substantially in scale. Absolute PE can break the model's translation equivalence *(433)*, and requires resolution adjustment (e.g. resampling) if the images used for training and prediction have different sequence lengths *(431)*. The MHSA in ViT is global, as each head independently computes attention weights between each token and all other tokens in the input sequence. However, the computational complexity of global attention scales quadratically with image size, which can limit the scalability of the network. Hence, most of the research on designing vision transformer backbones are concerned with 1) developing hierarchical architectures to establish multi-scale feature representations, 2) reducing the memory and computational cost of global MHSA calculations, and 3) introducing appropriate inductive biases and prior knowledge of the target task for better performance. CNNs reduce computation by using pooling operations that decrease resolution as the network progresses. However, applying traditional pooling operations to vision transformers is not straightforward because the tokens are permutation invariant. Additionally, fixed down-sampling methods can discard important information or include redundant features that don't contribute to classification accuracy. It is shown that Vision transformers like ViT have highly sparse activations *(431)* which implies that only a subset of the model's parameters are active or contribute significantly during training, while others remain inactive or have negligible impact, which shows promise for pruning strategies to achieve good efficiency/accuracy trade-offs.

---

[4] To compute the global scaled self-attention for a given head, the module first calculates the attention scores between each query and key vector in that head which captures the relationship between different representations of the whole input subspaces at different positions (i.e tokens). This is done by taking the dot product between the query and key vectors and dividing the result by the square root of the dimensionality of the projected space. The resulting attention scores are then passed through a softmax function to obtain the attention weights for each key vector. The attention weights are then used to compute a weighted sum of the value vectors, giving a single output vector for each head.

A common approach is to remove the least important tokens or patches from the input image, which reduces the number of computations required during training/inference while maintaining high accuracy. Identifying the least important tokens is typically based on calculating their weights (i.e. importance scores) or their contribution to the output of the model *(434, 432)*. For instance, Fayyaz et al. *(435)*, leveraging the concept that the amount of pertinent information in an image fluctuates based on its content, developed a differentiable, parameter-free module called Adaptive Token Sampler (ATS) that efficiently reduces the number of tokens. ATS can be integrated into existing vision transformer models, either during the training phase or during inference, and is attached to the MHSA of a transformer layer at each stage. It uses the attention weights of the classification token to assign significance scores to the input tokens, and then applies inverse transform sampling over these scores to select a relevant subset of tokens. Adaptive Vision Transformer (AdaViT) *(436)* enhances the transformer backbone with a model-agnostic decision network. This decision network is added to every block of the transformer and is supervised by a specialized loss function designed to measure the computational cost of the generated usage policies, which are optimized in conjunction with the backbone. The decision network in AdaViT employs three distinct linear layers with sigmoid activation to dynamically determine the inclusion or exclusion of self-attention heads, transformer blocks, and patch embeddings based on the input to each block.

Another approach for reducing the number of tokens is to cluster similar tokens based on the spatial redundancy in intermediate token representations that arises in well-trained ViTs, which reduces the number of tokens without reducing the receptive field. Token clustering entails computing pairwise distances between the tokens using a similarity metric (e.g. Euclidean distance), and then grouping proximate tokens using clustering algorithms such as K-means or hierarchical clustering techniques. Once the tokens have been grouped, the centroid token of each cluster is then selected as a proxy for all tokens in the cluster. These representative tokens are then used as inputs for the subsequent layers of the model. For instance, Liang et al. *(437)* converts the "high-resolution" path of the vision transformer to a "high-to-low-to-high resolution" path using a token clustering layer and a token reconstruction layer, both of which are non-parametric and unlike resampling strategy don't require fine-tuning to avoid accuracy degradation. The token reconstruction layer maps the low-resolution clustered representations back to the high-resolution space, thereby recovering the original image resolution.

The Pyramid Vision Transformer (PVT) *(438)* resolves the drawback of fixed scale in ViT by introducing a hierarchical architecture that uses a progressive shrinking strategy, in which patch embedding layers are added to each encoder stage to control the scale of feature maps. PVT uses a learnable absolute PE, which is calculated alongside the patch embedding at each stage of the encoder. PVT also introduces Spatial Reduction Attention (SRA) as a modification to the global MHSA, which reduces the spatial dimensions of inputs to the K and Q branches based on a user-defined ratio in order to lower the memory and computational overhead.

In contrast to absolute PE, relative PE only considers the spatial relationships between each patch and its neighbors to generate a position encoding for each patch (430) and is usually included in the encoder layers instead of being embedded in the input image. For instance, Conditional PVT (CPVT) (433) introduces a generator module consisting of padded 2D convolutions, placed after each encoder layer in each stage of the PVT, to calculate the conditional PE from the layer's output, then adding it back to the embeddings before passing to the next layer. The generator module dynamically develops positional information conditioned on the local neighborhood of the input tokens, making it possible to generalize to inputs of different lengths. PVT v2 (439) introduces several key improvements to enhance the performance of the first version. Unlike PVT v1, which treats an image as a sequence of non-overlapping patches, PVT v2 applies overlapping patch embedding using convolutions with zero-padding to maintain local continuity in the image. The second improvement is a linear complexity MHSA, which is achieved by using average pooling to reduce the spatial dimensions of the features to a fixed size before passing it to the attention module. The third improvement replaces the fixed-size position encoding in PVT v1 with conditional position encodings by adding a padded depthwise separable convolution layer after the first dense layer and before the GELU non-linearity in the FFN component, making it easier to process images of arbitrary resolution.

The Swin Transformer (440) is another hierarchical design that consists of a number of transformer layers, which reduce the sequence lengths by combining neighboring patches with a patch merging layer that is added after each stage, creating a multi-scale encoder. Swin follows a local MHSA by restricting each head to a local neighborhood of patches, which is known as windowed MHSA (W-MHSA). To keep the computational complexity linear while allowing cross-window token mixing, it shifts the window within a non-overlapping pattern over the patches between consecutive layers (SW-MHSA, Figure 29b). Similar to the adjacency matrix in spatial analysis, the shifted window mechanism is represented as a matrix of binary values, where each row and column correspond to a patch in the image, with values of 1 indicating which patches can pay attention to each other. The pattern of the shifted window mechanism can be adjusted by changing the stride and shift of the window, the sizes of which vary for different stages of the Swin Transformer. Swin Transformer V2 (441) introduces two adjustments to the architecture that help improve the network's scalability. These include placing the LN layer after the MHSA and MLP components, which keeps the activation amplitude between transformer layers consistent throughout the network. Scaled cosine similarity is also used to make self-attention calculations less dependent on the amplitudes of block inputs. Both versions of the Swin transformer use a learnable relative PE that is added to the similarity matrix of each head in both the W- and SW-MHSA.

Shao et al. (428) introduced a novel normalization technique called Dynamic Token Normalization (DTN) to enhance the performance of various vision transformers, including ViT, Swin, and PVT. The authors identified a limitation in Layer Normalization (LN), which only normalizes the embedding within each token. Consequently, LN fails to maintain distinct magnitudes for tokens at different positions, hindering the Transformers' ability to effectively capture important positional context in images. In contrast, DTN applies normalization both within each token (intra-token) and across different tokens (inter-token), enabling Transformers to capture both the global contextual information and the local positional context more effectively. This approach addresses the challenge of preserving inductive bias by keeping the variation between tokens without the need to use convolutions.

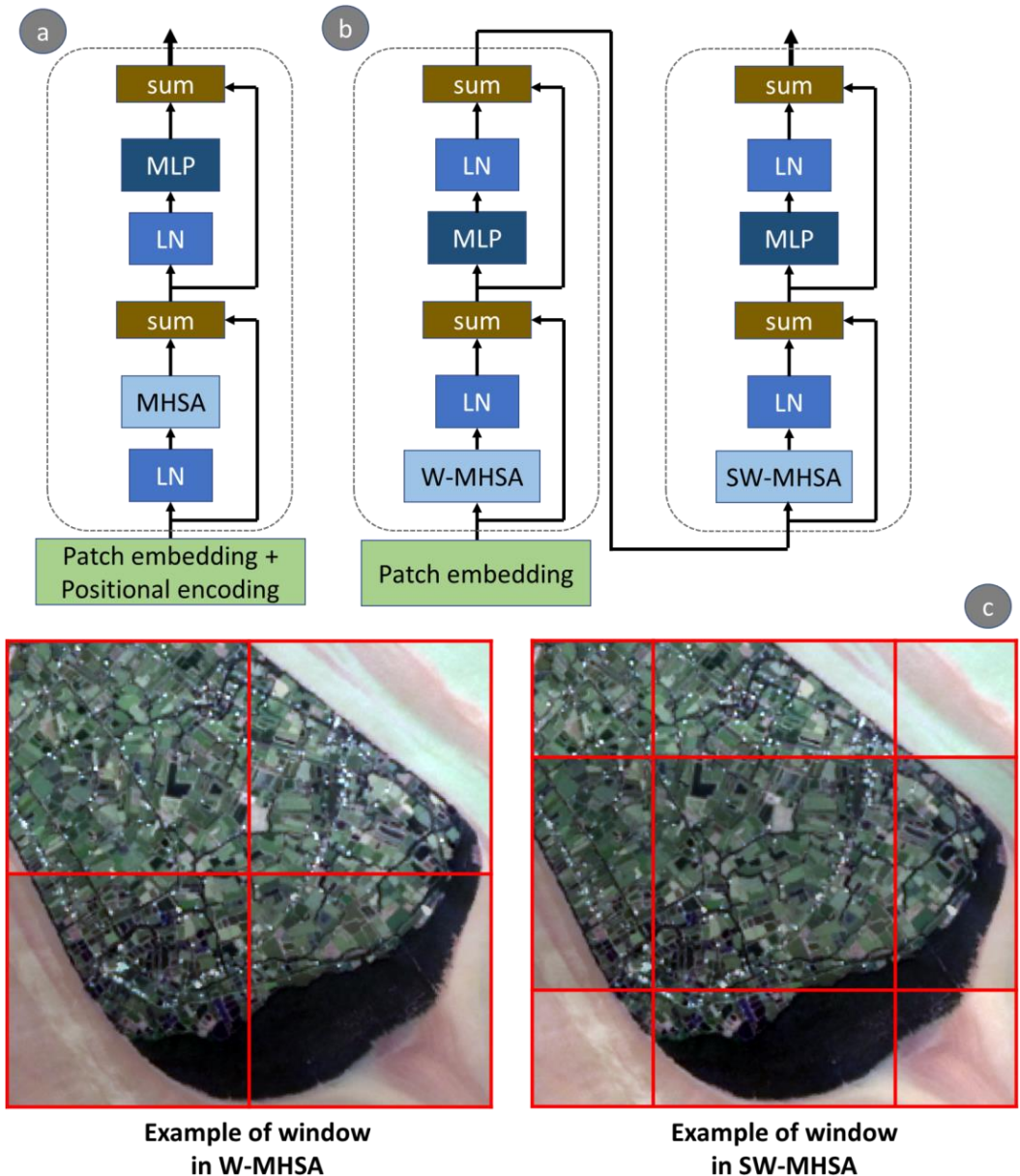

Figure 29. The encoder layer in a) ViT and b) Swin transformer V2. In each stage of the Swin encoder a window multi-head self-attention (W-MHSA) is followed by a shifted window MHSA (SW-MHSA) in the consequent layer, and this pattern repeats C) shows an example of how the windows change in consequent Swin layers.

CSWin Transformer (442) uses a cross-shaped window in the self-attention mechanism that splits heads into groups that separately process the horizontal and vertical dimensions of the image using stripes of equal width in parallel, which allows the model to capture both local and global features. Strip width is typically small for shallow layers and becomes larger for deeper layers in the network. In CSWin, a convolution layer is used between stages, which halves the number of tokens while doubling the number of feature channels. Different from Swin, CSWin generates a relative PE by transforming the V branch through convolution and adding it directly to the output of the MHSA, effectively decoupling the positional encoding from the attention calculations.

Twins *(443)* introduces two sequentially arranged attention modules that replace the shifted window mechanism used in the Swin architecture with two attention mechanisms in the same layer. The first of these is the locally-grouped self-attention (LSA), a window-based MHSA that divides the input sequence into multiple groups, in order to capture local within-group patterns. The second is global sub-sampled attention (GSA), which identifies relationships between groups using a single representative token computed from each group. Twin also uses conditional PE.

## 6.2 Transformer Decoder

Semantic-rich sequences of latent tokens produced by the encoder are reshaped to image patches and get upsampled to the original spatial dimensions. SEgmentation TRansformer (SETR) *(444)* is among the first designs with the semantic segmentation task in mind. The decoder structure in the SETR model is responsible for taking the features extracted by ViT as the encoder and producing the final pixel-wise segmentation mask. The encoder output is 16X coarser in the spatial scale than the original image. SETR introduces three decoder heads. The first is a naive upsampling decoder, which consists of a classifier module that uses a 1 × 1 conv, followed by a sync batch norm layer with ReLU activation and another linear 1 × 1 conv layer, which map the features to the category space, and then bilinearly upsamples the output to the full image resolution. The second is a progressive upsampling (PUP) module, which alternates convolutional layers with 2x upsampling operations to reach the full resolution, in order to mitigate the noisy predictions that often result from one-step upsampling. The third decoder head is Multi-Level feature Aggregation (MLA), which takes feature representations from M layers distributed uniformly across the encoder and processes them separately through a set of conv layers. The processed features from all layers are then upsampled 4x and fused via top-down aggregation and channel-wise concatenation, resulting in a final feature map with the same resolution as the input. Segmenter *(425)* also uses ViT as a backbone, but uses a masked decoder that contains two transformer layers. These layers take a set of K trainable class embeddings for each semantic class as input, along with the patch embedding produced by the encoder, and calculate the scalar product between the class embeddings and L2-normed patch embedding to generate K class masks. These masks are then reshaped to 2D images, upsampled bilinearly, and passed through a softmax and layer norm to produce the final prediction map. SegFormer *(427)* leverages the large receptive field of the transformer encoder and uses an MLP decoder that accepts patch embeddings from all encoder stages, projecting them to the same embedding dimension. It then up-samples each patch embedding to ¼ the original resolution, concatenates and fuses them through a 1×1 convolution and normalization. Finally, another 1×1 convolution maps the embeddings to the label space (figure 30). Dense Prediction Transformers (DPT) *(445)* uses ViT as its backbone but leverages a convolution-based decoder to combine the benefits of transformers with the spatial understanding of convolutions for robust dense prediction tasks.  First, the read operation handles the classification token containing the image-level information by simply ignoring it or adding/concatenating it to every other N tokens of size D. Following the 'Read' operation, the reshaped tokens are transformed back into an image-like representation by mapping each token to the position of its corresponding patch in the image, a process referred to as Concatenate by authors but mostly known as un-patchifying in the literature. Lastly, the 'Resample' operation rescales this representation to a specific size with a certain number of features per pixel, using a 1 x 1 convolutions feature dimension adjustment, followed by a 3 x 3 strided or transpose convolution, depending on the required scaling.

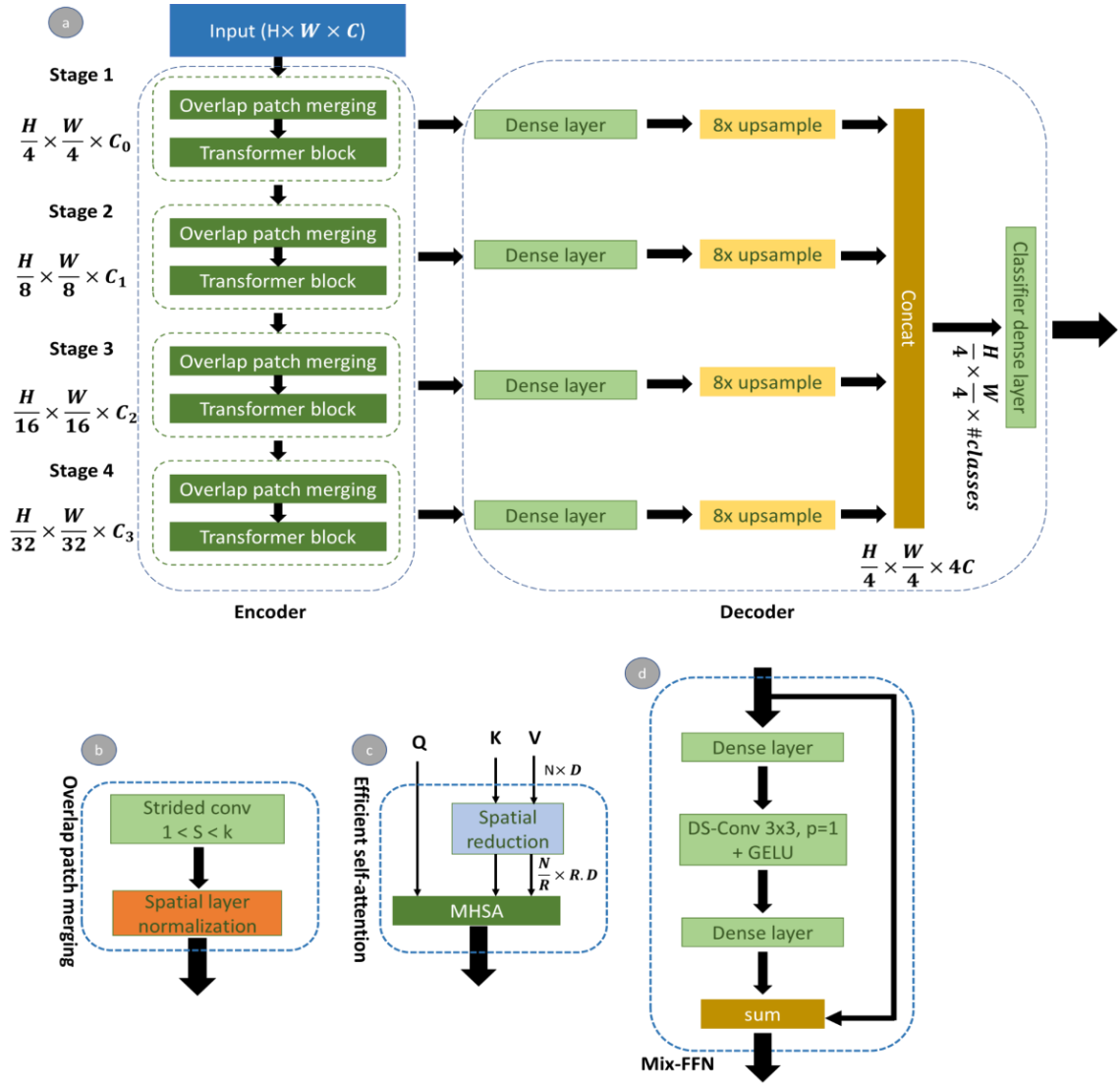

Figure 30. Structure of SegFormer. a) overall structure, b) patch merging strategy, c) efficient self-attention using the spatial reduction technique reducing the complexity to $O(\frac{N}{R})^2$ and d) is the feed forward network that uses conv layer with padding as substitute for PE.

### 6.3 Hybrid models

There are also hybrid models that combine CNN and transformer architectures to improve contextual modeling and feature extraction in semantic segmentation *(446, 447)*. The design patterns of such hybrids are diverse, including merging the layers in a single module like those that integrate convolution with MHSA in a single transformer block, as well as models that separate the convolution and transformer blocks, either within a single branch or in separate parallel branches. For instance, TrSeg *(448)* is a hybrid encoder-decoder architecture that has three main components: A Dilated Residual Network (DRN) backbone CNN, a multi-scale pyramid pooling module (PPM), and a transformer decoder subnetwork. The DRN backbone CNN extracts informative context from input sequences, and the multi-scale pooling module produces contextual information that is flattened and concatenated to a single sequence that is fed to the transformer. The transformer decoder subnetwork is a stack of six identical transformer decoder blocks, each of which contains an MHSA block (with 8 heads), a co-attention block, and a Feed-Forward Network (FFN) made up of two linear transformation layers, LayerNorm and ReLU activation. The MHSA acts on the PPM output, while the co-attention module uses the original encoder output to generate K and V branches, and the PPM output to make the Q branch, producing a gradual upsampling with robust multi-scale feature fusion. CoAtNet *(421)* uses a single branch design with four stages, in which the first two stages consist of conv blocks based on the inverted bottleneck depthwise convolution layers from Mobilenet V2 (MBConv), which was chosen because convolutions were found to be more effective for processing primitive patterns. The last two stages are transformer blocks from ViT that were modified to obtain relative positional encoding using a global convolutional kernel, which is added to the similarity matrix before softmax normalization, thereby producing translation equivalence. The Convolutional vision Transformer (CvT) *(449)* uses a three stage network consisting of improved ViT blocks with integrated convolution and attention. Each stage first uses token embedding that uses depthwise convolution to reshape tokens into 2D grids, using overlapping windows followed by layer normalization. Convolutional token embedding enables the network to learn a hierarchy of features by progressively reducing the number of tokens while increasing the feature dimension, similar to CNNs. The flattened output then goes through another depthwise separable conv layer, instead of the conventional linear projection, to create the Q, K, V triplet for the MHSA. This further increases the model's ability to encode inductive bias without sacrificing the global receptive field, which removes the need for a positional encoding layer, thereby enabling different sized images to be used for training and inference. This architecture further improves the computational efficiency of MHSA by compressing the embedding of K and V branches using a strided convolution projection layer. Similarly, SegFormer *(427)* uses an efficient hierarchical transformer encode that produces multi-level features using overlapped patch merging at the end of each stage of the encoder. This approach is implemented using 2D convolution with a carefully decided kernel size, stride, and padding (e.g. K = 7, S = 4, P = 3 in the first stage, and K = 3, S = 2, P = 1 for the other three stages in the network). This merging strategy forces the MHSA to focus on local patterns in early stages while gradually increasing the ERF. This optimized MHSA reduces computational cost by reducing tokens along the K branch. SegFormer avoids using PE by introducing positional information by adding a padded depth-wise conv layer in the FNN component after the first dense layer, before applying non-linearity. D'Ascoli et al. *(450)* introduce a modified MHSA known as gated positional self-attention (GPSA), designed with a soft convolutional inductive bias. The GPSA layers, initially mimicking the locality of convolutional layers, allow each attention head to adjust a gating parameter that regulates the balance between positional and content information. This is implemented through an initialization procedure that can imitate a convolution with a square kernel size equal to the square root of the number of heads. The researchers created the Convolutional Vision Transformer (ConViT) by replacing the MHSA in the first 10 blocks of ViT backbone in the Data-Efficient Image Transformer (DeiT) model augmented with GPSA layers which outperforms the DeiT, particularly in scenarios with limited data. HRViT (451) is a hybrid model, merging multi-branch architectures like HRNet with transformer layers for semantic segmentation. Starting with a strided convolution stem for initial feature extraction and spatial dimension reduction, the model contains four progressive Transformer stages, each with multi-scale branches. All stages begin with lightweight fusion layers, merging low- and high-resolution features to retain detailed positional data. Each stage also contains a patch embedding block for interpatch communication through a simplified pointwise and depthwise convolution, local self-attention blocks (HRViTAttn), and mixed-scale convolutional feedforward networks (MixCFN). The inputs for these stages are split into horizontal and vertical windows and processed individually, optimizing computational efficiency and focusing on local spatial relationships. Each stage concludes with a MixCFN, an inverted residual block consists of two multi-scale depthwise convolution paths inserted between two linear layers, leveraging 3×3 and 5×5 kernels to enhance the multi-scale local information extraction.

## 6.4 Transformers for SITS

TSViT (452) is tailored for SITS modeling, leveraging a factorized temporo-spatial adaptation of the ViT as its backbone. This design ensures a global receptive field in every layer across both time and space. Instead of the traditional ViT approach with a single cls token, TSViT utilizes multiple cls tokens corresponding to the number of object classes. These tokens refine the accumulated data from patch tokens, thereby enhancing the capacity for storing global representation and subsequently yielding more precise class probabilities. Furthermore, TSViT implements a distinct tokenization strategy for input image time-series, combined with temporal position encodings matched to acquisition times. This strategy emphasizes the extraction of date-specific features while compensating for irregular SITS acquisition intervals. After undergoing the temporal encoder, cls tokens are distinctly separated by class, allowing interactions solely within identical cls tokens in the spatial encoder, ensuring a class-centric focus. TSViT employs 3D convolution with spatial and temporal strides for non-overlapping token extraction, capturing comprehensive spatio-temporal data from the input SITS. Emphasizing its distinction from natural image processing, TSViT suggests that a temporal-then-spatial factorization is more aligned with the nuances of SITS processing. Zhao et al. (453) used a transformer model to classify each pixel of the VIIRS active fire product which has a low spatial resolution of 375m but leveraging a high temporal resolution. The tokenization process involves extracting a small patch of size w centered on every pixel position in the input. These patches are then flattened to create vectors of size $Cw^2$ for each time point, and the resulting sequence is fed into the transformer encoder. MHSA is also modified by applying a mask to each pair of Q and K vectors to limit the model's access only to the past and current tokens for classifying each time point.

There are also transformer-like networks designed to work on single-pixel datasets popular in crop type mapping where the spectral-temporal patterns are thought to be more helpful than the spatial dimension as the shape prior in this case is arbitrary. For example, Garnot et al. (454) propose a spatio-temporal framework known as the Pixel-Set Encoder (PSE) for processing medium-resolution Satellite Image Time Series (SITS) in crop type mapping. The model represents each parcel using a spatio-spectra-temporal tensor, where it randomly selects a set of S pixels from the N pixels in the parcel. Each selected pixel is then processed by a shared Multilayer Perceptron (MLP) module that includes a fully-connected layer, batch normalization, and a ReLU activation function. The processed values are pooled along the pixel axis, resulting in a vector encapsulating the statistical features of the entire parcel. This vector is complemented with pre-computed geometric features of the parcel such as the perimeter, pixel count, cover ratio, and the ratio between the perimeter and surface and passed through another MLP module (MLP2). Furthermore, the model adopts a fixed sinusoidal positional encoding, inspired by Vaswani et al. (2017). However, unlike the original Transformer network which outputs for each sequence element, the goal here is to encode an entire time series into a single embedding. To achieve this, a master query is defined for each attention head, calculated as a temporal average of the queries from all dates. This query is used to determine a single attention mask, which weights the input sequence of embeddings. The outputs from each of the H attention heads are concatenated and processed by another MLP (MLP3) to obtain the final output of the Temporal Attention Encoder (TAE). Unlike the Transformer network, this output is used directly as the spatio-temporal embedding without the need for residual connections. The model shows that each attention head specializes on a specific portion of the time series, indicating its ability to adapt to different temporal patterns in the data. Authors, also introduced a lightweight version (L-TAE) modified by eliminating redundant calculations and parameters (455).

## 7. Factors to consider for model training and prediction

The semantic segmentation of Earth Observation data using deep learning models requires careful consideration of several factors that influence their performance and generalization ability. Some of the critical factors include the size, quality, and representativeness of the samples used, as well as the relative abundance of categories. Previous studies by Razavian et al., Helber et al., Blaga and Nedevschi, and Elmes et al. (456–459) have highlighted the importance of these factors. Therefore, this section provides an in-depth analysis of the concerns regarding training data and the techniques developed to address these issues.

### 7.1 Common pre-processing

When preparing remote sensing imagery for DL model training, it often needs to be divided into smaller tiles or chips, as the large size of most images cannot be accommodated in the memory available on most hardware (55). Choosing the appropriate chip size involves considering several factors. First, the spatial dimensions of the image chips should be large enough to provide sufficient context to capture the patterns of interest. For example, in a study on tree species mapping using UAV-based RGB imagery, Schiefer et al. (460) found that chip size did not affect the overall model performance, but had noticeable impacts on specific classes due to within-chip class imbalances that varied with chip size. Second, to simplify implementation with hierarchical encoders, it is good practice to make the chip size divisible by the total downsampling factor of the feature extraction subnetwork (i.e. backbone network). Chipping is usually implemented with an overlap strategy set between 20-50% to increase the number of chips (292, 461, 462). There are also chipping schemes that allow to define tiles in units of the raster image CRS which works best on datasets containing multi-sensor with different GSD to have consistent context (463). Chipping is also necessary for mini-batch training, a commonly used strategy that involves randomly selecting a batch of samples without replacement from all available chips. The mini-batch size must be large and representative in order to generate stable loss curves as the network weight update at the end of each mini-batch depends on the average loss of the mini-batch. A larger batch size is also required if the network uses batch normalization (BN) layers, which depend on the mini-batch driven statistics. As a heuristic rule, the patch size can be determined based on the effective receptive field (RF) of the last layer of the backbone, which is typically the largest effective RF of the network. This is also confirmed by Pintea et al. (142), who found the patch size and convolution kernel size as important factors in defining the lower bound of the resolution dimension, which is the inner scale used to observe objects in an image.

Chip size also affects the quality of predictions during model inference (464). Increasing chip size can provide more context around objects, reducing edge effects that can degrade prediction quality. Reina et al. (153) demonstrated that incrementally increasing the prediction chip size up to half of the max tile size of 640×640 improved the dice metric in the SpaceNet-Vegas dataset. Similarly, EfficientNet (162) showed that increasing patch size can boost prediction accuracy up to a certain point depending on the dataset and model architecture, beyond which the accuracy declines when chip sizes become very large with irrelevant content.

Another issue related to chipping is the occurrence of checkerboard patterns at the boundaries of stitched chips, which is attributable to the interaction between the convolutional operation and the spatial distribution of objects in each image chip. Predictions for objects near the edges of a chip are usually less certain than for those near the center, where there are contextual clues available in all directions to aid discrimination. To minimize these artifacts, a common approach is to provide an overlap between adjacent chips and then combine predictions in the overlap zone by averaging (295, 301) or a voting strategy (292) to decide on the prediction in that area. When averaging, it is more accurate to convert score maps to binary predictions for a given class first (153). Another way to reduce checkerboarding is to make image chips larger than the actual size required for prediction output, effectively adding a buffer around each chip that overlaps with neighboring chips. This overlap serves to provide the necessary spatial context to the model and is discarded after prediction to enable stitching the central part of chips for a seamless prediction (465).

Rescaling is a common pre-processing step applied to input imagery that aims to normalize the brightness values across different images to improve model performance. There are a variety of rescaling methods but the most common are: standardization and min/max rescaling. Standardization involves converting brightness values to z scores (466), while min/max linearly transforms the original range of brightness values, typically to the range between 0 and 1, using the minimum and maximum values in the image (467). A comparative study on seagrass mapping using drone optical images demonstrated that min/max rescaling provides better results for a variety of network architectures compared to standardization (468). Both techniques are usually applied separately to each band in multispectral imagery to prevent any single band from dominating the calculated statistics. Another important choice is to calculate the statistics locally for each image chip, or to do the calculation over the whole dataset. Such a choice is particularly important for SITS where per-band min/max rescaling is preferable, when the statistics are calculated across the full temporal extent of the time series *(363)*. To prevent outliers from affecting the range statistics, usually the 2nd and 98th percentile values are used as the min/max values.

## 7.2 Limited training data

DL-based models require a large number of training samples to achieve satisfactory generalization performance *(469)*. Unfortunately, collecting a large amount of annotated RS imagery can be expensive, especially when the scenes are complex and contain numerous objects with ambiguous boundaries. Given the costs, a number of techniques have been developed that help to increase the size of training and validation datasets, while minimizing the overall collection effort. These include approaches that leverage un- or partially labeled data, those that generate synthetic labels, and transfer learning methods.

### 7.2.1 Augmentation techniques

One approach to overcome limited training data is to augment existing labels, which involves applying geometric or photometric transformations to the image chips to increase their variability and boost the sample size. Geometric transformations, such as the affine family, alter the position and orientation of objects in the chips while preserving topology *(470)*. These augmented chips can then be used as additional samples added to the originals or as on-the-fly transformations of the original chips. Common geometric transformations include translation, reflection, cropping, scaling and rotation. Photometric transformations, which include brightness shift, gamma correction, blurring, random contrast change, or added random noise, are applied to pixel values over the channel dimension to make the model robust to changes in color or contrast *(471, 472)* by forcing the model to rely more on the shape information textural patterns rather than shape information in categorizing objects *(422)*. For 1D time-series data, popular augmentation approaches include temporal warping, adding noise, and brightness jittering. In temporal warping, the temporal sequence of the original time series data is randomly stretched or compressed in time by applying a random non-linear function that can change the temporal spacing between the data points. This results in an artificially augmented dataset with variations in the temporal scale, while preserving the temporal dynamics of the original dataset. Beside enlarging the dataset, augmentations have a regularization effect, as they make the model invariant to those transformations, thereby providing a better generalization capability *(471)*.

Most augmentation methods, especially brightness transformations, designed for ground-view natural images and might not be as effective for medium resolution, multi- or hyper-spectral overhead imagery, where spectral channels are carefully calibrated, and the between-band correlations usually contain rich discriminative information with analytical justifications. Illarionova et al. *(473)* developed MixChannel Augmentation, which chooses one time point as the anchor image and replaces some of its bands with the equivalent bands from other non-anchor images, preserving the spectral distribution of each separate original band within the synthesized band, without strictly maintaining the joint distribution over all bands within the augmented image. CutMix *(474)* is an augmentation method where two images from the training dataset are merged to create a new sample. This fusion is accomplished by applying a binary mask that determines which regions of each image should be combined, effectively exchanging patches between the two images. The corresponding labels are also mixed proportionally to the area of the patches. This technique maximizes the use of training pixels while maintaining the regional dropout's regularization effect. As a result, CutMix enhances the model's object localization capabilities and overall performance. Similarly, chessMix *(475)* is an augmentation technique designed for the RS imagery, which generates synthesized images by arranging the geometrically transformed image chips into a gridding pattern, using a cut and mix strategy in a way that corrects for class imbalance. The gridding structure also has blanks in between the actual image patches to avoid adjacent discontinuation problems and to force the model to learn more descriptive features. This approach was particularly promising for small, heavily imbalanced datasets *(475)*. Kemker et al. *(46)* generate large quantities of synthesized multi-spectral imagery using DIRSIG software package *(476)* to boost the volume of their training dataset, and found improved accuracy for all models trained using the augmented dataset versus the original dataset. Nalepa et al. *(477)* introduced a rapid PCA-based augmentation for Hyper-spectral RS images, which simulates data that match the distribution of the original dataset, and can be applied for both model training and testing.

Advanced data augmentation techniques harness the potential of Generative Adversarial Networks (GANs) *(478)* to synthesize new images by approximating the intrinsic distribution within the original dataset, using a parametrized model distribution through a competitive training process. This creation of additional images via GANs amplifies the sample size and diversity of the training dataset, thereby improving model generalization and performance *(479–482)*. Conditional Generative Adversarial Networks (cGANs) *(483)* stand out among various GAN architectures due to their ability to generate samples under specific conditions. Fundamentally, a cGANs consists of two interconnected components: a generator and a discriminator. The generator ingests random noise along with certain conditions (such as class labels, other images, or raster layers like DEM) to fabricate new samples tailored to those conditions. While it's typical to also provide these conditions to the discriminator to enhance its evaluative prowess, it's possible to only condition the generator, as evidenced in works like Baier et al. *(484)*. By feeding the conditions to the discriminator, it's empowered to not just gauge the realism of the generator's output, but also its adherence to the imposed conditions with the cost of more complex training. GANs training is originally formulated as a two-player zero-sum game represented as a min-max loss function which consists of two parts, used to train the generator and discriminator networks concurrently, whereby the problem is a maximization task for the discriminator and a minimization task for the generator *(478)*. The generator loss is designed to encourage the generator to produce samples that are indistinguishable from real samples and in line with the desired conditional vectors. It's defined as the negative log-likelihood of the discriminator's output when fed with the generated samples, thus the generator tries to minimize the log-probability that the discriminator correctly identifies the generated sample as fake. The discriminator loss is aimed at enabling the discriminator to correctly differentiate between real and fake samples, given the corresponding conditional vectors. The discriminator loss is the sum of two terms: the log-probability of correctly classifying real samples and the log-probability of incorrectly classifying fake samples. cGANs often implement a Spatially-Adaptive Denormalization (SPADE) normalization layer *(485)*, which adjusts the normalization statistics of the generator network based on the semantic layout of the input image, thus enhancing the quality of the generated images. For instance, Baier et al. (484) use GAN conditioned on a land cover map and DEM layers to generate synthesized SAR images, and use the semantic map and SAR images, while the semantic map and SAR images were used to generate visible band composite images for both 10 and 1 meter resolutions. Similarly, Le et al. (486) used a cGAN with an encoder that produces mean and variance of real 4-bands satellite imagery, a generator consisting of SPADE ResNet blocks conditioned on the class mask the encoder output, and a patch-GAN discriminator subnetwork. Using a subset of the Chesapeake Land Cover dataset for the segmentation task, they found that a Unet trained on this synthesized data reached the same level of performance compared to the same model trained on real imagery.

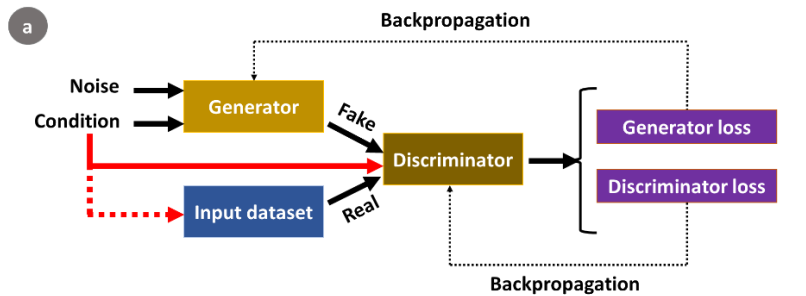

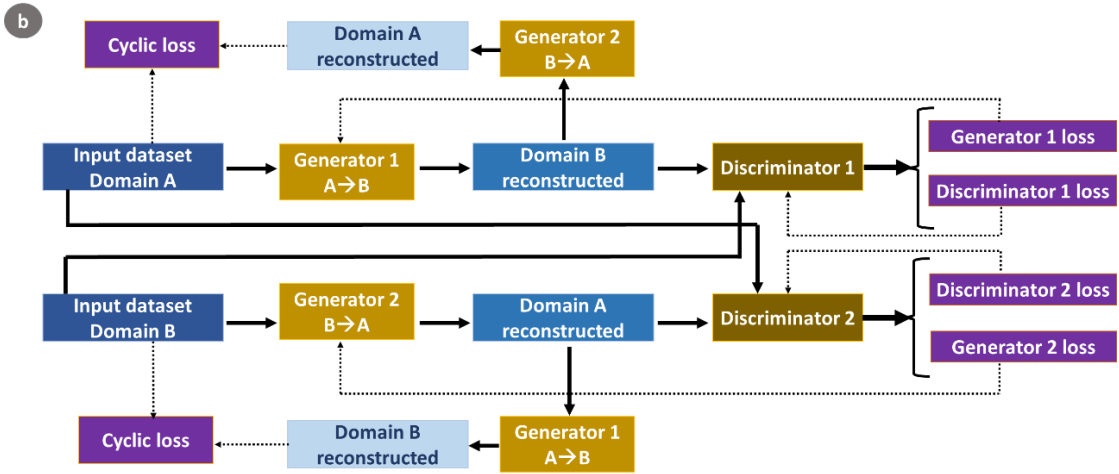

Figure 31. General architecture of a) cGAN and b) CycleGAN. Notice that conditioning the discriminator in cGAN can be achived in two primary ways. Either through direct conditioning (red arrow) where the discriminator evaluates the authenticity of a sample given its associated condition. Or fusion with real

dataset (dashed red arrow) where the discriminator isn't explicitly conditioned in the traditional sense; instead, it evaluates the authenticity of the combined sample.

In an ideal scenario, GAN models converge when the generator produces outputs that mirror the real data so closely that the discriminator cannot differentiate between them. As highlighted by Goodfellow et al. *(478)*, a discriminator trained to perfection drives the generator to minimize the Jensen-Shannon (JS) divergence to measure the similarity between real data and the generator's output. However, there may be complications when the actual data and the generator's output occupy separate lower-dimensional manifolds within the larger space of potential outputs. In such instances, these entities might not overlap or intersect only at isolated points *(487)*, causing JS divergence to be constant and KL-divergences (and their reverse counterparts) to reach infinity. These situations render divergence measures unproductive for training the generator as they fail to provide a helpful gradient for the generator's learning *(488)*.

To address these issues, Wasserstein GANs (WGANs) were proposed by Arjovsky et al. *(487)*. WGANs substitute the original GAN minimax objective function with the Wasserstein (Earth Mover's) distance which smoothly measures the amount of probability mass to be moved to transform one distribution into another, providing more stable and meaningful gradients throughout the learning process. WGANs also introduce a Lipschitz constraint, limiting the extent of changes the discriminator (referred to as a "critic" in WGAN) can make in response to input alterations. This is achieved by weight clipping or gradient penalty, ensuring the critic approximates a 1-Lipschitz function, which is crucial for an accurate estimation of the Wasserstein distance. Unlike traditional GAN discriminators that output a probability between 0 and 1, the critic in a WGAN can output any real value, with higher values for real samples and lower for generated ones, indicating the estimated Wasserstein distance. There are also variants that relax the Lipschitz constraint to further ease the training process *(489)*. Notably, WGANs and their variants, such as the gradient penalty WGAN *(490)*, are more robust to mode collapse, a scenario where the generator tends to produce limited output diversity, failing to capture the entire distribution of the training data. For example, in satellite imagery, a generator might only produce urban landscapes, overlooking other types like forests or agricultural fields. This lack of diversity hampers the GAN's utility for tasks like semantic segmentation. Moreover, monitoring the training and validation losses to avoid overfitting is not as straightforward with GANs compared to traditional deep networks, as the two loss functions are optimized for different goals and do not converge like traditional loss functions used in dense CNNs *(475)*.

Evaluating the performance and quality of GANs can be a complex task, as it involves assessing diverse, and sometimes, opposing aspects like image diversity, realism, and structural consistency. There is no universally accepted metric that captures all these characteristics in a unified manner, hence a combination of objective and subjective measures are usually employed. In scenarios where abundant real samples are available, the quality of synthesized images can be gauged by comparing the accuracy of a model trained solely on the synthesized images with an identical model trained on real samples as a measure of how close the synthesized images are to the actual data distribution *(484)*. In addition to the aforementioned approach, Inception Score (IS) *(491)* and Fréchet Inception Distance (FID) score *(492)* are two popularly used objective metrics. The IS uses a pre-trained Inception network, originally trained on ImageNet, to measure the quality and diversity of the generated images. The IS essentially measures two desired properties in the generated samples: their ability to be easily classified (indicating good sample quality) and their diversity with respect to class labels. The Inception Score computation involves calculating the average KL-divergence between two specific distributions: the conditional label distribution $p(y|x)$, and the marginal distribution $p(y)$. The conditional label distribution, $p(y|x)$, is the predicted class probabilities for a generated image, x. The marginal distribution, $p(y)$, on the other hand, is the overall distribution of predicted class labels over all generated images. A high Inception Score implies that the generated images are of good quality and exhibit diversity as it favors situations where the conditional entropy is low (indicating that the images are easily classifiable), and the marginal entropy is high (indicating diversity across the generated images). IS doesn't directly evaluate diversity in the sense of checking if the generated samples cover all modes (variety in the data) of the real data distribution, but is based on the assumption that a model generating diverse outputs would lead to a more uniformly distributed class probability in the pre-trained Inception network. Therefore, a GAN can memorize the training set, simply reproducing the training images, and still achieve a high IS if the training set is diverse and covers a wide range of classes. The FID metric, proposed as an enhancement to the IS, which considers the characteristics of image features, rather than the images themselves and directly compares the distribution of generated and real images. FID uses a specific layer of the Inception network (or any other suitable CNN) to embed the generated and real images into a high-dimensional feature space which is assumed to follow a multivariate Gaussian distribution and computes the mean and covariance as the key characteristics of the distributions of image features. The FID is then computed as the Fréchet distance (a measure of similarity between two distributions) between the Gaussian distributions of the real and the generated image features. A lower FID score corresponds to more realistic and high-quality image generation. It has been shown that FID is more robust than IS with respect to variations in dataset size and complexity, making it a more reliable measure in many scenarios especially if fine-tuned on the target domain *(493)*. Unlike IS, FID is also capable of detecting intra-class mode dropping - a situation where a model generates only one type of image per class. Despite these objective metrics, subjective evaluation conducted by human raters remains an invaluable part of GAN evaluation. This is because human perception can discern subtle nuances in image quality, such as color balance and object relationships, which can be missed by existing objective metrics. Yates et al. *(494)* assessed the efficacy of unconditional GANs in generating realistic aerial images using FID and Kernel Inception Distance (KID) metrics. Their findings revealed a strong positive correlation between KID and FID, though neither correlated significantly with human accuracy, suggesting a potential divergence between these metrics and human perception of realism. They found that urban images had better FID/KID scores than rural ones but were more easily identified by human participants, indicating that achieving low FID and KID scores might not always be crucial for photorealism. The study also showed the potential unreliability of these metrics when comparing single-sample datasets with real datasets due to high variance, and that FID scores can vary significantly when the pretrained model is fine-tuned on EO data, resulting in lower scores. Borji *(495)* provides a comprehensive overview of different metrics used to evaluate the performance of GANs for different types of tasks.

Cyclic Consistent Generative Adversarial Networks (CycleGAN) *(496)* are another type of GAN architecture that learns to translate images from one domain to another without requiring paired training data. It has two generators and two discriminators, where the generators produce images in each of the two domains, and the discriminators distinguish between the generated and real images in each domain (*497*). The two generators and discriminators are trained adversarially, but CycleGAN also introduces a cycle consistency loss that enforces the cycle consistency constraint, by measuring the difference between the original and reconstructed images after translation. Specifically, for an image in domain A, it is translated to domain B using generator G_AB, and then translated back to domain A using generator G_BA. Similarly, for an image in domain B, it is translated to domain A using generator G_BA, and then translated back to domain B using generator G_AB. For each case, the distance between the original and reconstructed images are calculated, using the L1 distance or perceptual loss as a distance measure. By optimizing the adversarial and cycle consistency loss jointly, CycleGAN can learn to translate images between the two domains without paired training data. For example, it was used successfully to translate visible bands of Sentinel-2 contaminated with thin clouds into cloud-free images (*498, 499*). With a similar goal, Li et al. (*500*) combined cGAN and CycleGAN with a physical model of cloud distortion to remove thin clouds from Sentinel-2 images. Mumuni and Mumuni (*501*) provide a comprehensive overview of augmentation techniques used in computer vision.

### 7.2.2 Transfer Learning and domain adaptation

Transfer learning (TL) is a widely adopted approach to address the limitation of training data. It is based on the assumption that a visual representation learned from one dataset (referred to as the source) can be effectively applied to another dataset (the target) that exhibits a roughly similar distribution in terms of both domain and task. The term domain pertains to the input features, including reflectance values or derived features, as well as the probability distribution associated with each feature (i.e., marginal probability). In the domain of satellite imagery, changes in the feature space often intertwine with alterations in the marginal distributions which can arise for numerous reasons including variations in the factors that shape the cover types and their relative abundances (e.g. seasonal variations, geographic transitions), change in sensor characteristics and Calibration especially when the sensor GSD varies (*331, 502–504*), and atmospheric conditions at the time of image acquisition (e.g. soil moisture, lighting, haze) (*505, 506*). The task within TL encompasses the label space, which represents semantic categories, and their corresponding marginal distribution. Discrepancies in the task arise when there are notable variations in the number of classes to be segmented or significant differences in the label distribution between the source and target domains that can manifest as dissimilar numbers of instances for each class.

If source and target datasets are different in both domain and task then the problem is categorized as inductive TL, while transductive TL, also known simply as domain adaptation, applies to situations in which only the domains vary between the source and target (*507*). Achieving good results necessitates a high degree of similarity between domains, as attempting to directly transfer a model to a substantially different or shifted domain usually leads to subpar performance. For instance, Zhang et al. (*508*) reported that the generalization power of a PSPNet for mapping cropland declines when the terrain being mapped changes from mountainous to plains. Another study found that models trained to map roads with data from specific cities performed poorly when used to extract roads in cities in different regions (e.g. a model trained on Las Vegas roads data applied to Kumasi, Ghana) (*509*). To overcome challenges related to domain shift, substantial effort has been devoted to domain adaptation (DA) techniques that help reduce differences between source and target domains, in order to make the model more generalizable and mitigate the small training samples *(510)*. Commonly used DA techniques in RS can be grouped based on the availability and size of annotated labels in the source and target domains.

### 7.2.2.1 Model-based DA

Model-based DA techniques are appropriate when there are sufficient labeled samples in the source domain and at least a small number available in the target domain. Such techniques assume substantial overlap between the visual representations in the source and target domains, especially in early layers of the network, where less task-specific primitive patterns manifest (e.g. object edges). A common practice is to *fine-tune* a pre-trained network by freezing the weight updates for earlier layers, and then train the remaining layers for a number of extra epochs using a small dataset in the target domain *(511)*. Models pre-trained on ImageNet data *(512)* and then fine-tuned for out-of-domain inference on So2Sat *(513)* and Chesapeake Land Cover datasets had 7-10% greater accuracy than models trained from scratch on So2Sat with random weight initialization *(514)*. Similarly, pre-trained ImageNet models aided classification of aerial images in the UCMerced dataset *(515, 516)*. There are numerous additional studies that report on the effectiveness of model-based fine tuning for semantic segmentation with satellite imagery *(45, 46, 517)*.

The efficiency of TL degrades as the source and target domains diverge *(518)*. Liu et al. *(504)* found that when transferring between imagery with different spatial resolution for classifying marsh vegetation, TL works best when source and target have similar spectral ranges. TL methods typically rely on the large datasets produced for the CV community (e.g. ImageNet) that consist of RGB natural images, which can be hard to adapt to multi- or hyperspectral RS imagery *(519, 506)*. This imposes restrictions on the choice of architecture and the number of input features that sometimes force the researchers to do an expert input feature selection to reduce the number of input bands to RGB to such pre-trained models *(520)*, usually leading to the loss of useful discriminative information *(521)* that can limit remote sensing applications. For example, Mahdianpari et al. *(522)* tested multiple DCNN models to map wetland classes using either different RGB combinations or all 5 band of RapidEye optical imagery, and concluded that in all cases models trained from scratch on all 5 bands were superior to models pre-trained on ImageNet, and then fine-tuned on the RGB bands.

Several solutions have been proposed for TL, ranging from RGB inputs derived from spectral dimensionality reduction *(523)*, to approaches that alter the first layer of the network to accommodate the extra bands of the RS images. For example, CoinNet *(524)* redesigns the first network layer by copying the weights for the RGB bands to the extra MSI bands, and then re-trains the whole network on the target domain, enabling the model to fully utilize the spatial and texture information from the source domain in the low-level layers of the network, but with the drawback that this approach can ignore spectral clues. Osco et al. *(525)* simply revised the first layer of the pre-trained network to handle the extra channels, which were initialized with random weights. TL-DenseUNet *(526)* applied TL to work with four band Gaofen-2 imagery in two steps. In the first step, the first layer of DenseNet-121 was adapted by randomly initializing the weights for the extra fourth band throughout the entire network, as well as the weights in the decoder subnetwork. The rest of the network was then frozen and the remaining active layers were fine-tuned on the target dataset. In step two, the whole network was fine-tuned on the target domain for several additional epochs.

**7.2.2.2 Instance-based DA**

In instance-based domain adaptation the focus is on adapting the individual source instances to the target domain rather than adapting the entire feature space or distribution of the source domain to the target domain. This approach includes techniques such as domain adaptation through sampling bias correction techniques, such as instance weighting, which assigns weights to each instance in the training dataset based on its similarity to the target domain *(527)*. With modern ANN training, instance-based DA is typically used in conjunction with other techniques to improve performance.

**7.2.2.3 Feature-based DA**

Feature-based DA techniques are employed to minimize the differences between the distributions of features in the target and source domain, typically by finding a mapping function that translates features directly from one domain into the other, or by finding the shared domain-invariant feature representation that works well in both domains. Feature-based DA techniques are better than instance-based ones if the feature support (i.e. range of values that a particular feature or variable can take) and marginal distributions both change between the source and target domains *(528)*, but these assume high similarity between the conditional distributions from each domain *(529)*. Two commonly pursued procedures include domain discrepancy reduction and generative approaches. Domain discrepancy reduction works by mapping both distributions into a feature space that minimizes the difference in the statistical characteristics of source and target domain distributions. Some methods, like Maximum Mean Discrepancy (MMD) *(530)*, focus on aligning the means of the distributions, while others, like Correlation Alignment (CORAL), also aim to align the covariance structures. A kernel function, such as the Gaussian kernel, is typically used to map the features to a space that minimizes the discrepancy between the source and target domain distributions. In an example of MMD, Wang et al. *(531)* used a two-branch architecture to classify crop types using Sentinel-2 time series, with one branch for each domain, in this case different crop-growing regions. The branches in the model shared parameters, which were trained on the source domain with a large number of labels and then applied to the target domain. Both branches consist of four bidirectional GRU layers and a self-attention module, and were supervised using a compound loss with a term for multi-kernel Gaussian MMD, which helped compensate for phenological differences between regions.

Generative approaches utilize adversarial learning methods to align the source and target distributions *(532)*. In this process, the generator serves as a mapping function that takes input from the source domain and generates synthetic samples that closely resemble the target domain samples, while the competing discriminator is a binary classifier that distinguishes between the source and target domain samples. To quantify the divergence between the source and target distributions in the joint feature space, various divergence metrics can be used, such as Jensen-Shannon divergence, MMD, or Wasserstein distance. These functions measure the dissimilarity between the distributions and guide the optimization process to minimize this divergence, which aligns the two domains and facilitates the transfer of knowledge from the source to the target domain. For instance, Benjdira et al. *(510)* tried to improve TL for multi-sensor datasets using a two stage process that first applied a cyclic GAN to map the source domain (the Potsdam dataset) into the target domain (Vaihingen dataset), and in the second stage fine-tuned the model using this mapped representation. A family of techniques more tuned for LU/LC mapping is class-aware domain adaptation, which is based on the notion that the marginal distribution alignment between source and target domains are not sufficient to consider the distribution of semantic classes (e.g. $p^{y|x}$) in the DA process. These usually consist of multiple adversarial networks that ensure both global and local consistency in mapping between domains, such as the hybrid adversarial framework introduced by Fang et al. *(533)*, which embeds a geometry-consistent generative adversarial network (GcGAN) as a feature translator inside a co-training adversarial learning network (CtALN). The GcGAN module uses structure-preserving transformations to ensure the DA is sensitive to land cover categories. Tuia et al. *(534)* provides a comprehensive overview of common DA techniques for RS data.

### 7.2.3 Weakly-supervised learning

When dealing with incomplete annotations, specifically when only easily identifiable pixels are annotated, weakly-supervised learning can be employed. Weak annotations are diverse including partially labeled images also known as scribbles *(535, 536)* or other forms of approximate annotations, such as bounding boxes *(537)*, or proposal masks *(538)*, which require much less effort to collect than detailed pixel-level annotations *(539)*. In the context of weakly supervised learning, image-level annotation is the least informative type of annotation as it doesn't provide any spatial prior information. However, it is highly sought after as it requires the lowest data collection cost *(540)*. For example, The SEC *(541)* framework uses a three-step process for weakly supervised semantic segmentation: Seed, Expand, and Constraint. The Seed phase uses a pre-trained network, such as VGG, for multi-label classification, generating reliable localization heat maps, or 'seeds'. These seeds are combined into a single weak localization mask. The Expand phase uses global weighted rank pooling to turn score maps from a pre-trained segmentation network into classification scores, adjusting these for each foreground class and object seeds with a specific decay parameter. The Constraint phase combats the common issue of imprecise boundaries in image-level labels training by introducing a constraint-to-boundary loss, formulated as the mean KL-divergence between the network and fully-connected CRF outputs. Nivaggioli and Randrianarivo *(542)*, use a four step framework. The first step involves training a multi-label image classification CNN using image-level annotations, followed by extraction of a set of class activation maps (CAMs) for each image which focus on the discriminative areas in the image where the relevant features are located. In the second step, an affinity network is trained to understand the relationships between each pair of pixels in each input image. The third step integrates the CAMs and affinity labels through a random walk process to generate segmentation labels. Finally, in the fourth step, the segmentation labels produced by the third stage are used to train a segmentation model.

Wei and Ji *(538)* developed a context-aware propagation algorithm for semantic segmentation that translates noisy OpenStreetMap road centerlines into a pixel-wise proposal mask using combined buffer and graph-based methods. Their dual branch encoder-decoder model uses a shared ResNet-34 encoder, and two decoders for generating segmentation and boundary masks. The boundary branch is fed with intermediate encoder features and ASPP-processed encoder outputs, which after upsampling and convolution, generates a boundary map. The features from the boundary branch are then integrated with segmentation scores to refine the segmentation branch results. Loss is calculated using mean squared error for boundary predictions compared to reference boundary masks generated using a pre-trained holistically-nested edge detection (HED). A cross-entropy plus conditional random field term is used for segmentation predictions, with total loss being a linear combination of the two. Hua et al. (543) formulate learning from sparse annotations as a semi-supervised learning to leverage the information content of the unlabeled image pixels which uses the clustering assumption in both spatial and feature domain. The training loss is the summation of CE loss on the labeled samples and a compound loss with three terms that calculates the Euclidean distance between each sample and nearest sample in the feature space and spatial domain, plus the third term that calculates the cosine similarity of the sample and farthest sample in the feature space. Instead of mimicking full supervision by creating fake fully-labeled masks from partial input, Tang et al. (544) seek to incorporate unsupervised loss terms. These terms are inspired by the mechanisms of graph cut (Normalized cut loss, (545)) and dense Conditional Random Field (CRF) algorithms and act as regularizer on the segmentation loss over partial input.

**7.2.4 Semi-supervised learning**

Semi-supervised learning (SSL) combines supervised and unsupervised learning methods, and is typically undertaken by refining a model trained on a limited number of annotated samples with complementary information derived from local dependencies within the feature space of a larger set of unlabeled samples (546). This approach often outperforms fully-supervised models trained on a smaller number of labels (543). SSL methods may be combined with DA techniques when the labeled and unlabeled samples originate from different distributions.

Several techniques are employed in SSL which typically incorporate multiple networks or branches that use composite loss functions consisting of supervised and unsupervised terms. A good example of generative models is Li et al. (547) which introduces a novel approach for cross-domain RS image semantic segmentation, leveraging DualGAN under two scenarios, varying in both geographic location and imaging mode. This method employs a unique objective function that features multiple weakly-supervised constraints, namely the weakly-supervised transfer invariant constraint (WTIC), weakly-supervised pseudo-label constraint (WPLC), and weakly-supervised rotation consistency constraint (WRCC). WTIC utilizes a very similar architecture to cycleGAN called DualGAN (548) to conduct unsupervised style transfer, thus forming a bridge between the source and target domains. WPLC employs a classification confidence filter to adaptively select anchor points with pseudo-labels from target domain images. In contrast, WRCC builds upon the premise that the inverse transformation of a segmented result of a rotated image should coincide with the original image's result, depicting the images' generalized rotation consistency property from the target domain. The study further presents a dynamic optimization strategy to dynamically adjust the constraint weights during training, effectively preventing gradient degradation. Hung et al. (549) proposed an adversarial non-generative setting comprised of a segmentation network (e.g. DeepLab) that replaces the generator and outputs class probability maps and a fully convolutional discriminator network, which produces feature maps in which each pixel's value represents the probability that it comes from the labels (p=1) or was predicted by the segmentation network (p=0). The segmenter network is trained using both labeled and unlabeled images, using a composite loss that uses cross-entropy loss for the labeled samples, a semi-supervised loss for the unlabeled data, and an adversarial loss to interact with the discriminator network. The discriminator network is trained only with labeled data and supervised using cross-entropy loss. The generator creates an initial segmentation of the unlabeled image, which the discriminator evaluates to produce a confidence map. This map is then binarized, using a threshold, into reliable regions that serve as supplementary training signals for the segmentation network, employing a masked cross-entropy loss.

Pseudo-labeling, also known as self-labeling, is another technique used both in semi- and self-supervised training, where a model is initially trained on labeled data from the source domain and then used to predict on the unlabeled data from the target domain. High-confidence predictions are then added to the labeled dataset for model retraining, and this process is repeated until convergence. For example, the Instance Adaptive Self-Training (IAST) framework (550) includes an instance adaptive selector (IAS) and region-guided regularization components. The IAS selects an adaptive pseudo-label threshold for each semantic category to create a multi-level confidence map of the predictions, aiming to reduce noise in the pseudo-labels and improve training data quality. Region-guided regularization is an additive term to the loss function, designed to smooth the prediction of the confident region and sharpen the prediction of the ignored region.

Within the unsupervised term of the composite loss functions that are used in SSL methods, consistency regularization and entropy minimization are commonly used techniques. Consistency regularization is used to improve model stability by applying transformations (e.g. translations, rotations) or perturbations (e.g. noise) to the input or latent features respectively or use different initialization of the network to produce predictions that are regularized to get similar. This regularization is achieved by minimizing a distance function (e.g. mean squared error (MSE), Kullback-Leiber divergence (KL)) between the prediction outputs (e.g. original and transformed), based on two assumptions. The first is that samples close to each other are likely to have the same label (i.e. smoothness assumption), and the second is that the decision boundary between classes should lie in low-density regions (i.e. cluster assumption), which encourages the model to make similar or consistent predictions for an unlabeled sample. For instance, Ouali et al. (551) proposed cross-consistency training (CCT), in which the architecture was designed with a shared encoder and multiple decoders (e.g. a main decoder and a set of auxiliary decoders.), based on the finding that low-density decision boundaries are more obvious in the hidden representations of the model after the encoder. The model applies different perturbations to the encoder's outputs, and forces the model predictions to remain consistent under these perturbations. The main decoder is trained in a supervised manner using the labeled data while the auxiliary decoder networks are trained on the unlabeled set by maintaining a consistency of predictions between the main and auxiliary decoders. Each auxiliary decoder is fed a perturbed version of the encoder's output, while the main decoder receives the uncorrupted intermediate representation. The loss function is the linear combination of CE on the labeled samples and MSE between the output of auxiliary decoders compared to the main decoder, which serves to enhance the representation learning of the encoder using the unlabeled examples, and subsequently, the performance of the segmentation network (figure 32).

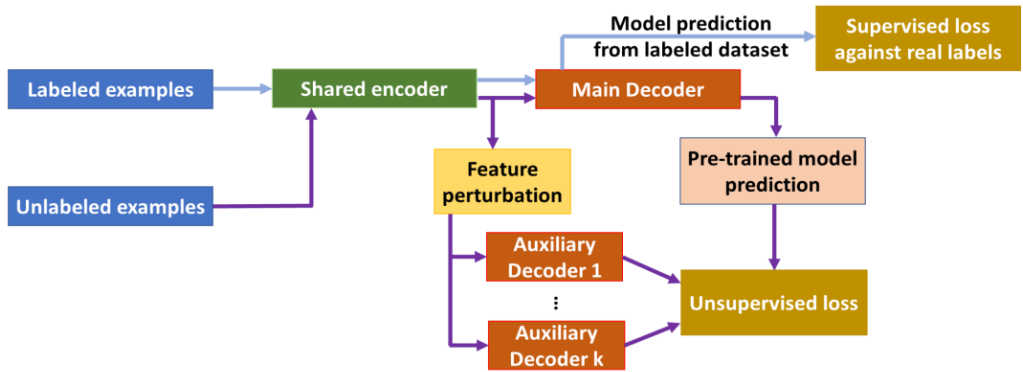

Figure 32. Semi-supervised semantic segmentation using consistency training as introduced by Ouali et al. (2020).

The Mean Teacher model (552) is a dual-network strategy, involving a student and a teacher network, both with identical architectures. However, the teacher network's parameters are calculated as an Exponential Moving Average (EMA) of the student's sequential weights during a given training step. In terms of data usage, both the student and teacher models receive input samples that are augmentations of the original sample. For labeled data, the student model is trained in a supervised manner. For unlabeled data, the teacher's prediction, which serves as pseudo-labels, is used to guide the training of the student model. And a consistency loss is applied on the predictions from the student and teacher networks.

A limitation of consistency learning is that it presumes accurate predictions for unlabeled images. If this assumption isn't met, perturbations could inadvertently push image features past the true classification boundary, resulting in incorrect training signals. This challenge is amplified in network perturbation-based consistency learning, where incorrect predictions from one network harm the training of other networks. Using only CE loss exacerbates this issue, as it can cause overfitting to prediction errors and lead to confirmation bias. While studying the L2 distance between patches focused on adjacent pixels, French et al. (553) found that significant differences between such patches don't always indicate different classes as regions of low data density in the input often don't correspond accurately to actual class boundaries. To promote consistent predictions, the authors advocate for robust mask-based augmentation techniques like cutOut, cutMix, and classMix. For instance, ClassMix (554) is an augmentation technique specifically developed for semi-supervised semantic segmentation, creating new images and corresponding artificial labels from unlabeled samples. It works by selecting two unlabeled images from the dataset and predicting their semantic classes using the model pre-trained on available annotated samples. A binary mask is generated, by randomly selecting half the classes from the argmaxed prediction of one image. This mask is used to mix the two images and their predictions, resulting in an augmented image and its corresponding artificial label. This process produces novel and diverse outputs that still resemble other images in the dataset. Even with the possibility of artifacts from this mixing process, performance improves as training progresses due to effective consistency regulation and can easily get integrated with the mean-teacher and pseudo-labeling strategies. A common approach is to only use a subset of unlabeled dataset which produce high confidence predictions. As an example, FixMatch *(555)* combines consistency regularization and pseudo-labeling, employing distinct weak and strong data augmentations. It begins with generating weakly augmented versions of unlabeled images, predicting class distributions for each, and assigning the class with the highest probability as a pseudo-label. Concurrently, strongly augmented versions of the same images are produced. Then, these strongly augmented images are compared to the pseudo-labels from the weakly augmented versions, ensuring consistency in predictions. DASH (Dynamic Thresholding) *(556)* is a semi-supervised learning (SSL) algorithm that strategically incorporates unlabeled data into training. It starts by setting a threshold based on the loss values of the labeled data. As training proceeds, the algorithm dynamically selects unlabeled data that have loss values smaller than this threshold for inclusion in the training set. The threshold value is progressively lowered throughout the optimization process, which enhances the effective utilization of the unlabeled dataset and guarantees non-asymptotic convergence. Wang et al. *(557)* propose a mean-teacher framework to balance training, stating that using only high confidence predictions as pseudo-labels can overlook low-confidence predictions, resulting in an imbalanced learning. In their architecture, both labeled and unlabeled images are sampled equally at each training step. The standard cross-entropy loss is minimized for labeled images, while unlabeled images are processed via the teacher model to get predictions. Pixel-level entropy with adaptive weight adjustment, determined by the inverse of the pixel percentage below a certain entropy threshold, is used to discern reliable pseudo-labels from unreliable ones for calculating unsupervised loss. The unreliable predictions, typically confused among a few classes, are considered as negative samples and utilized through a contrastive loss function. These negative samples are stored in a FIFO queue for each category to ensure balance. The overall loss then becomes the summation of supervised, unsupervised, and contrastive losses. The threshold for separating reliable and unreliable predictions is adaptively adjusted using a linear strategy as the model improves.

Liu et al. *(558)* extended the Mean Teacher model by introducing an auxiliary teacher and employing confidence-weighted Cross-Entropy (CE) for the consistency loss. The output prediction is derived from the aggregate predictions of both teacher models. Notably, during the training process, only one teacher network is updated in each epoch. Authors use adversarial feature noise produced using the more accurate teacher models, and then apply this estimated noise to the student model's features—a method referred to as T-VAT. Beside the latent features, inputs to the teacher model are augmented weakly (e.g. geometric augmentation) but the input to the student network are subjected to strong augmentations (e.g. photometric augmentation, cutMix). Cross-pseudo supervision *(559)*, uses two similar segmentation networks that are initialized differently. A pixel-wise cross-entropy function governs the training loss on the labeled images, while a bidirectional pseudo-supervision loss uses the confidence map of the first network to supervise the pixel-wise confidence map from the second network, and uses the output of the second network to supervise the first.

As an uncertainty measure, entropy minimization pushes the model to make confident predictions on unlabeled data. For example, Wu et al. *(560)* developed a model that makes use of gradient information, aleatoric uncertainty, and consistency constraints to enhance semi-supervised learning for semantic segmentation. The architecture generates classification and segmentation mean and variance from the encoder ($out \in R^C$) and decoder ($out \in R^{H \times W \times C}$) outputs, respectively. For labeled data, the authors further use the linear combination of random noise sampled from the standard Gaussian distribution and the mean and variance parameters, which are then fed to the cross entropy loss. The unsupervised loss consists of three terms: feature gradient map regularization (FGMR), adaptive sharpening, and class consistency. FGMR leverages gradient maps from lower layers, in order to improve the encoding capacity of deeper layers and enlarge the inter-class distance. Adaptive Sharpening uses the variance of the model prediction to filter noisy samples and by adjusting the temperature of the categorical distribution to pay more attention to non-noise. It also uses unlabeled data and class consistency to alleviate the noise that might be introduced through the sharpening process.

### 7.2.5 Self-supervised learning

When dealing with scenarios where annotated training data are unavailable, self-supervised learning—a new branch in unsupervised learning—becomes a prevalent strategy. These techniques can be generally categorized into generative or discriminative approaches, underpinned by reconstruction-based and contrastive losses, respectively. After training on a preliminary or pretext task, the model ideally extracts valuable features that can be transferred to other downstream tasks with minimal modification. Reconstruction-based methods, which are a subtype of generative models, leverage various pretext tasks including color transformations, context-based tasks like predicting relative locations (*561*), Jigsaw puzzle (*562*, *563*) and masked image modeling (MIM) (*564*). Although less frequently mentioned in literature, GANs also find applications in self-supervised learning. For instance, Treneska et al. (*565*) used a cGAN for image colorization pre-text task. In this task, the network was conditioned on a grayscale input image to produce a colored output image, a proxy task for visual comprehension. The knowledge acquired was subsequently transferred to two downstream tasks—multilabel image classification and semantic segmentation.

Contrastive learning is a form of self-supervised learning that encourages a model to learn features in a way that similar examples (positive pairs) are proximal in the representation space, while dissimilar examples (negative pairs) are distanced. The most common approach, Instance-level contrastive learning, exemplified by SimCLR (*566*), regards features of each image as an independent class. Positive pairs are formed from distinct transformations of a single image, while all other images and their augmentations are considered negative examples. The contrastive loss function is typically based on the noise-contrastive elimination (NCE) strategy such as Normalized Temperature-scaled CE (NT-Xent) or InfoNCE, which promotes the model to develop representations where an image and its augmentations are closer in representation space than to any other image (noise). For instance, NT-Xent is the ratio of the sum of similarities between all positive pairs divided by the negative log-likelihood of these pairs being similar (*567*). This form of instance-based contrastive learning can extend to SITS, where spatially identical but temporally varying pairs of satellite images are considered positive (*568*). DenseCL (*569*), designed for dense prediction tasks, is a self-supervised learning framework comprising a backbone network and a projection head with two parallel sub-heads—a global projection head and a dense projection head. The backbone is retained after pre-training, while the projection head is discarded. Like in traditional Instance-level contrastive models, the global projection head outputs a global feature vector for each view via global pooling, while the dense projection head produces dense feature vectors, preserving spatial information. The model is trained end-to-end by optimizing a joint pairwise contrastive loss at both global and local feature levels, performing contrastive learning densely using a fully convolutional network, analogous to the target dense prediction tasks. Ayush et al. *(570)* proposed a self-supervised learning framework based on MoCo-v2 to leverage the spatio-temporal structure of remote sensing data through geography-aware contrastive learning and usage of RS-tailored methods to generate positive and negative pairs. Specifically, the contrastive framework constructs temporal positive pairs from spatially aligned images over time, thereby learning representations that are invariant to minor changes over time, such as seasonal variations. Furthermore, the framework incorporates a pretext task of predicting the geographical origin of an image which is done by clustering the images in the dataset using their coordinates and assigning each cluster a categorical geo-label which is trained using a CNN with CE loss. The pretext task and contrastive learning are combined through a CE loss that takes encoded representations of the randomly perturbed query branch as the input to CNN instead of the original positive temporal image, fostering the development of representations that carry geographical information, which can be beneficial in remote sensing tasks.

In the Jigsaw Puzzle approach, an image is split into non-overlapping patches, which are then randomly shuffled or permuted, creating a sort of jigsaw puzzle. The task of the model is to predict the initial configuration of these patches which enables the model to comprehend the spatial and contextual relationships between different image components *(571)*. For instance, an approach called Completing damaged jigsaw puzzles *(572)* uses puzzles with one piece missing and the other pieces devoid of color and the model's task is to solve the puzzles, generate missing content, and colorize the pieces. Similarly, in the relative location prediction technique, an image is split into a grid of patches. Two patches are randomly selected: one is treated as the anchor patch and the other as the context patch. The model's task is to predict the relative spatial position of the context patch with respect to the anchor patch (e.g. to the left, to the right, above, below, etc.) *(573)*.

MIM is an image inpainting technique that involves masking or erasing sections of the input image. The model is tasked with reconstructing the original image by filling in these missing or masked portions *(564, 574, 575)*. Both methods enable the model to learn high-level feature representations of data, which can then be leveraged for downstream tasks. For instance, after the self-supervised learning phase, the model can be fine-tuned on a smaller labeled dataset for image classification or semantic segmentation tasks. The efficacy of self-supervised learning methods is evaluated by their performance on these downstream tasks *(576)*. A striking example of this is the Masked Autoencoder[5] (MAE) *(564)*, an encoder-decoder design that masks a large portion of the input (e.g., 75%) and feeds the remainder into a Vision Transformer (ViT) encoder. The MAE decoder receives a full set of tokens, including encoded visible patches and mask tokens indicating the presence of missing patches to predict. Positional embeddings are added to all tokens, providing information about their location in the image. While only the encoder is used for image representation in recognition tasks, the decoder's flexible design is crucial for the pre-training image reconstruction task. The decoder output forms a reconstructed image and the loss function computes the mean squared error (MSE) between the reconstructed and original images in the pixel space, considering only masked patches. ConvMAE (574) is an extension of MAE that uses hybrid convolution-transformer encoder, capable of learning discriminative multi-scale visual representations. The encoder employs masked convolution blocks for local content encoding at early stages and transformer blocks to aggregate global context at later stages and utilizes a block-wise masking strategy, progressively upscaling the mask to larger resolutions in early convolutional stages. This approach ensures that tokens processed by the late stages are segregated into masked and visible tokens, maintaining the computational efficiency of MAE. SatMAE (575), an extension of the Masked Autoencoder (MAE), has been specifically designed to handle the temporal and multispectral characteristics of EO satellite imagery. In this model, the authors introduce a temporal encoding scheme, which is integrated alongside the positional encoding. This combined encoding effectively caters to the spatial-temporal aspects of the imagery. In terms of handling the multispectral aspect, the channels are divided into groups, each of which is embedded separately. The final embedding tokens are formed by the concatenation of these individual group channel embeddings, along with an added spectral encoding. This approach allows the model to better comprehend the complex, multispectral nature of the EO satellite imagery. Furthermore, the authors introduce a unique masking strategy. They independently mask image patches across different time points, compelling the model to rely on unmasked band groups to reconstruct the corresponding region within a masked band group. This strategy further enhances the model's ability to analyze and learn from the complex, time-variant nature of EO satellite imagery. Scale-MAE (577) is an extension of MAE, designed to learn relationships between data at multiple scales during the pre-training process. Key features of Scale-MAE include a Ground Sample Distance Positional Encoding (GSDPE) that informs the model (e.g. ViT) about the position and scale of an input image through scaling the positional encoding relative to the area covered in an image, regardless of the resolution of the image. The second innovation is using a Laplacian-pyramid decoder to encourage the network to learn multi-scale representations which, inspired by super-resolution strategy, decodes the embeddings into two images: one with low-frequency information and the other with high-frequency residuals. This structure allows the ViT decoder to utilize fewer parameters than standard MAE while still generating strong representations across various scales. The Scale-MAE decoding process involves three steps: initial decoding of encoded values, progressive upsampling through deconvolution before passing through Laplacian Blocks, which reconstruct low and high-frequency features at different scales. The model's performance is evaluated using an aggregate loss calculated from the low and high-frequency predicted features against low and high high-frequency features extracted from the ground truth, with L1 loss for high-frequency output to better reconstruct edges, and L2 loss for low-frequency output to better reconstruct average values. SITS-Former, introduced by Yuan et al. (578), consists of the image patch embedding and Transformer encoder modules. In the embedding phase, each image patch (5x5) and its acquisition time are transformed into an embedding vector, which combines the spatio-spectral features — derived through a lightweight 3D convolutional network — with a pre-defined positional encoding vector that uses the sinusoidal positional encoding technique to handle irregularly sampled time series. The Transformer encoder subsequently refines these embeddings, allowing patches to interact and identify temporal patterns. SITS-Former utilizes a self-supervised pre-training stage focusing on "missing-data imputation", where 15% of the time series patches are randomly masked in each training epoch, followed by a supervised fine-tuning phase for Sentinel-2 time series classification.

---

[5] An Autoencoder (AE) is a type of neural network that learns to reproduce its input to its output without supervision. It does this by learning to compress data into a lower-dimensional representation, often called the latent space, via an encoder function, and then reconstructing the original input from this latent representation using a decoder function. As a result, autoencoders can learn important properties of the input data and are commonly used for dimensionality reduction, feature learning, and generative modeling. Due to its ability to capture complex data relationships in an unsupervised manner, autoencoders have found widespread use in various domains, including pattern recognition and prediction tasks based on spatio-temporal data.

Active learning is a machine learning approach designed to select the most informative and representative instances from an unlabeled dataset for labeling using some information measure. With active learning integration, the sampling strategy in self- and semi-supervised learning can be optimized to select a more diverse set of training data from abundant unlabeled instances (579). For instance, Li et al. (580) developed an active self-learning temporal-ensembling convolution neural network (A-SL CNN) framework, which combines self-learning and active learning strategies through the interaction of student and teacher CNN branches. The approach generates less error-prone pseudo-labels, and results in significant improvement in flood mapping from SAR imagery over urban areas.

Patel et al. (581) assess the effectiveness of self-supervised and semi-supervised learning techniques in performing three distinct remote sensing segmentation tasks: riverbed segmentation, land cover mapping, and flood mapping under label scarcity and geographical domain-shifts. They adopt SimCLR for self-supervised learning, which leverages a contrastive loss method, and FixMatch for semi-supervised learning, utilizing a consistency regularization strategy. The findings suggest that these methods markedly enhance generalization performance in comparison to supervised training, regardless of whether the latter is pre-trained on ImageNet or randomly initialized. This enhancement is particularly significant in scenarios where labeled data is scarce (e.g. 1%) and where geographic domain shifts occur between the training dataset and the validation or test datasets. According to Cha et al. (568), foundation models for RS demonstrate improved performance with an increase in the number of parameters. They also benefit significantly from training on extensive and diverse datasets, which enhances their robustness and boosts their performance across varied tasks. Furthermore, pretraining these models with multi-modal information, such as geo-location, audio, and multispectral imagery, yields superior performance compared to using a single modality. Wang et al. (582) and Tao et al. (583) also provide excellent reviews of self-supervised learning.

## 7.3 Imbalance in training data

Imbalance in training data is a frequent challenge in remote sensing analyses, and is often a result of the natural spatial distribution and varying sizes of semantic objects, but can also arise through the choice of sampling design. The consequences of such imbalance can severely impact model performance (584), and the repercussions become even more pronounced as the complexity of the semantic segmentation task increases (585). Imbalance manifests in two primary forms, which are semantic imbalance, or variations in the prevalence of categories in the training sample, and difficulty imbalance, which refers to differences in the proportions of easy and hard to predict samples (586).

Semantic imbalance can be addressed using a sampling strategy that changes the class distribution, either by oversampling the minority class and/or undersampling the majority class in the input. In practice, these manipulations are often implemented together during the chipping process *(587)*. As examples of dataset level manipulation, Mei et al. *(411)* randomly selected a fixed number of labeled samples from each class, while Small and Ventura *(588)* increased the fraction of minority class samples without reducing the model's accuracy for the majority class. Semantic imbalance can also be dealt with by using loss functions that apply a weighting scheme to compensate for imbalance, such as Weighted Cross Entropy (WCE) and Balanced Cross Entropy (BCE). Rezaei-Dastjerdehei et al. *(589)* showed that a careful choice of the weight parameter in WCE can significantly increase the recall score (~10%) in the output prediction with only a small decrease (~3%) in the precision score. Common weighting schemes include the inverse of class frequencies *(266)* and median frequency balancing *(212, 198, 46)*. Cui et al. *(590)* argues that the commonly pursued strategy of chipping with overlap changes the true proportion of semantic classes in the training phase, leading to false class weights. They developed a method to compensate for such bias by calculating the effective number of samples for each semantic category and using that number with the chosen weighting scheme. Johnson and Khoshgoftaar *(591)* provides additional information on issues related to class imbalance in deep learning models.

Difficulty imbalance refers to the variance in the level of difficulty to correctly classify different types of pixels. This problem often arises along class boundaries where difficult-to-classify pixels are often found, which also tend to be relatively few in number *(592)*. It can also occur with objects of small size, or on non-boundary edges in object interiors. Difficulty imbalance is often addressed by using a loss function that lowers the influence of easy to classify samples, such as Focal loss *(593)* and its integration with variations of Dice loss *(594, 294)*, which calculate loss as the arithmetic mean of false positive (FP) and false negatives (FN). Another alternative is Tversky-Focal loss *(595)*, which evaluates the trade-off between FP and FN through a weighting hyperparameter. Doi and Iwasaki *(586)* reported that the Focal loss function led to a substantial improvement in tackling difficulty imbalance, although its effectiveness varied with model architecture, and the focal hyperparameter required careful tuning. Wu et al. *(466)* used adaptive thresholding based on the Jaccard similarity score to choose the hardening threshold for the class probability for each pixel, which helped to improve predictions for small, hard-to-classify objects. Zhao, et al. *(596)* leverage the linear correlation between the predicted map and the ground truth label map to prioritize pixel positions in corresponding regions that contribute to greater structural differences. They do this by reweighting the cross-entropy loss based on structural similarity, and discarding samples with high similarity scores to streamline the process. This method enhances the focus on areas with larger structural disparities, thereby improving the accuracy of segmentation.

The impact of imbalance is inadequately measured by many common model evaluation metrics, which are mostly derived from the summarized pixel-based comparisons between the model predictions and reference labels provided by the confusion matrix *(597)*, which can obscure the true performance of the model for small, hard-to-predict classes *(598)*. Certain measures are indeed more sensitive to data imbalance. For instance, overall accuracy can be misleading in imbalanced datasets. If 95% of the data belongs to Class A and 5% to Class B, a naive model that predicts Class A for all instances will achieve 95% accuracy, even if it is completely unable to identify Class B. Other sensitive measures include Error Rate, User's Accuracy (Precision; Positive Predictive Value), True Negative Accuracy (Inverse Precision; Negative Predictive Value), Commission Error (False Discovery Rate), False Omission Rate, Matthews

Correlation Coefficient, Markedness and Intersection Over Union (Jaccard Index). On the other hand, measures like recall (Producer's accuracy, Hit Rate; True Positive Rate; Sensitivity) and specificity (True Negative Rate, inverse recall), False Positive Rate (false alarm rate; Fallout), Omission Error (False Negative Rate; Miss Rate), Likelihood Ratio, Geometric Mean, and Bookmaker Informedness (Youden's index) are considered less sensitive to class imbalance. F1-score is the harmonic mean of precision and recall, two metrics that provide different perspectives on a model's performance[6] which makes it a safer option in the context of imbalanced binary datasets. In multiclass problems, a version of the F1-score is typically calculated for each class, but when the number of classes is high, these scores are commonly aggregated to get a single measure. The way in which this averaging is done can result in the F1-score being sensitive to class imbalance. For example, if you simply calculate the mean F1-score (also known as macro-average F1), this could give too much weight to small classes, since each class contributes equally to the final score. Conversely, if you calculate a weighted average based on the size of the classes (micro-average F1), this could give too much weight to large classes. Luque et al. (599) provide an in-depth analysis of the impact of imbalanced data on evaluation metrics and propose formulas to remove the imbalance bias for a number of the mentioned metrics for binary cases.

## 7.4 Label quality

In addition to the size and representativeness of the training dataset, the quality of the assigned labels is commonly a significant determinant of model performance, regardless of the chosen model (600, 601). Label noise refers to inaccuracies or errors in the target labels in the training data or validation dataset. Empirical studies within the domain of general machine learning algorithms show that the detrimental effects of poor label quality often surpass those induced by input feature noise (602). Elmes et al. (459) provides a systematic overview of the taxonomy of label noise, or error, and its effects on a general learning system within the field of remote sensing. They attribute such noise primarily to errors originating from the design and collection of the training samples. Design-related errors often arise due to a temporal and/or resolution mismatch between the base imagery and the very high resolution (VHR) or in-situ campaign used for collecting training labels. Collection-related errors, on the other hand, are typically a result of mistakes or inconsistencies during the manual image interpretation process with contributing factors such as poorly defined or communicated semantic classes, complex scenery, the interpreter's skill and contextual knowledge, as well as the ambiguity of class boundaries. For quantification purposes, label error is typically decomposed into semantic (e.g. thematic) and geometric (e.g. positional) errors. Geometric label errors are commonly caused due to the mis-alignment issue and their treatment is not considered in this text. Thematic noise is mainly considered for polygons where the polygon interior is assumed to have high semantic quality and as we get closer to the object boundaries in the annotation layer this quality degrades. This is even true for crisp semantic boundaries in large training datasets due to the fact that capturing fine semantic borders requires taking many more vertices than coarse annotation, which is time consuming. In their study, Elmes et al. introduced varying degrees of label noise into the SpaceNet Vegas buildings dataset, including instances of shifted or deleted building footprints. They observed that as the level of noise in the labels increased, there was a corresponding decrease in the prediction quality from the U-Net architecture used in the case study. Schmitt et al. (*603*) investigate the application of weakly-supervised learning strategies to generate high-resolution land cover predictions. This approach integrates high-resolution satellite imagery with existing global land cover maps, which, despite their extensive coverage, tend to be low-resolution products of traditional, automated classification algorithms. Using the SEN12MS dataset, the study highlights the potential challenges and improvements in this field. Initial results suggest that easily identifiable classes like Forest, Urban, and Water are well distinguished. However, classes such as Shrubland, Barren, and Wetlands, which are more challenging to identify and carry significant label noise in low-resolution land cover maps, can considerably decrease overall accuracy. Exploring medical image segmentation under different noise regimes, Heller et al. (*604*) found that CNNs' performance degraded steadily as the boundary noise increased, but tested models were robust to high degrees of non-boundary label noise. Zlateski et al. (*605*) investigated the effects of label quality on the model's performance for semantic segmentation of street-view images and reported that a large dataset of lower annotation quality can perform as well as a smaller dataset of higher quality annotations. They conclude the best trade-off between effort for better quality labels and model performance can be achieved through a two stage training, in which the model is first trained on the large but low quality dataset and then fine-tuned using the smaller, higher quality dataset. Such an approach was adopted in a road-mapping study, in which the model was initially trained using noisy labels derived from OSM road layers, and then fine-tuned the model using a smaller dataset with high label quality (*606*).

---

[6] Precision evaluates how many of the predicted positive instances are actually positive, whereas recall assesses how many of the actual positive instances are correctly identified by the model. The F1-score balances these two aspects, rewarding models that maintain both high precision and high recall.

The typical strategy to mitigate the impact of label noise is to use loss functions that are robust to label noise which mostly require multi-stage training. For instance, Patrini et al. (*607*) proposed two architectural and application-domain agnostic procedures for loss correction in the presence of class-dependent label noise. Training of the noise-aware model is decoupled into two stages, where the first stage calculates the required knowledge of the probability of each class being corrupted into another using an error transition matrix (confusion matrix), estimated directly from the noisy dataset. The second stage then learns with corrected loss. The two stage training is implemented either through manually cleaning a subset of the training dataset and then training with the corrected loss. Alternatively, the network is trained with the un-corrected loss on noisy data, and then the trained parameters are used to initialize the second stage, which uses the corrected loss. They also noted that networks that only use ReLU for activation are more robust to label noise. The proposed framework by Vahdat (*608*) also used a small set of clean samples along with the noisy, and used an undirected graphical model (e.g. CRF) that represents the relationship between noisy and clean labels, which they trained in a semi-supervised manner using a variation of Expectation-Maximization as the loss. Acuna et al. (*609*) introduce Semantically Thinned Edge Alignment Learning (STEAL), which iteratively refines noisy annotated boundaries during the training stage. STEAL consists of a module and a customized loss function as the linear combination of BCE, non-maximum suppression loss, which directly optimize for NMS[7] edges resulting in thinner edges, and a direction loss

---

[7] For edge or boundary detection, the response of the edge detector (which could be a filter like Sobel or Canny, or a deep learning model) typically produces a map of edge strength where higher values correspond to a higher likelihood of an edge being present. However, this edge map often includes many adjacent pixels along the actual edges, leading

based on the mean squared angular distance between the normal directions of predictions and reference annotation. STEAL can be added to any existing boundary detection networks and can actively handle coarse labels with mis-aligned boundaries using a level set formulation. The training procedure consists of two steps, in which the first step refines boundaries, and the second optimizes the model's parameters by learning from the refined boundaries. These two steps are performed alternately, with each step improving the solution of the other until a stopping criterion is met. This type of procedure is a common optimization method for problems where the objective function depends on multiple sets of variables, and is often known as block coordinate descent or alternating optimization.

Although CNN models have the capacity to fit completely labels that consist entirely of random, research shows that during training, CNNs initially fit clean samples before addressing noisy ones *(610, 611)*. It's also observed that overparameterized networks trained with first-order optimization methods, such as stochastic gradient descent, can withstand label noise if they incorporate early stopping *(612)*. Bai et al. *(613)* introduces a novel variation of the early stopping strategy, termed progressive early stopping (PES) which is based on the assumption that later layers are more susceptible to label noise than their preceding layers. In the PES method, the earlier layers of the DNN are initially trained over a large number of epochs, thus optimizing these layers extensively. Subsequent layers are then progressively trained over fewer epochs, with the earlier layers kept static to minimize the adverse effects of label noise. Leveraging model behavior in the early training stage, Arazo et al. *(614)* applied a two components beta mixture model (BMM) to each sample's loss values, generating posterior probabilities of clean and noisy samples. These probabilities were then used in a dynamically weighted bootstrapping loss to manage noisy samples without eliminating them. In a different approach, Xia et al. *(611)* propose dividing the network parameters into critical and non-critical categories. Critical parameters are those important for learning from clean labels, while non-critical parameters tend to overfit to noisy labels. To optimize these parameter categories differently, a robust early-learning strategy was employed. Critical parameters undergo a positive update, guided by gradients from the objective function and weight decay, facilitating learning from clean labels. In contrast, non-critical parameters undergo a negative update using only weight decay, without incorporating gradients from the objective function, pushing them towards zero, thereby reducing their contribution and preventing overfitting to noisy labels.

Label smoothing (LS), first introduced in Szegedy et al. *(165)*, is a regularization technique developed to prevent overfitting and improve performance in deep models *(615)*. LS works by converting strict one-hot encoded labels into soft labels by blending the one-hot vector with a uniform distribution over all the classes. This disperses a small part of the confidence for the correct label across all other classes, preventing the model from extreme certainty in its training predictions, thus enhancing generalization. For example, in a three-class land cover semantic segmentation task, let's say the classes are aquaculture pond, mangrove forest, and crop. A one-hot label [1,0,0] designates that a specific pixel belongs to the aquaculture pond class and is not associated with mangrove forest or crop. Conventional label smoothing could tweak this to [0.8, 0.1, 0.1], indicating that the pixel could have slight associations with the mangrove forest and

---

to thick edges in the output. Non-maximum suppression is a method used to thin out these edges by retaining only the pixels with the maximum value along the direction of the gradient and suppressing (setting to zero) the responses of the other, non-maximum pixels. This results in thin, one-pixel wide edges in the output.

crop classes. Variations like margin-based smoothing *(616)* or knowledge distillation *(617)* introduce additional complexity, focusing on widening class margins or mimicking more knowledgeable models using student and teacher settings, respectively, for further improvements. Lukasik et al. *(618)* found LS can be on par with loss correction techniques in mitigating negative effects of the label noise with different rates especially when the smoothing rate is higher than the true label noise rate. Wei et al. *(619)* demonstrated that using negative label smoothing (NLS) on datasets with high label noise rate is more effective than positive LS. NLS is an extension of traditional label smoothing that assigns negative weights to incorrect classes, serving as a form of penalty. For example, for our three-class example, NLS might adjust the labels to [1.2, -0.1, -0.1] which not only signifies high confidence in the aquaculture pond class, but also penalizes predictions for the mangrove forest or crop classes. With a loss function that is linear in labels, the total loss would decrease for predicting aquaculture pond class and increase for mangrove forest or crop, which reinforces the model's preference for the correct class and discourages it from predicting the incorrect ones.

## 8. Conclusion

The advent of deep learning has led to a rapid shift in the field of remote sensing away from conventional classifiers and earlier generation machine learning towards neural network-based approaches for semantic segmentation tasks, commencing an epoch of heightened prediction accuracy and more detailed feature identification. With the burden of feature engineering now eased, research focus has pivoted towards model engineering and learning frameworks that address different aspects of input datasets. In this review, we sought to offer a comprehensive exploration of CNNs, RNNs, and VTs, discussing key factors to consider in implementing these architectures, and their potential benefits and limitations. Our scope was not confined to the field of remote sensing, but rather extended to include cutting-edge work on deep semantic segmentation from diverse disciplines. We introduced ideas and methodologies from these varied fields, as they hold potential for further improving the semantic segmentation of Earth Observation imagery.

Our review underscores the vast array of design choices that need to be made when applying neural networks to semantic segmentation problems, not only with respect to architectural choice (typically the encoder-decoder design), but also other requirements, such as those related to specific loss functions, which can be context-aware and capable of handling unbalanced issues. This profusion of options can be overwhelming, but several common themes and trends emerge from the complexity. Encoders, commonly tasked with capturing a broad spectrum of spatial information from a hierarchy of scales, create multi-scale feature maps. Attention mechanisms have gained traction, particularly for integrating information from different branches or multi-scale features in CNN or RNN architectures. Transformers, known for their attention-based architectures, are increasingly employed as backbones to enhance the model's understanding of the global context, on par or surpassing the performance of CNNs. Decoders generally lean towards models that capture both local and global contexts, hence facilitating the fusion of detailed and broader contextual information. Employing skip connections is widely advocated to conserve spatial information that could potentially be lost during downsampling in the encoder. Additionally, the use of boundary prediction modules is an important factor in decoder design. More recently, hybrid designs incorporating elements of CNNs, RNNs, and/or transformers have become popular due to their complementary abilities for bolstering performance. A rising design trend, Neural Architecture Search (NAS), has potential to reduce manual choice in model design, and may facilitate more rapid advances in the field *(620–624)*. Despite being computationally intensive, NAS is rendered more efficient through the adoption of modular networks, weight sharing *(625)* and strategic scaling techniques *(626)*. However, the success of NAS and the chosen architecture is heavily reliant on task specifications and dataset characteristics, which further underscores the need for meticulous selection and tuning *(627)*.

In addition to the technical facets of model design, we also delved into the inherent challenges creating and using supervised datasets. Conventional wisdom holds that a sufficiently large labeled dataset that reliably represents the domain of interest results in the best performance with supervised learning. However, in practice labeled data are scarce, expensive to procure, and often have limited lifespans, which limits the effectiveness of supervised models. Given this reality, generative models, such as GANs, are increasingly used to augment training data. GANs can generate synthetic yet realistic data, which can increase the diversity and size of training sets, thereby mitigating the challenge of data scarcity, while improving model robustness to common variations in environmental and geographic conditions, such as changing illumination, seasons, or landscape configuration. Furthermore, GANs can be combined with semantic segmentation networks in a technique known as feature-based adversarial training. This approach forces the representations from different domains to be closer in feature space, promoting domain-invariant learning that improves model performance on the target domain. Transfer learning and domain adaptation techniques, such as the use of encoders pre-trained on existing natural image or Earth Observation datasets, are also effective in cases where the target domain differs from those in available labeled datasets. Beyond developing synthetic labels, there is increasing attention on weakly-supervised and semi-supervised learning. By harnessing the potential of unlabeled or partially labeled data, these approaches greatly reduce the need for manual annotation, making them highly valuable in settings where labeled data are sparse. There are several innovative frameworks for weakly-supervised and semi-supervised semantic segmentation, such as simSLR and MixMatch. Each of these methods has the ability to learn robust feature representations that are similar to those obtained in supervised learning, but for a fraction of the labeled data. Notably, GANs are also used within these approaches. Finally, a trend that is further challenging the supremacy of the supervised paradigm is the use of self-supervision for feature extraction and the development of large, pre-trained foundation models. These models require minimal annotated data for fine-tuning to downstream tasks, including semantic segmentation and often achieve results surpassing state-of-the-art supervised models.

In summary, this review emphasizes the importance of tailoring neural network design and training strategies to suit both the specific demands of the task at hand and the unique characteristics of satellite imagery. It highlights the rapid pace of advancements in deep learning, acknowledging that while some studies can yield contradictory outcomes or provide only incremental enhancements over previously tested architectures - often within the confines of benchmark datasets - the future promises a more mature understanding and increased explainability of these models as they get adopted and tested on diverse datasets. These developments, when paired with the innovative integration of various model designs, indicate a promising trajectory for the field of remote sensing towards achieving unprecedented levels of precision and computational efficacy.

321. L.-C. Chen, Y. Yang, J. Wang, W. Xu, A. L. Yuille, "Attention to Scale: Scale-Aware Semantic Image Segmentation" in *2016 IEEE Conference on Computer Vision and Pattern Recognition (CVPR)* (IEEE, Las Vegas, NV, USA, 2016; http://ieeexplore.ieee.org/document/7780765/), pp. 3640–3649.

322. X. Zhang, J. Jin, Z. Lan, C. Li, M. Fan, Y. Wang, X. Yu, Y. Zhang, ICENET: A Semantic Segmentation Deep Network for River Ice by Fusing Positional and Channel-Wise Attentive Features. *Remote Sens.* **12**, 221 (2020).

323. Z. Huang, X. Wang, Y. Wei, L. Huang, H. Shi, W. Liu, T. S. Huang, "CCNet: Criss-Cross Attention for Semantic Segmentation" in *In Proceedings of the IEEE/CVF International Conference on Computer Vision* (2019; http://arxiv.org/abs/1811.11721), pp. 603–612.

324. Li, R., Su, J., Duan, C., Zheng, S., Linear Attention Mechanism: An Efficient Attention for Semantic Segmentation, 4 (2020).

325. R. Li, Zheng, S., Zhang, C., C. Duan, Su, J., Wang, L., Atkinson, P. M., Multi-attention network for semantic segmentation of fine-resolution remote sensing images. *IEEE Trans. Geosci. Remote Sens.*, 13 (2021).

326. H. Zhou, L. Qi, Z. Wan, H. Huang, X. Yang, CANet: Co-attention network for RGB-D semantic segmentation. *Pattern Recognit.* **124**, 108468 (2022).

327. J. Hu, L. Shen, S. Albanie, G. Sun, E. Wu, "Squeeze-and-Excitation Networks" in *In Proceedings of the IEEE conference on computer vision and pattern recognition* (2018; http://arxiv.org/abs/1709.01507), pp. 7132–7141.

328. J. Hu, L. Shen, S. Albanie, G. Sun, A. Vedaldi, Gather-Excite: Exploiting Feature Context in Convolutional Neural Networks. *Adv. Neural Inf. Process. Syst.* **31** (2019) (available at http://arxiv.org/abs/1810.12348).

329. H. Lee, H.-E. Kim, H. Nam, "SRM: A Style-Based Recalibration Module for Convolutional Neural Networks" in *2019 IEEE/CVF International Conference on Computer Vision (ICCV)* (IEEE, Seoul, Korea (South), 2019; https://ieeexplore.ieee.org/document/9008782/), pp. 1854–1862.

330. Q. Wang, B. Wu, P. Zhu, P. Li, W. Zuo, Q. Hu, ECA-Net: Efficient Channel Attention for Deep Convolutional Neural Networks (2020), (available at http://arxiv.org/abs/1910.03151).

331. T. Panboonyuen, K. Jitkajornwanich, S. Lawawirojwong, P. Srestasathiern, P. Vateekul, Semantic Segmentation on Remotely Sensed Images Using an Enhanced Global Convolutional Network with Channel Attention and Domain Specific Transfer Learning. *Remote Sens.* **11**, 83 (2019).

332. M. A. Islam, M. Kowal, S. Jia, K. G. Derpanis, N. D. B. Bruce, Global Pooling, More than Meets the Eye: Position Information is Encoded Channel-Wise in CNNs. *ArXiv210807884 Cs* (2021) (available at http://arxiv.org/abs/2108.07884).

333. L. Ding, H. Tang, L. Bruzzone, LANet: Local Attention Embedding to Improve the Semantic Segmentation of Remote Sensing Images. *IEEE Trans. Geosci. Remote Sens.*, 1–10 (2020).
122

https://link.springer.com/10.1007/978-3-030-01270-0_31), vol. 11220 of *Lecture Notes in Computer Science*, pp. 524–540.

545. M. Tang, A. Djelouah, F. Perazzi, Y. Boykov, C. Schroers, "Normalized Cut Loss for Weakly-Supervised CNN Segmentation" in *2018 IEEE/CVF Conference on Computer Vision and Pattern Recognition* (IEEE, Salt Lake City, UT, 2018; https://ieeexplore.ieee.org/document/8578293/), pp. 1818–1827.

546. Y. Ouali, C. Hudelot, M. Tami, An Overview of Deep Semi-Supervised Learning (2020), (available at http://arxiv.org/abs/2006.05278).

547. Y. Li, T. Shi, Y. Zhang, W. Chen, Z. Wang, H. Li, Learning deep semantic segmentation network under multiple weakly-supervised constraints for cross-domain remote sensing image semantic segmentation. *ISPRS J. Photogramm. Remote Sens.* **175**, 20–33 (2021).

548. Z. Yi, H. Zhang, P. Tan, M. Gong, "DualGAN: Unsupervised Dual Learning for Image-to-Image Translation" in *2017 IEEE International Conference on Computer Vision (ICCV)* (IEEE, Venice, 2017; http://ieeexplore.ieee.org/document/8237572/), pp. 2868–2876.

549. W.-C. Hung, Y.-H. Tsai, Y.-T. Liou, Y.-Y. Lin, M.-H. Yang, Adversarial Learning for Semi-Supervised Semantic Segmentation (2018), (available at http://arxiv.org/abs/1802.07934).

550. K. Mei, C. Zhu, J. Zou, S. Zhang, "Instance Adaptive Self-Training for Unsupervised Domain Adaptation" in *In Computer Vision–ECCV 2020: 16th European Conference* (Springer International Publishing, Glasgow, UK, 2020; http://arxiv.org/abs/2008.12197), vol. Part XXVI 16, pp. 415–430.

551. Y. Ouali, C. Hudelot, M. Tami, "Semi-Supervised Semantic Segmentation With Cross-Consistency Training" in *2020 IEEE/CVF Conference on Computer Vision and Pattern Recognition (CVPR)* (IEEE, Seattle, WA, USA, 2020; https://ieeexplore.ieee.org/document/9157032/), pp. 12671–12681.

552. A. Tarvainen, H. Valpola, "Mean teachers are better role models: Weight-averaged consistency targets improve semi-supervised deep learning results" in *Advances in Neural Information Processing Systems* (2017), vol. 30, pp. 1195–1204.

553. G. French, S. Laine, T. Aila, M. Mackiewicz, G. Finlayson, Semi-supervised semantic segmentation needs strong, varied perturbations (2020), (available at http://arxiv.org/abs/1906.01916).

554. V. Olsson, W. Tranheden, J. Pinto, L. Svensson, "ClassMix: Segmentation-Based Data Augmentation for Semi-Supervised Learning" in *2021 IEEE Winter Conference on Applications of Computer Vision (WACV)* (IEEE, Waikoloa, HI, USA, 2021; https://ieeexplore.ieee.org/document/9423297/), pp. 1368–1377.

555. K. Sohn, D. Berthelot, C.-L. Li, Z. Zhang, N. Carlini, E. D. Cubuk, A. Kurakin, H. Zhang, C. Raffel, "FixMatch: Simplifying Semi-Supervised Learning with Consistency and Confidence" in *Advances in neural information processing systems* (2020), vol. 33, pp. 596–608.

556. Y. Xu, L. Shang, J. Ye, Q. Qian, Y.-F. Li, B. Sun, H. Li, R. Jin, "Dash: Semi-Supervised Learning with Dynamic Thresholding" in *In International Conference on Machine Learning* (2021), pp. 11525–11536.